\documentclass{article}
\usepackage[preprint,nonatbib]{conf}
\usepackage[utf8]{inputenc} % allow utf-8 input
\usepackage[T1]{fontenc}    % use 8-bit T1 fonts
\usepackage{xcolor}         % colors
\definecolor{bleudefrance}{rgb}{0.19, 0.55, 0.91}
\usepackage[pagebackref,breaklinks,colorlinks,citecolor=bleudefrance]{hyperref}    % hyperlinks
\usepackage{url}            % simple URL typesetting
\usepackage{booktabs}       % professional-quality tables
\usepackage{amsfonts}       % blackboard math symbols
\usepackage{nicefrac}       % compact symbols for 1/2, etc.
\usepackage{microtype}      % microtypography

\usepackage{graphicx}

\usepackage[accsupp]{axessibility}  % Improves PDF readability for those with disabilities.

\usepackage{array}
\usepackage{multirow}
\usepackage{makecell}
\usepackage{multicol}
\usepackage{wrapfig}

\usepackage[capitalize]{cleveref}

\usepackage{animate}

\usepackage{enumerate}
\usepackage{enumitem}
\usepackage{chngcntr}

\usepackage{tcolorbox}

\usepackage{xspace}
\newcommand*{\eg}{e.g.\@\xspace}
\newcommand*{\ie}{i.e.\@\xspace}

\newcommand*{\newcite}[1]{~\cite{#1}}
\newcommand{\norm}[1]{\left\lVert#1\right\rVert}
\newcommand{\myquote}[1]{``#1''}

\newcommand{\ourdm}{VSTAR}
\newcommand{\ourrecaption}{VSP}
\definecolor{RWTHblue}{RGB}{0,85,169}%{0,84,159}

\renewcommand{\paragraph}[2][\ ]{\vspace{4pt}\noindent{\bf #2#1}}

\usepackage{comment}

\title{VSTAR: Generative Temporal Nursing for Longer Dynamic Video Synthesis}

\author{%
  Yumeng Li$^{1,2}$ \  \  William Beluch$^{1}$\  \  Margret Keuper$^{2,3}$\ \ Dan Zhang$^{1,4}$ \ \ Anna Khoreva$^{1}$ \\
  $^1$Bosch Center for Artificial Intelligence $^2$University of Mannheim \\
$^3$Max Planck Institute for Informatics 
$^4$University of T\"ubingen \\
\texttt{\{yumeng.li, william.beluch, dan.zhang2, anna.khoreva\}@de.bosch.com} \\
\ \ \texttt{keuper@uni-mannheim.de} \\
\href{https://yumengli007.github.io/VSTAR}{Project page: https://yumengli007.github.io/VSTAR}
}

\begin{document}
\maketitle

\begin{abstract} 
Despite tremendous progress in the field of text-to-video (T2V) synthesis, open-sourced T2V diffusion models struggle to generate longer videos with dynamically varying and evolving content. They tend to synthesize quasi-static videos, ignoring the necessary visual change-over-time implied in the text prompt. At the same time, scaling these models to enable longer, more dynamic video synthesis often remains computationally intractable. To address this challenge, we introduce the concept of Generative Temporal Nursing (GTN), where we aim to alter the generative process on the fly during inference to improve control over the temporal dynamics and enable generation of longer videos. We propose a method for GTN, dubbed {\ourdm}, which consists of two key ingredients: 
1) \textbf{V}ideo \textbf{S}ynopsis Prompting (VSP) - automatic generation of a video synopsis based on the original single prompt leveraging LLMs, which gives accurate textual guidance to different visual states of longer videos, and 2) \textbf{T}emporal \textbf{A}ttention \textbf{R}egularization (TAR) - a regularization technique to refine the temporal attention units of the pre-trained T2V diffusion models, which enables control over the video dynamics. 
We experimentally showcase the superiority of the proposed approach in generating longer, visually appealing videos over existing open-sourced T2V models. We additionally analyze the temporal attention maps realized with and without {\ourdm}, demonstrating the importance of applying our method to mitigate neglect of the desired visual change over time.

\vspace{-1em}

\end{abstract}

\begin{figure}[ht!]
\vspace{-1.0em}
\begin{centering}
\setlength{\tabcolsep}{0.0em}
\renewcommand{\arraystretch}{0}
\par\end{centering}
\begin{centering}
\hfill{}%
 \begin{tabular}{
    m{0.0\linewidth}<{\centering} @{}
    m{0.16\linewidth}<{\centering} @{\hspace{0.015\linewidth}}
    m{0.16\linewidth}<{\centering} @{}
    m{0.16\linewidth}<{\centering} @{} %
    m{0.16\linewidth}<{\centering} @{}
    m{0.16\linewidth}<{\centering} @{}
    m{0.16\linewidth}<{\centering} @{}
    }
\tabularnewline
& & \multicolumn{5}{c}{\begin{tabular}{c}
\small \myquote{The process of a lava eruption, from smoke emission to lava cooling}
\end{tabular}} 
\tabularnewline
    & \multirow{2}{*}{\begin{tabular}{c}
   \footnotesize VideoCrafter2\\
    \animategraphics[autoplay,loop,width=0.98\linewidth,height=0.65\linewidth]{6}{figs/teaser/volcano/baseline/frame_}{0}{63}
    \end{tabular}}
    & \includegraphics[width=0.98\linewidth,height=0.65\linewidth]{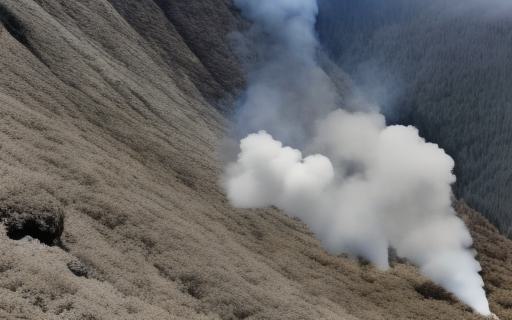}
    & \includegraphics[width=0.98\linewidth,height=0.65\linewidth]{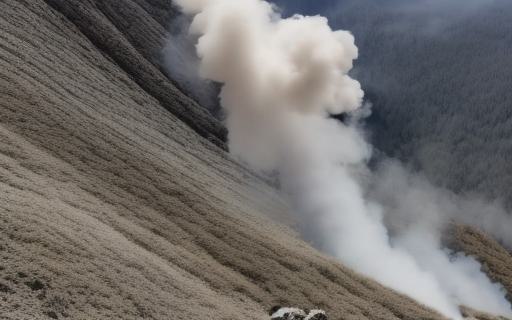} 
    & \includegraphics[width=0.98\linewidth,height=0.65\linewidth]{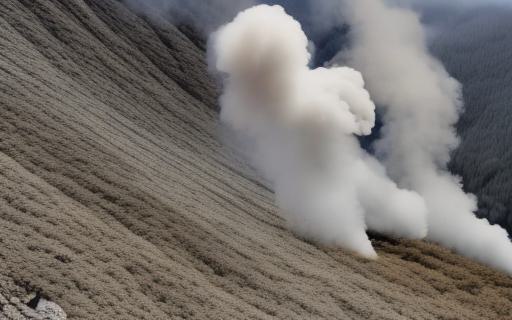} 
    & \includegraphics[width=0.98\linewidth,height=0.65\linewidth]{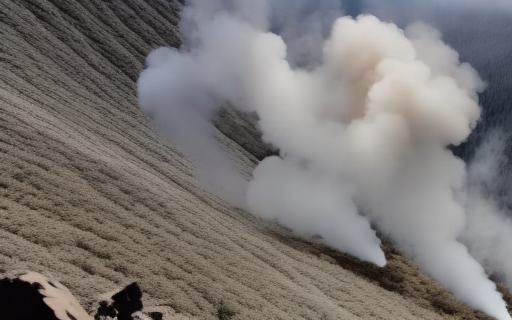}  
    & \includegraphics[width=0.98\linewidth,height=0.65\linewidth]{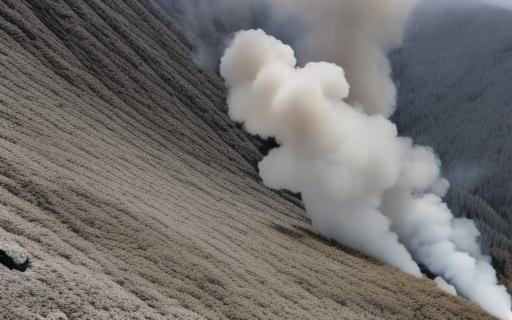}
\tabularnewline
    & %
    & \includegraphics[width=0.98\linewidth,height=0.65\linewidth]{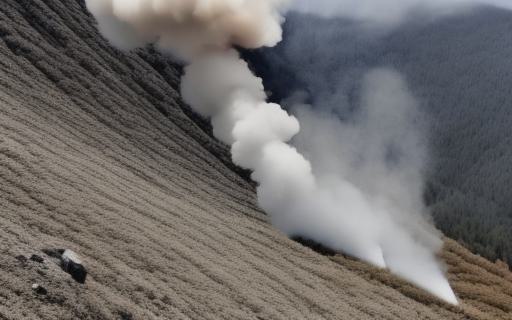}
    & \includegraphics[width=0.98\linewidth,height=0.65\linewidth]{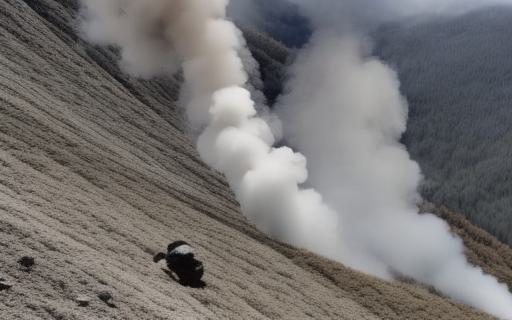} 
    & \includegraphics[width=0.98\linewidth,height=0.65\linewidth]{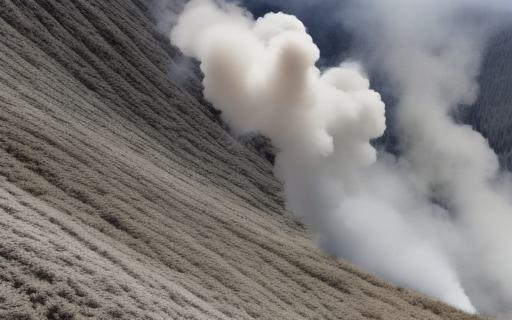} 
    & \includegraphics[width=0.98\linewidth,height=0.65\linewidth]{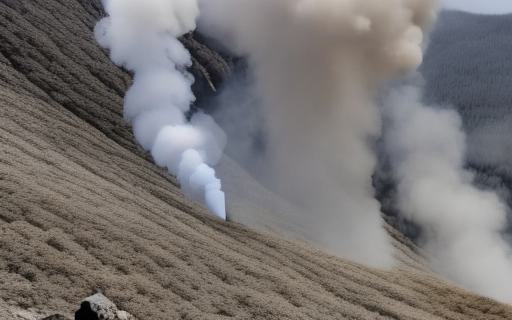} 
    & \includegraphics[width=0.98\linewidth,height=0.65\linewidth]{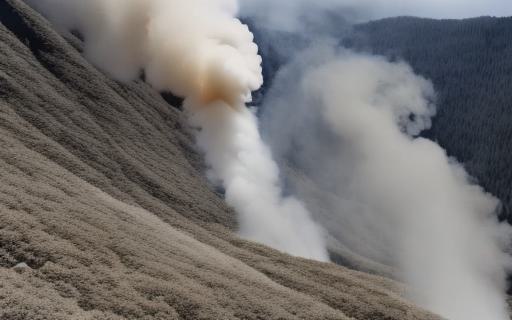}
\tabularnewline

\tabularnewline
    & \multirow{2}{*}{\begin{tabular}{c}
    \footnotesize \textbf{VSTAR} \\
    \animategraphics[autoplay,loop,width=0.98\linewidth,height=0.65\linewidth]{6}{figs/teaser/volcano/ours_935/frame_}{0}{63}
    \end{tabular}}
    & \includegraphics[width=0.98\linewidth,height=0.65\linewidth]{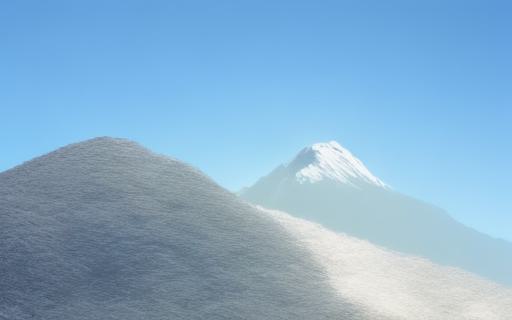}
    & \includegraphics[width=0.98\linewidth,height=0.65\linewidth]{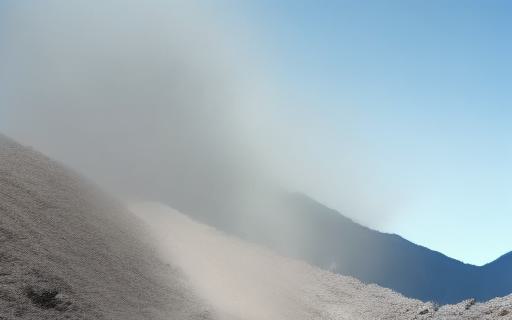} 
    & \includegraphics[width=0.98\linewidth,height=0.65\linewidth]{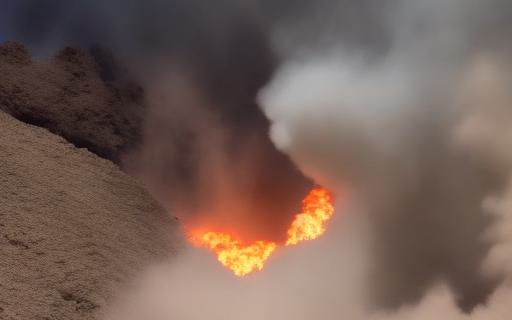} 
    & \includegraphics[width=0.98\linewidth,height=0.65\linewidth]{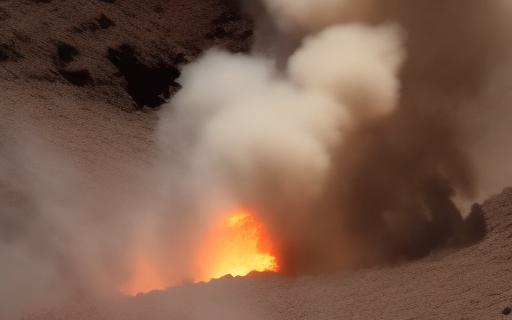}  
    & \includegraphics[width=0.98\linewidth,height=0.65\linewidth]{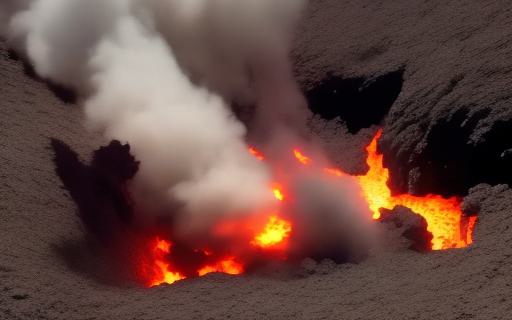}
\tabularnewline
    & %
    & \includegraphics[width=0.98\linewidth,height=0.65\linewidth]{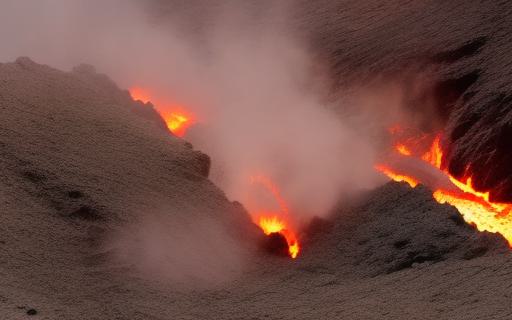}
    & \includegraphics[width=0.98\linewidth,height=0.65\linewidth]{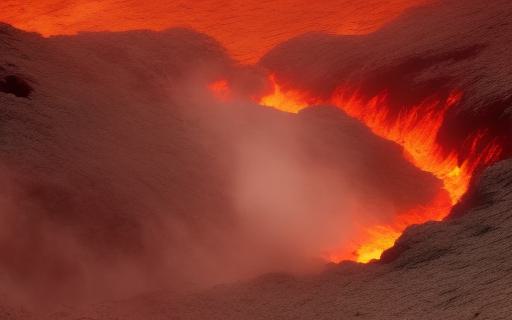} 
    & \includegraphics[width=0.98\linewidth,height=0.65\linewidth]{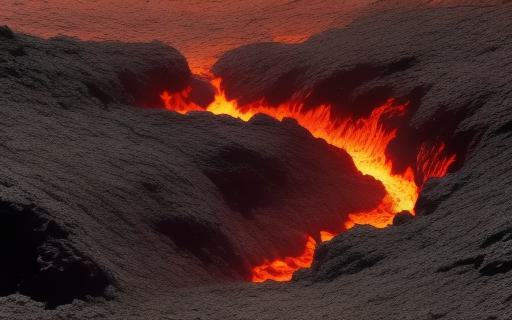} 
    & \includegraphics[width=0.98\linewidth,height=0.65\linewidth]{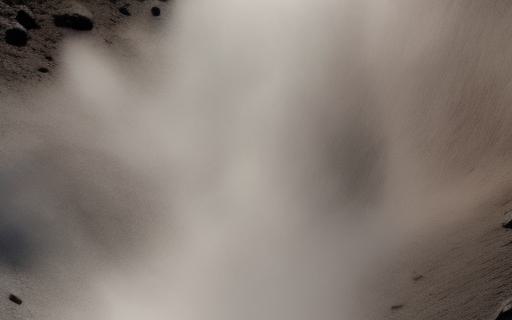} 
    & \includegraphics[width=0.98\linewidth,height=0.65\linewidth]{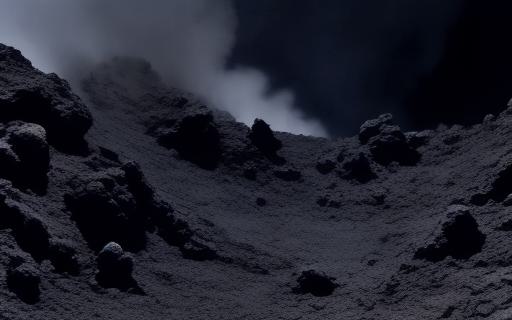}
\tabularnewline
\end{tabular}
\hfill{}
\par\end{centering}
\vspace{-0.5em}
\caption{Our {\ourdm} can generate a 64-frame video with dynamic visual evolution in a \emph{single} pass. Images are subsampled from the video.
Note that the first column is a GIF, best viewed in \emph{Acrobat Reader}.
}\label{fig:teaser}
\vspace{-0.5em}
\end{figure}

\section{Introduction}\label{sec:intro}
\vspace{-0.5em}
Driven by a whirlwind of activity from both published research and the open-source community, text-to-image synthesis and its natural extension to text-to-video synthesis have undergone remarkable progress in the past few years. Having transformed the idea of content creation, they are now widespread as both a research topic and an industry application. In the realm of text-to-video (T2V) synthesis specifically, recent advancements in video diffusion models\newcite{blattmann2023align,guo2024animatediff,wang2023lavie,chen2023videocrafter1,chen2024videocrafter2,wang2023modelscope,sora} have sparked promising progress, offering improved possibilities for creating novel video content from textual descriptions.

However, despite these advancements, we observe two common issues in current open-source T2V models\newcite{guo2024animatediff,wang2023lavie,chen2023videocrafter1,chen2024videocrafter2,wang2023modelscope}: limited visual changes within the video, and a poor ability to generate longer videos with coherent temporal dynamics. 
More specifically, the synthesized scenes often exhibit a high degree of similarity between frames (see \cref{fig:teaser}), frequently resembling a static image with minor variations as opposed to a video with varying and evolving content.
Additionally, these models do not generalize well to generate videos with more than the typical 16 frames in one pass (see \cref{fig:compare_t2v_32}). While several recent works attempt to generate long videos in a sliding window fashion\newcite{qiu2023freenoise,wang2023genLvideo}, the methods not only introduce considerable overhead due to requiring multiple passes, but also face the new challenge of preserving temporal coherence throughout these passes.

To mitigate the aforementioned issues, we propose the concept of \emph{``Generative Temporal Nursing''}(GTN), which aims to improve the temporal dynamics of (long) video synthesis on the fly during inference, without re-training T2V models, and using a single pass to not induce a high computational overhead. As a form of GTN, we propose {\ourdm}, consisting of \textbf{V}ideo \textbf{S}ynopsis Prompting (VSP) and \textbf{T}emporal \textbf{A}ttention \textbf{R}egularization (TAR).

Current open-sourced T2V models, such as ModelScope\newcite{wang2023modelscope}, LaVie\newcite{wang2023lavie} and VideoCrafter\newcite{chen2023videocrafter1,chen2024videocrafter2}, are built upon T2I models, and process all frames within one batch. The single text prompt is conditioned via cross-attention in the spatial transformer of the UNet and shared by all frames.
However, it is challenging for the T2V models to transform the semantics from a single prompt into the required visual change across frames, especially when a video with high dynamics is desired, as shown in \cref{fig:teaser}.
For dynamic video synthesis faithful to the input prompt, the generation could benefit from a \textit{synopsis} that describes the main events of the video, with explicit descriptions about the desired visual development over time.
As a method to provide this guidance and better disseminate the single input prompt across frames, 
the first strategy of GTN, Video Synopsis Prompting ({\ourrecaption}) leverages the ability of large language models (LLMs), \eg, ChatGPT\newcite{chatgpt}, to decompose the single input prompt describing a dynamic transition into several stages of visual development. More specifically, thanks to their in-context learning capability\newcite{brown2020GPT,hu2022context}, LLMs can be instructed to perform such synopsis prompting automatically by providing a few (or even one) concrete examples. {\ourrecaption} can thus provide the T2V model more accurate guidance on individual visual states, encouraging diversity from the spatial perspective.

Next, we investigate the architectural units of T2V models introduced to capture the temporal interactions between frames. These units, newly incorporated into the T2I backbone, are based on temporal transformers consisting of self-attention layers\newcite{guo2024animatediff,chen2023videocrafter1,wang2023lavie,wang2023modelscope}. 
Naturally, this temporal attention serves as a critical component in driving the dynamic aspects of video synthesis.
Previous work on T2I generation has shown that cross-attention, as the only interaction between the UNet and the input text prompt, can be manipulated to steer the image generation process, \eg control the image layout or improve attribute binding\newcite{chefer2023attendandexcite,li2023divide,prompt2prompt,feng2022trainingfree,chen2024trainingfreeCAcontrol}. 
A resulting natural question is, \emph{can we improve the dynamics of video synthesis by manipulating the temporal attention?}
Observing the visual gap between real videos and synthesized ones leads us to compare their temporal attention maps (see \cref{fig:compare_TA_real_fake}). 
We discover that real videos have a band-matrix-like structure, indicating high temporal correlation among adjacent frames and reduced correlation with frames further apart.
Intriguingly, the attention maps of the synthesized ones are less structured, potentially explaining their inferior temporal dynamics.

Inspired by this observation, we propose a simple yet effective Temporal Attention Regularization (TAR) strategy to improve the video dynamics of generated videos. %
More specifically, we design a symmetric Toeplitz matrix with values along the off-diagonal direction following a
Gaussian distribution. The standard deviation of this distribution can control the regularization strength, \ie, the visual variation along the temporal dimension. Adding it to the existing temporal attention maps strengthens the temporal correlation between adjacent frames, while reducing it between more distant frames. 
Notably, TAR is readily applicable to pre-trained T2V models and requires no optimization, thus introducing no extra inference overhead.
Equipped with both strategies, our {\ourdm} can produce long videos with appealing visual changes in one single pass.

Finally, we analyze the temporal attention mechanisms of different T2V models, establishing valuable connections between their capability to generate longer videos and their architectures. Following the analysis, we offer several training suggestions for enhancing the generalization ability of future models.

In summary, our contributions include:
\begin{itemize}
    \item We introduce a novel concept of \myquote{Generative Temporal Nursing}, aiming to improve temporal dynamics, especially for long videos, without requiring any training or introducing high computational overhead at inference time.
    \item We propose {\ourdm}, a method for Generative Temporal Nursing, consisting of two simple yet effective strategies: Video Synopsis Prompting and Temporal Attention Regularization, which enable long video generation in a single pass with improved video dynamics. %
    \item We are the first to provide an analysis of temporal attention within video diffusion models, and unleash its potential for controlling the video dynamics. Based on the analysis, we provide insights on how to improve the training of the next generation of T2V models.  
\end{itemize}

\section{Related Work}\label{sec:related-work}
\vspace{-0.8em}
\paragraph{Text-to-Video Diffusion Models.}
\label{subsec:related-t2v}
Recent text-to-video diffusion models\newcite{blattmann2023align,guo2024animatediff,wang2023lavie,chen2023videocrafter1,chen2024videocrafter2,wang2023modelscope} are commonly built upon large-scale pretrained T2I model, \eg, Stable Diffusion\newcite{rombach2022SD}.
Such methods generally introduce a temporal dimension to the T2I model and incorporate temporal transformer for temporal modeling and fine-tune on a video dataset, however differ in their design choice of the temporal units and fine-tuning process.
ModelScope\newcite{wang2023modelscope} and VideoCrafter\newcite{chen2023videocrafter1} similarly inserting the temporal attention after spatial units within the UNet.
LaVie\newcite{wang2023lavie} and AnimateDiff\newcite{guo2024animatediff} additionally employed Rotary Positional Encoding\newcite{touvron2023llama} and Sinusoidal Positional Encoding based on the frame indices, respectively. 
More recently, VideoCrafter2\cite{chen2024videocrafter2} adopted the architecture of its predecessor, and advance the fine-tuning process by enriching existing video datasets with high-quality image data, achieving state-of-the-art T2V performance. 

Since long video generation is especially difficult, there are also works\newcite{qiu2023freenoise,wang2023genLvideo} focusing specifically on this application.  FreeNoise\newcite{qiu2023freenoise} proposed noise rescheduling %
combined with local window based attention fusion. 
Gen-L-Video\newcite{wang2023genLvideo} casts the problem as fusing multiple short video clips with temporal overlapping. However, they necessitate several passes for generation, significantly raising the inference overhead.
Different from these methods, our {\ourdm} targets at long video generation with a pretrained T2V model in one \emph{single} pass.

\paragraph{Attention Manipulation.}\label{subsec:related-attention}
In the realm of T2I models, many works\newcite{cao2023masactrl,chefer2023attendandexcite,li2023divide,chen2024trainingfreeCAcontrol,prompt2prompt,feng2022trainingfree} have identified the attention layers as potential targets to manipulate for improving synthesis. 
\newcite{chefer2023attendandexcite,li2023divide} employs inference time latent optimization based on the cross-attention maps, to enhance faithfulness to the input prompt, \eg, encourage object presence and proper attribute binding. However, such optimization increases the computation cost at inference time.
There are also methods\cite{prompt2prompt,feng2022trainingfree} directly modifies or reweights the attention maps to enable text-controlled image editing or improve attribute binding and compositionality.  
Nonetheless, for T2V diffusion models, there still lacks of comprehensive understanding of the temporal attention mechanism. Our work is the first to investigate this aspect and unleash its manipulation potential for improving video generation of pretrained T2V models without extra optimization overhead during inference.

\begin{figure*}[t]
\vspace{-0.5em}
\centering
\includegraphics[width=0.95\linewidth]{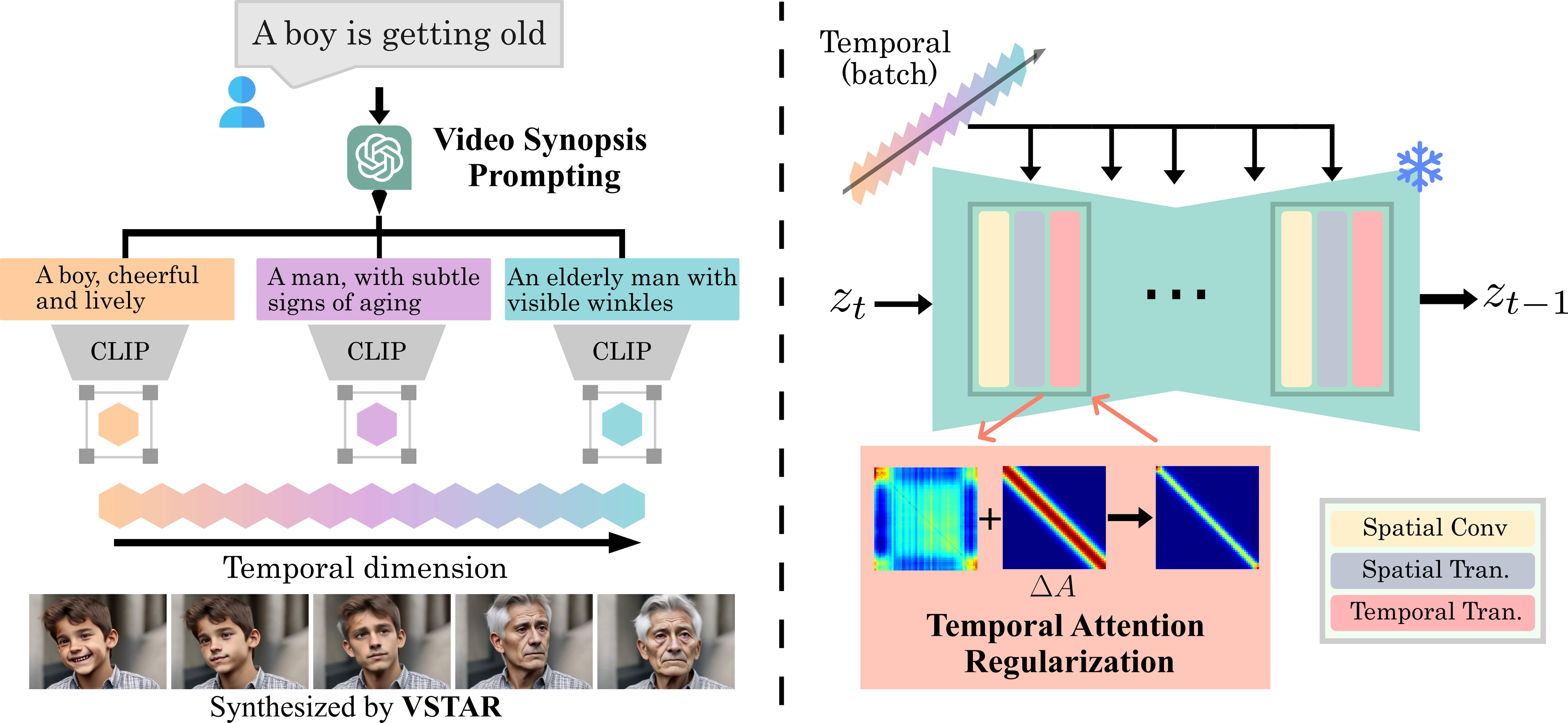}
\caption{
Method overview. Our {\ourdm} consists of two strategies: Video Synopsis Prompting (left) and Temporal Attention Regularization (right).
}
\label{fig:overview}
\end{figure*}

\vspace{-0.5em}
\section{Method}\label{sec:method}
\vspace{-0.5em}

Our concept of Generative Temporal Nursing (GTN) aims at improving the video dynamics of pre-trained T2V diffusion models. Besides the text prompt, we identify in \cref{subsec:method-preliminary} that the temporal attention layer is a further key component of T2V models responsible for determining video dynamics. Our first GTN strategy, Video Synopsis Prompting (\cref{subsec:method-Recap}), expands the initial text prompt for the whole video into a sequence of detailed descriptions that control the video progression respectively on different frames. Being inspired by the temporal attention analysis in \cref{sec:attention-analysis} on real videos, we next design a simple yet effective Temporal Attention Regularization (\cref{subsec:method-TA-reg}), encouraging the temporal attention of synthetic videos to mimic the attention of real videos.

\subsection{Preliminary: Text-to-Video Diffusion Model} \label{subsec:method-preliminary}
Current open-sourced text-to-video (T2V) diffusion models\newcite{wang2023modelscope,wang2023lavie,chen2023videocrafter1,chen2024videocrafter2} share a similar high-level design, even if training strategies and specific implementations vary. Based on the text-to-image (T2I) latent diffusion model, \eg, Stable Diffusion (SD)\newcite{rombach2022SD}, two main changes are introduced for video diffusion models: inflating the 2D UNet to a 3D UNet and adding temporal transformers to capture the requisite temporal relationship found between video frames. With the addition of a temporal axis to the 2D convolutional kernels of SD, the resulting pseudo-3D convolutional layers can handle the input video latent $z\in \mathbb{R}^{N\times C \times H \times W}$, where $N$ is the number of frames and $C,H,W$ represent the channel and spatial dimension of each frame in the latent space, respectively. 
To generate a video of $N$ frames given a text prompt, current T2V methods~\cite{wang2023modelscope,wang2023lavie,chen2023videocrafter1,chen2024videocrafter2,guo2024animatediff} process all $N$ frames within one batch, and simply repeat the same prompt embedding for all frames.  Inherently, the provided text prompt is conditioned via cross-attention of the spatial transformer in the UNet. 
The temporal transformer consists of several self-attention layers that operate along the temporal axis. More specifically, the spatial dimension of the intermediate features is merged into the batch dimension, resulting in a shape of $(B \times h\times w, N)$. Since the spatial layers inherited from SD can only handle each frame independently, the temporal attention layers thus play a crucial role for modeling the video dynamics.

\vspace{-0.5em}
\subsection{Video Synopsis Prompting (VSP)}
\label{subsec:method-Recap}
Similar to T2I models, T2V models shall generate the desired content information based on the text prompt. T2I models already struggle with handling complicated text prompts, particularly when required to properly compose a scene and correctly place relative content spatially~\cite{yang2023reco,feng2022trainingfree,yang2024mastering,wang2024instancediffusion}. The lack of semantic understanding, reasoning, and planning of the synthesis models results in low quality outputs. The issue becomes more critical when moving from image to video synthesis, as the evolution of the scenes must also now be considered. For example, the text prompt \myquote{A landscape transitioning from winter to spring} is highly abstract; the seasonal change from winter to spring inherently can consist of several visual states. As shown in \cref{fig:compare_t2v_32}, the SOTA T2V model VideoCrafter2~\cite{chen2024videocrafter2} fails to generate such dynamic changes. 

\begin{wrapfigure}{R}{0.5\textwidth}
\vspace{-1em}
\centering
\includegraphics[width=0.99\linewidth]{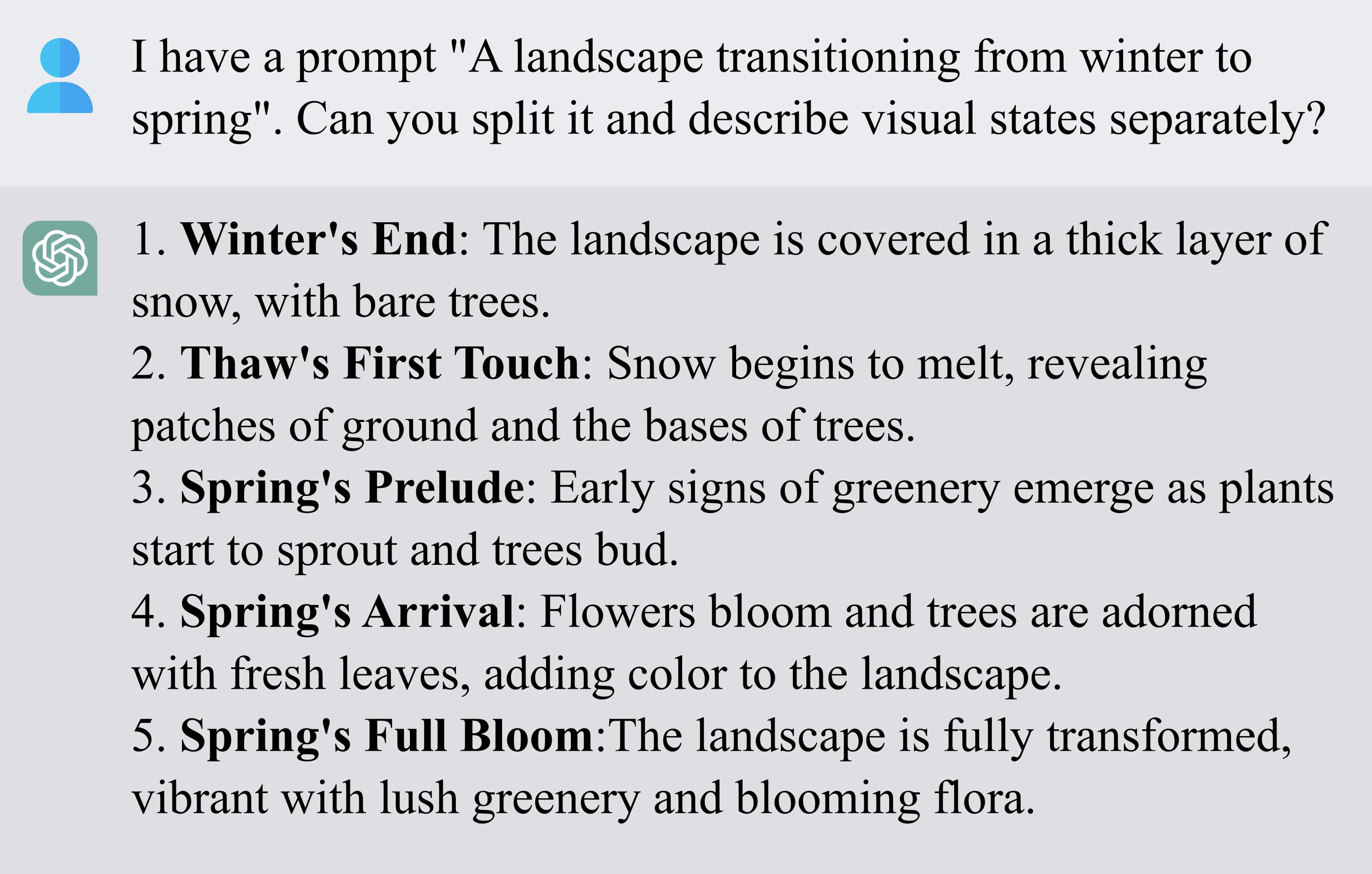}
\caption{ An illustration example of VSP. With the aid of LLMs, we can obtain more descriptive video synopsis for key stages.
}
\label{fig:method-VSP}
\vspace{-1em}
\end{wrapfigure}

Inspired by the creation of long dynamic videos in real life, we propose to offload the task of interpreting the text prompt, reasoning about it, and creating a video synopsis to LLMs. 
This task can be effectively managed in the language space, where LLMs have presented strong generalization across various tasks.  
When we ask ChatGPT\newcite{chatgpt} to parse the same text prompt, i.e., \myquote{A landscape transitioning from winter to spring}, into a sequence of text descriptions that well describe the dynamics, the result is more convincing and semantically informative as shown in \cref{fig:method-VSP}.
The detailed instruction template can be found in Supp. Material. %

It is sufficient to generate text descriptions for the main event changes in a video rather than for each frame. A text encoder \eg, CLIP text decoder\cite{radford2021clip}, is then applied to extract the text embeddings of these descriptions, which are then interpolated to guide each frame's synthesis via cross attention as illustrated in \cref{fig:overview}. 
This process yields more accurate guidance for transitioning visual stages, while ensuring smooth conditioning without abrupt changes between frames.

\begin{figure*}[t]
\vspace{-0.5em}
\begin{centering}
\setlength{\tabcolsep}{0.0em}
\renewcommand{\arraystretch}{0}
\par\end{centering}
\begin{centering}
\hfill{}%
\footnotesize
	\begin{tabular}{@{\hspace{0.2em}}c@{\hspace{0.2em}}c@{\hspace{0.2em}}
 c@{\hspace{0.2em}}c@{\hspace{0.2em}}c@{\hspace{0.2em}}c@{\hspace{0.2em}}c@{\hspace{0.2em}}c@{\hspace{0.2em}}c
 }
	\centering
    &
    \multicolumn{2}{c}{
    \begin{tabular}{c}{Real 16 frames}\end{tabular}
    } 
    & 
    \multicolumn{2}{c}{
    \begin{tabular}{c}{Synthetic 16 frames}\end{tabular}
    } 
    &
    \multicolumn{2}{c}{
    \begin{tabular}{c}{Real 48 frames}\end{tabular}
    } 
    & 
    \multicolumn{2}{c}{
    \begin{tabular}{c}{Synthetic 48 frames}\end{tabular}
    } 
    
    \tabularnewline
     & Res. 64 & Res. 32 & Res. 64 & Res. 32 & Res. 64 & Res. 32 & Res. 64 & Res. 32 
    \tabularnewline
    &
	\includegraphics[width=0.12\linewidth, height=0.07\textheight]{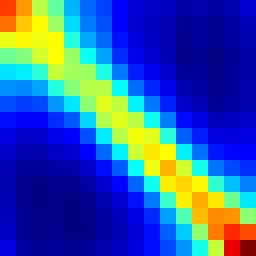}
    &
	\includegraphics[width=0.12\linewidth, height=0.07\textheight]{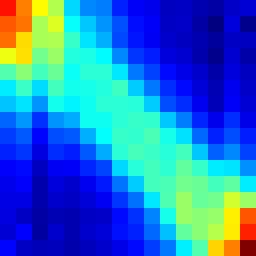}
    & 
	\includegraphics[width=0.12\linewidth, height=0.07\textheight]{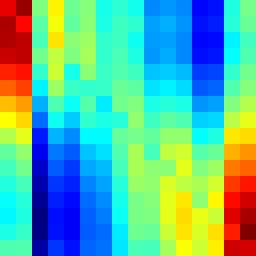}
    & 
    \includegraphics[width=0.12\linewidth, height=0.07\textheight]{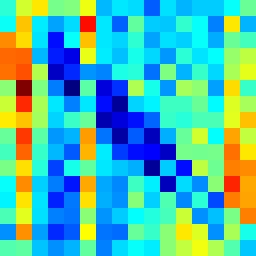}
    &
	\includegraphics[width=0.12\linewidth, height=0.07\textheight]{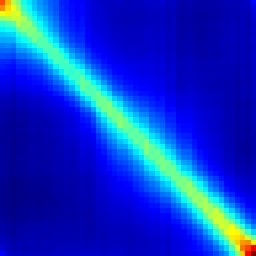}
    &
	\includegraphics[width=0.12\linewidth, height=0.07\textheight]{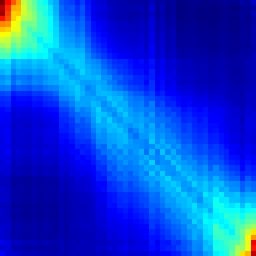}
    & 
	\includegraphics[width=0.12\linewidth, height=0.07\textheight]{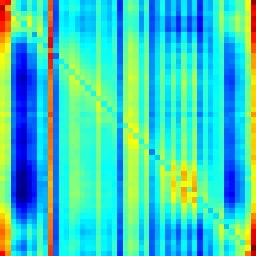}
    & 
    \includegraphics[width=0.12\linewidth, height=0.07\textheight]{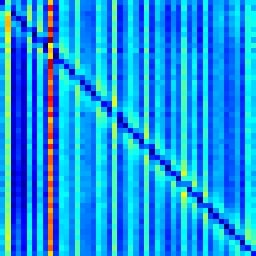}
  \tabularnewline
\end{tabular}
\hfill{}
\par\end{centering}
\vspace{-0.5em}
\caption{Temporal attention visualization of real and synthetic videos of 16 and 48 frames. Attention of real videos exhibits a band-matrix like structure, indicating high correlation with adjacent frames. Synthetic videos exhibit less-structured attention maps, especially for 48 frames, which explains the low quality of long video generation.
} 
\label{fig:compare_TA_real_fake}
\end{figure*}
\begin{figure}[t]
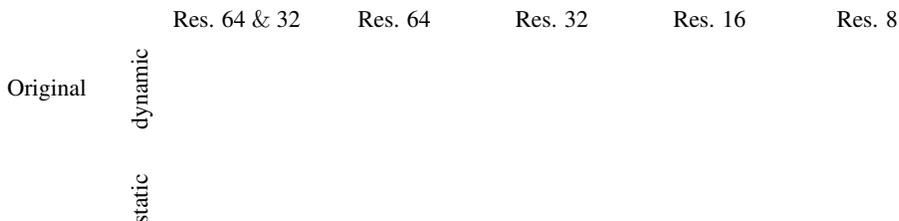

\begin{centering}
\setlength{\tabcolsep}{0.0em}
\renewcommand{\arraystretch}{0}
\par\end{centering}
\begin{centering}
\hfill{}%
 \footnotesize
 \begin{tabular}{
    m{0.15\linewidth}<{\centering} @{\hspace{0.02\linewidth}}
    m{0.03\linewidth}<{\centering} @{}
    m{0.15\linewidth}<{\centering} @{}
    m{0.15\linewidth}<{\centering} @{} %
    m{0.15\linewidth}<{\centering} @{}
    m{0.15\linewidth}<{\centering} @{}
    m{0.15\linewidth}<{\centering} @{}
    }
     &  & Res. 64 $\&$ 32 & Res. 64 & Res. 32 & Res. 16 & Res. 8
\tabularnewline
    \multirow{2}{*}{ \begin{tabular}{c}
    Original\\\animategraphics[autoplay,loop,width=0.98\linewidth,height=0.65\linewidth]{4}{figs/per_res_TA/orig/frame_}{0}{15}
    \end{tabular}
    } 
    &
    \multirow{1}{*}
    {\rotatebox{90}{
        \footnotesize  dynamic
        \hspace{-1.3\linewidth}
    }} 
    & \animategraphics[autoplay,loop,width=0.98\linewidth,height=0.65\linewidth]{4}{figs/per_res_TA/ones_64_32/frame_}{0}{15}
    & \animategraphics[autoplay,loop,width=0.98\linewidth,height=0.65\linewidth]{4}{figs/per_res_TA/ones_64/frame_}{0}{15}   
    & \animategraphics[autoplay,loop,width=0.98\linewidth,height=0.65\linewidth]{4}{figs/per_res_TA/ones_32/frame_}{0}{15}   
    & \animategraphics[autoplay,loop,width=0.98\linewidth,height=0.65\linewidth]{4}{figs/per_res_TA/ones_16/frame_}{0}{15}
    & \animategraphics[autoplay,loop,width=0.98\linewidth,height=0.65\linewidth]{4}{figs/per_res_TA/ones_8/frame_}{0}{15}
\tabularnewline

    &
    \multirow{1}{*}
    {\rotatebox{90}{
        \hspace{-1.3\linewidth}
        \begin{tabular}{c}
        \footnotesize static
        \end{tabular}
         \hspace{-1.3\linewidth}
    }}
    & \animategraphics[autoplay,loop,width=0.98\linewidth,height=0.65\linewidth]{4}{figs/per_res_TA/zeros_64_32/frame_}{0}{15}
    & \animategraphics[autoplay,loop,width=0.98\linewidth,height=0.65\linewidth]{4}{figs/per_res_TA/zeros_64/frame_}{0}{15}   
    & \animategraphics[autoplay,loop,width=0.98\linewidth,height=0.65\linewidth]{4}{figs/per_res_TA/zeros_32/frame_}{0}{15}   
    & \animategraphics[autoplay,loop,width=0.98\linewidth,height=0.65\linewidth]{4}{figs/per_res_TA/zeros_16/frame_}{0}{15}
    & \animategraphics[autoplay,loop,width=0.98\linewidth,height=0.65\linewidth]{4}{figs/per_res_TA/zeros_8/frame_}{0}{15}
\tabularnewline
\end{tabular}
\hfill{}
\par\end{centering}
\vspace{-0.5em}
\caption{Per-layer temporal attention analysis. 
We replace the temporal attention maps at different resolutions with a diagonal matrix (1st row) and an all-ones matrix (2nd row), which leads to a more dynamic or a more static video, respectively. We observe that high resolution attention has a larger impact on the video dynamics.  
Note that this is a GIF, best viewed in \emph{Acrobat Reader}. 
}
\label{fig:ablation_per_res_TA}
\end{figure}

\vspace{-0.5em}
\subsection{Temporal Attention Analysis}\label{sec:attention-analysis}

To properly synthesize videos that capture the dynamics conveyed in the input prompt,
we delve into the synthesis model itself. An examination of the components new to T2V models, beyond the common building blocks already used in T2I models, leads naturally to the temporal attention layers.  These new modules 
are crucial for facilitating proper video synthesis, \ie, generating sequential frames with dynamic yet consistent content that reflect the input text information. 
We hypothesize that the ineffectiveness of current T2V models arises from unstructured interactions among frames in the same video within the temporal attention layers. To verify our hypothesis, we conduct a systematic analysis comparing the attention maps of real and synthetic videos.
Specifically, the attention map A is expressed as:
\begin{align} \label{eq:attention}
A = Softmax\left( \phi\left(Q,K\right) \right) = 
 Softmax\left(\frac{QK^T}{\sqrt{d}}\right) \in \mathbb{R}^{N\times N}, 
\end{align}
where $Q$ and $K$ represent the query and key of the self-attention layer, and $d$ is the latent dimension. 
This attention matrix essentially depicts the pairwise correlation between the $N$ frames of one video. For real videos, their attention maps can be obtained by adding noise to their clean latent and extracting the attention during the denoising process. For synthetic videos, we can read out their attention maps directly during their synthesis passes.

As shown in \cref{fig:compare_TA_real_fake}, for both 16 and 48 frame real videos, the attention matrix manifests as a band-matrix-like structure. Intuitively, closer frames should have a higher correlation with each other to maintain temporal coherency. 
Compared to real videos, attention matrix of the synthetic ones is less structured, especially for 48 frames. That explains why the model generalizes even worse to longer videos. High correlation is spread across a wide range of frames, resulting in a harmonized sequence with similar appearances.

Further, we conducted a per-resolution ablation as shown in \cref{fig:ablation_per_res_TA}. We replace the attention map at each individual resolution, \ie, $64$, $32$, $16$, and $8$, while keeping the other resolution untouched. We experiment with two extreme cases: using the Identity matrix ($I_N$) and the all-ones matrix ($J_N$). The former regularizes the frames to be mutually independent, while the latter oppositely requires full correlation, i.e., static sequence. 
The observations from \cref{fig:ablation_per_res_TA} are highly consistent. When utilizing $I_N$ to encourage independence among frames, the temporal coherence of the synthesized frames is indeed compromised. 
Conversely, employing $J_N$ can significantly diminish the video dynamics, leading to a quasi-static video. 
This controlled experiment clearly demonstrates how the temporal attention layer impacts the dynamics of the video synthesis model.

Finally, we investigate the effect of the interplay between attention and resolution on the content dynamics of videos. As also shown in \cref{fig:ablation_per_res_TA}, the replacement at the higher resolutions of 64 and 32 has a more evident effect than at lower resolutions. Applying the changes jointly at both resolutions, 64 \& 32, further amplifies the effect. In contrast, the videos are much less responsive to the attention replacement at resolution 8. Likely, the low resolution features encode high-level semantics, while with higher resolution features there is more capacity for representing varying local details in the scene; such details are necessary for reflecting coherent change over frames.

Based on these controlled experiments, we can conclude that manipulating temporal attention allows us to alter the video dynamics, \ie, making the visual process either more static or more dynamic. 
In particular, adjustments at higher resolutions, \eg 64 \& 32, are more effective.

\vspace{-0.5em}
\subsection{Temporal Attention Regularization (TAR)}
\label{subsec:method-TA-reg}

From the experiments above, we have clearly observed the role of temporal attention layers in determining the dynamics of videos. Naturally, the attention matrices of synthetic videos should be similar to that of real videos. 
Therefore, we propose a simple regularization technique applied on the temporal attention layers for pretrained T2V model. %
Note that, our proposal is directly applied to pretrained T2V models without requiring re-training, and incurs no additional optimization costs during inference.

\begin{figure}[t]
\vspace{-0.5em}
\begin{centering}
\setlength{\tabcolsep}{0.0em}
\renewcommand{\arraystretch}{0}
\par\end{centering}
\begin{centering}
\hfill{}%
 \begin{tabular}{
    m{0.07\linewidth}<{\centering} @{}
    m{0.15\linewidth}<{\centering} @{}
    m{0.15\linewidth}<{\centering} @{}
    m{0.15\linewidth}<{\centering} @{} %
    m{0.15\linewidth}<{\centering} @{}
    m{0.15\linewidth}<{\centering} @{}
    m{0.15\linewidth}<{\centering} @{}
    }
\tabularnewline
& \multicolumn{6}{c}{\begin{tabular}{c}
\myquote{A peony starts to bloom, in the field}
\end{tabular}} 
\tabularnewline
    \multirow{1}{*}
    {\rotatebox{90}{
        \scriptsize None
        \hspace{-0.4\linewidth}
    }} 
    & \animategraphics[autoplay,loop,width=0.98\linewidth,height=0.65\linewidth]{8}{figs/ablation_std/peony_536/emb/frame_}{0}{15}
    & \includegraphics[width=0.98\linewidth,height=0.65\linewidth]{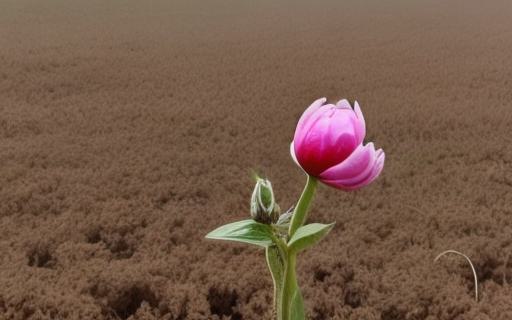}
    & \includegraphics[width=0.98\linewidth,height=0.65\linewidth]{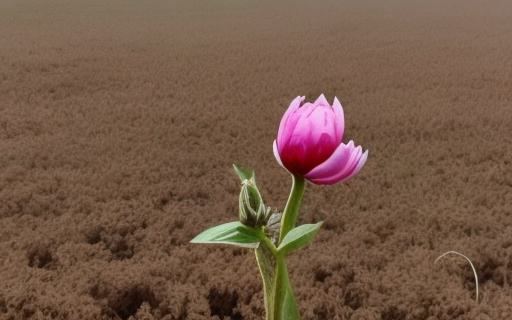} 
    & \includegraphics[width=0.98\linewidth,height=0.65\linewidth]{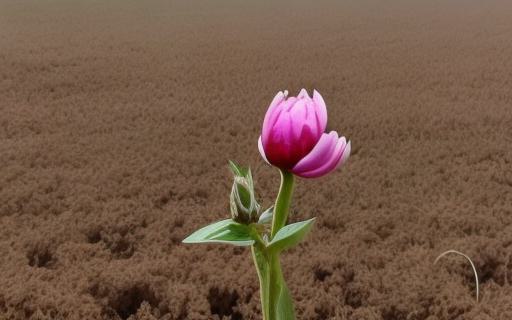} 
    & \includegraphics[width=0.98\linewidth,height=0.65\linewidth]{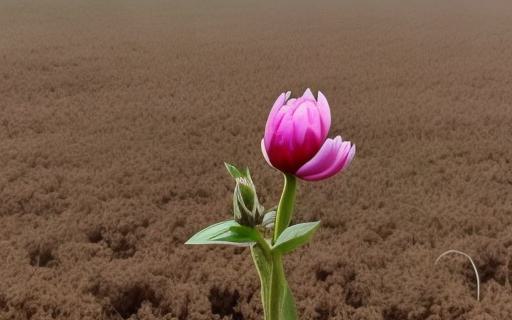} 
    & \includegraphics[width=0.98\linewidth,height=0.65\linewidth]{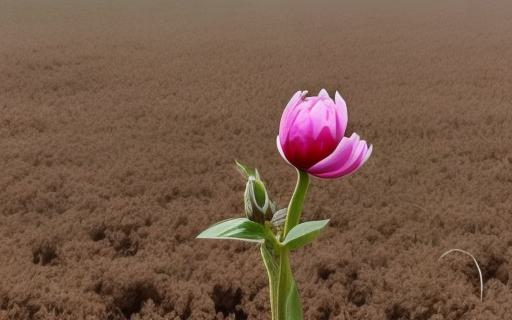}
\tabularnewline
\multirow{1}{*}
    {\rotatebox{90}{
        \begin{tabular}{c}
        \scriptsize $\sigma_{64}=8$
        \end{tabular}
        \hspace{-0.6\linewidth}
    }} 
    & \animategraphics[autoplay,loop,width=0.98\linewidth,height=0.65\linewidth]{8}{figs/ablation_std/peony_536/64D8/frame_}{0}{15}
    & \includegraphics[width=0.98\linewidth,height=0.65\linewidth]{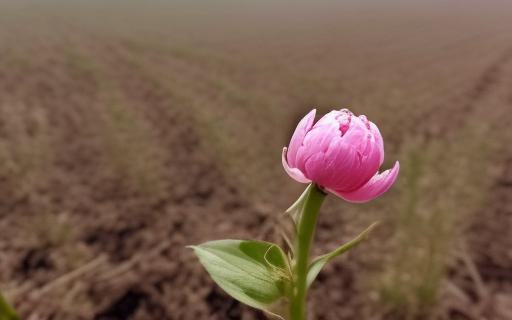}
    & \includegraphics[width=0.98\linewidth,height=0.65\linewidth]{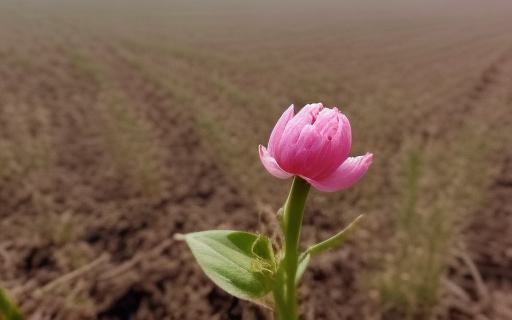} 
    & \includegraphics[width=0.98\linewidth,height=0.65\linewidth]{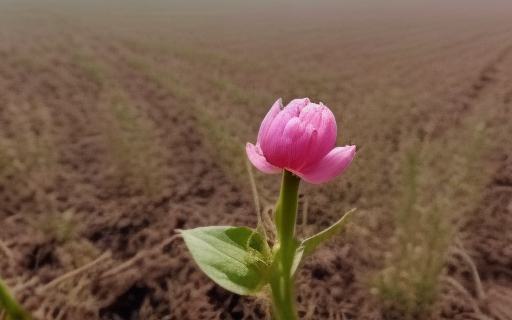} 
    & \includegraphics[width=0.98\linewidth,height=0.65\linewidth]{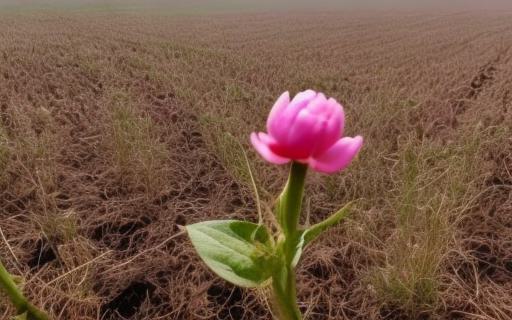} 
    & \includegraphics[width=0.98\linewidth,height=0.65\linewidth]{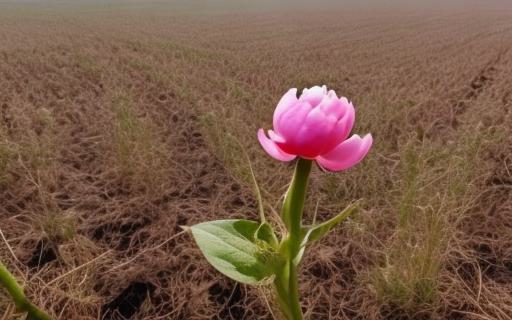}
\tabularnewline
\multirow{1}{*}
    {\rotatebox{90}{
        \begin{tabular}{c}
        \scriptsize $\sigma_{64}=4$
        \end{tabular}
         \hspace{-0.6\linewidth}
    }} 
    & \animategraphics[autoplay,loop,width=0.98\linewidth,height=0.65\linewidth]{8}{figs/ablation_std/peony_536/64D4/frame_}{0}{15}
    & \includegraphics[width=0.98\linewidth,height=0.65\linewidth]{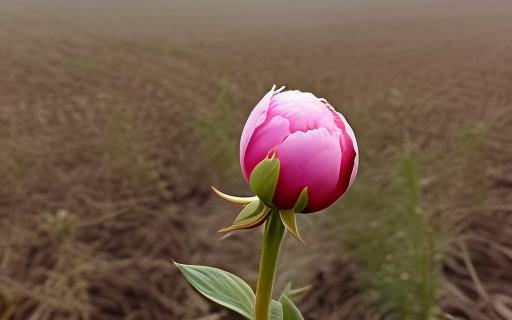}
    & \includegraphics[width=0.98\linewidth,height=0.65\linewidth]{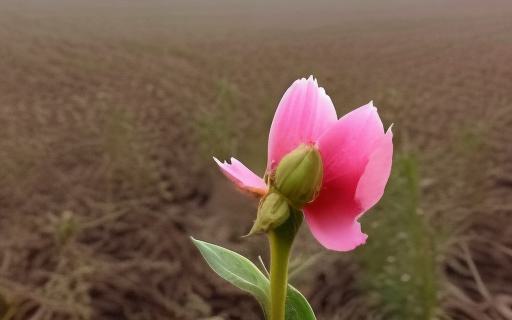} 
    & \includegraphics[width=0.98\linewidth,height=0.65\linewidth]{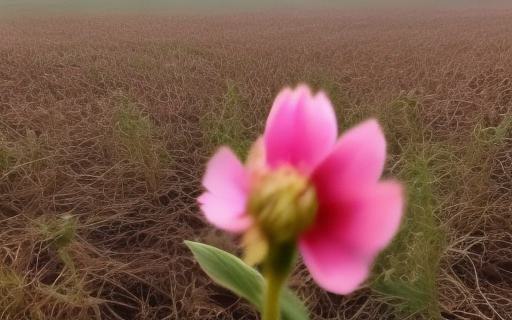} 
    & \includegraphics[width=0.98\linewidth,height=0.65\linewidth]{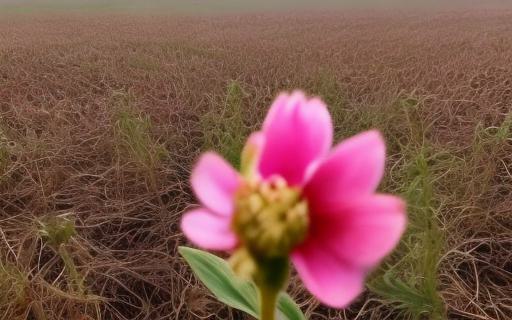} 
    & \includegraphics[width=0.98\linewidth,height=0.65\linewidth]{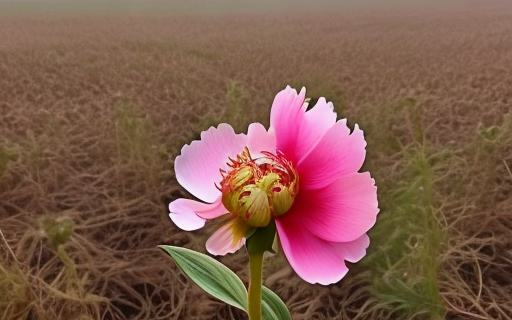}
\tabularnewline
\multirow{1}{*}
    {\rotatebox{90}{
        \begin{tabular}{c}
        \scriptsize $\sigma_{64}=1$
        \end{tabular}
         \hspace{-0.6\linewidth}
    }} 
    & \animategraphics[autoplay,loop,width=0.98\linewidth,height=0.65\linewidth]{8}{figs/ablation_std/peony_536/64D1/frame_}{0}{15}
    & \includegraphics[width=0.98\linewidth,height=0.65\linewidth]{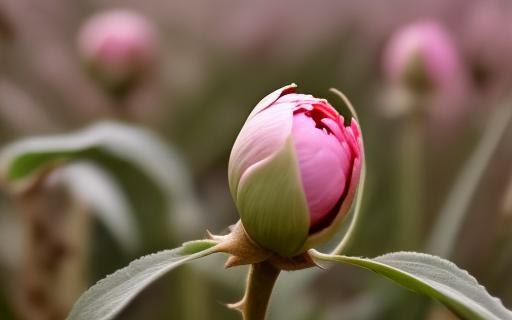}
    & \includegraphics[width=0.98\linewidth,height=0.65\linewidth]{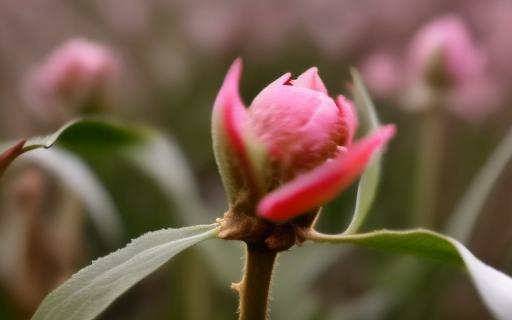} 
    & \includegraphics[width=0.98\linewidth,height=0.65\linewidth]{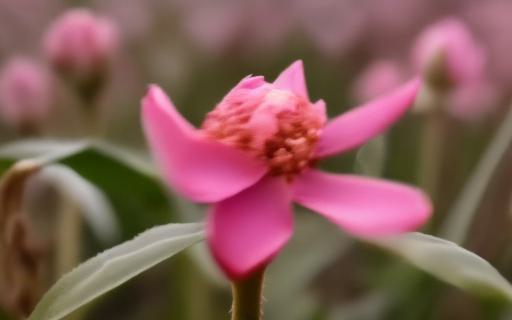} 
    & \includegraphics[width=0.98\linewidth,height=0.65\linewidth]{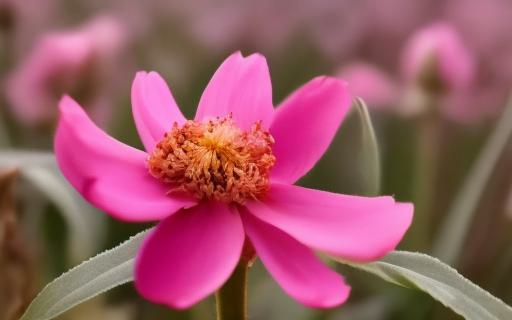} 
    & \includegraphics[width=0.98\linewidth,height=0.65\linewidth]{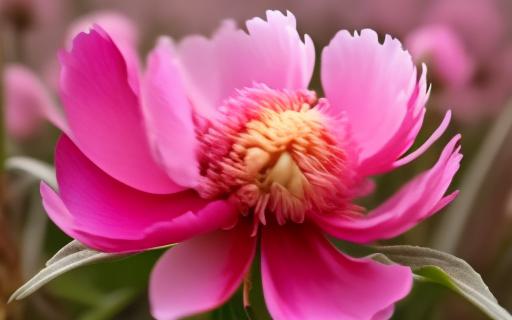}
\tabularnewline
\multirow{1}{*}
    {\rotatebox{90}{
        \begin{tabular}{c}
        \scriptsize $\sigma_{64}=1$ \\  
        \scriptsize $\sigma_{32}=8$ 
        \end{tabular}
        \hspace{-0.6\linewidth}
    }} 
    & \animategraphics[autoplay,loop,width=0.98\linewidth,height=0.65\linewidth]{8}{figs/ablation_std/peony_536/64D1_32D8/frame_}{0}{15}
    & \includegraphics[width=0.98\linewidth,height=0.65\linewidth]{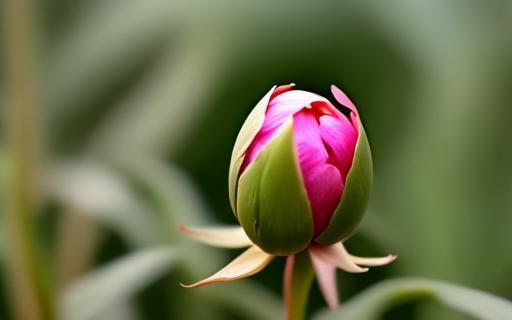}
    & \includegraphics[width=0.98\linewidth,height=0.65\linewidth]{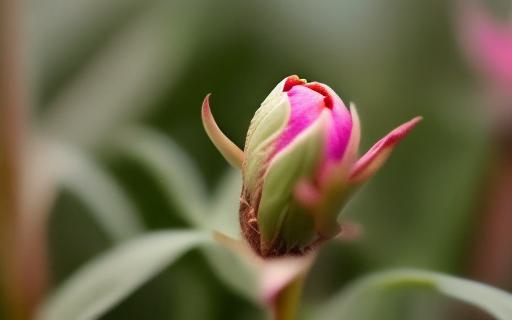} 
    & \includegraphics[width=0.98\linewidth,height=0.65\linewidth]{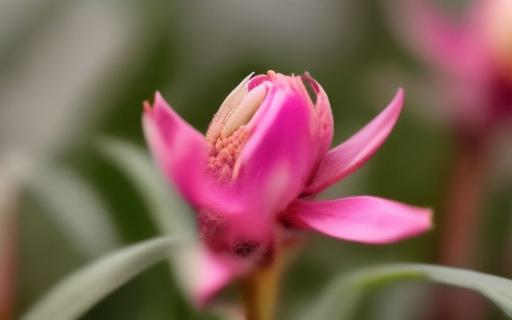} 
    & \includegraphics[width=0.98\linewidth,height=0.65\linewidth]{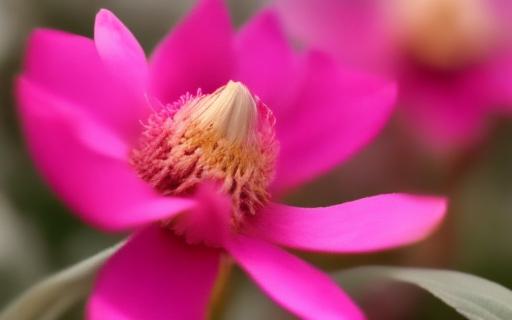} 
    & \includegraphics[width=0.98\linewidth,height=0.65\linewidth]{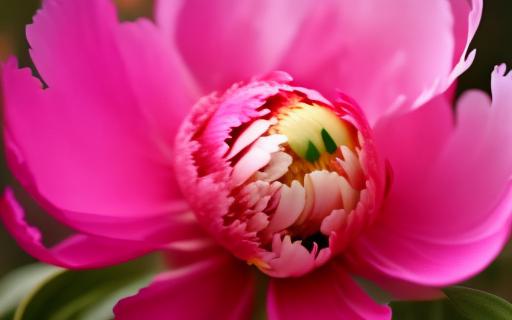}
\tabularnewline
\multirow{1}{*}
    {\rotatebox{90}{
        \begin{tabular}{c}
        \scriptsize $\sigma_{64}=1$ \\  
        \scriptsize $\sigma_{32}=1$ 
        \end{tabular}
        \hspace{-0.6\linewidth}
    }} 
    & \animategraphics[autoplay,loop,width=0.98\linewidth,height=0.65\linewidth]{8}{figs/ablation_std/peony_536/64D1_32D1/frame_}{0}{15}
    & \includegraphics[width=0.98\linewidth,height=0.65\linewidth]{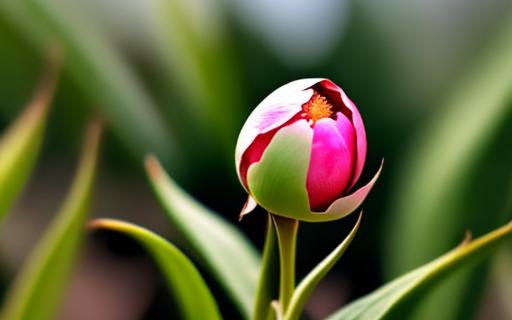}
    & \includegraphics[width=0.98\linewidth,height=0.65\linewidth]{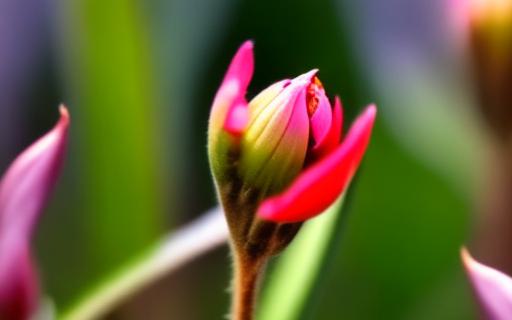} 
    & \includegraphics[width=0.98\linewidth,height=0.65\linewidth]{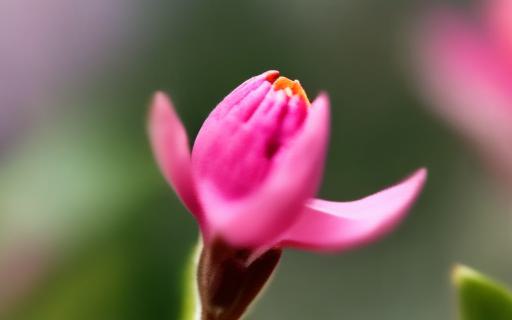} 
    & \includegraphics[width=0.98\linewidth,height=0.65\linewidth]{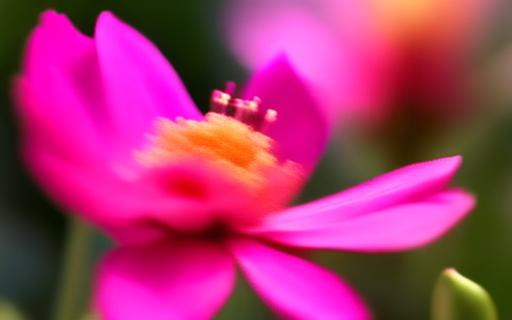} 
    & \includegraphics[width=0.98\linewidth,height=0.65\linewidth]{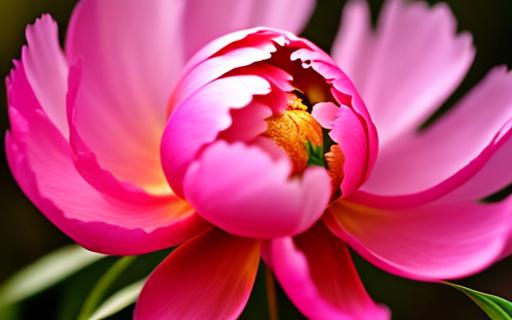}
\tabularnewline
\end{tabular}
\hfill{}
\par\end{centering}
\vspace{-0.5em}
\caption{Ablation of attention regularization matrix $\Delta A$. 
Smaller $\sigma$ induces a stronger regularization effect, leading to increasing temporal dynamics. When applying regularization at both 64 \& 32, the video becomes more dynamic, \ie, the peony is fully bloomed. Yet, excessive regularization, \ie, $\sigma_{64}=\sigma_{32}=1$, can leave the impression of temporal incoherency. 
Contains GIFs, best viewed in \emph{Acrobat Reader}.  
}
\label{fig:ablate_reg_matrix}
\vspace{-0.3em}
\end{figure}

As illustrated in \cref{fig:compare_TA_real_fake}, the attention correlation of the real video resembles a band-matrix-like structure, with high correlation between neighboring frames and lower correlation the larger the frame offset. To approximate such a structure, we design a symmetric Toeplitz matrix as the regularization matrix $\Delta A$, with its values along the off-diagonal direction following the Gaussian distribution:
\begin{align} \label{eq:delta-Gaussian}
\Delta A_{i,j} =  e^{-\frac{1}{2}(\frac{j-i}{\sigma})^2}, 
\end{align}
where $i,j \in \{1,...,N\}$ represent the entry index of the attention regularization map, and $\sigma$ is the standard deviation of the normal distribution. As indicated in \cref{fig:ablate_reg_matrix}, the standard deviation $\sigma$ can control the regularization strength, \ie larger $\sigma$ leading to less visual variations along the temporal dimension. 
This regularization matrix is then added to the original attention matrix in (\ref{eq:attention}), \ie
\begin{align} \label{eq:delta-regularization}
A' \leftarrow Softmax\left(\phi(Q,K) + \max[\phi(Q,K)] \cdot \Delta A\right).
\end{align}
To balance both terms, we additionally introduce $\max[\phi(Q,K)]$, which weights $\Delta A$ based on maximum in the attention matrix $\phi(Q,K)$. As illustrated in \cref{fig:overview}, the regularized attention map $A'$ will be inserted back for further processing.  

With the combination of both VSP and TAR, our {\ourdm} can effectively provide temporal nursing for video generation, enabling the synthesis of long videos with appealing visual evolution using pretrained T2V models, while also introducing no optimization overhead. 
We find temporal attention analysis to be a powerful tool for understanding the temporal modeling of video diffusion models and leverage it to analyze other T2V models in the next section. We establish valuable connections to their architecture designs, and provide guidance for the future training of T2V models for long video generation.

\vspace{-0.5em}
\section{Experiments} \label{sec:experiments}
\vspace{-0.5em}

\begin{figure}[t]
\vspace{-0.5em}
\begin{centering}
\setlength{\tabcolsep}{0.0em}
\renewcommand{\arraystretch}{0}
\par\end{centering}
\begin{centering}
\hfill{}%
 \begin{tabular}{
    m{0.03\linewidth}<{\centering} @{}
    m{0.16\linewidth}<{\centering} @{}
    m{0.16\linewidth}<{\centering} @{}
    m{0.16\linewidth}<{\centering} @{} %
    m{0.16\linewidth}<{\centering} @{}
    m{0.16\linewidth}<{\centering} @{}
    m{0.16\linewidth}<{\centering} @{}
    }
\tabularnewline
& \multicolumn{6}{c}{\begin{tabular}{c}
\myquote{A Ferrari driving on the road, starts to snow}
\end{tabular}} 
\tabularnewline
    \multirow{1}{*}
    {\rotatebox{90}{
        \scriptsize ModelScope
        \hspace{-1.5\linewidth}
    }} 
    & \animategraphics[autoplay,loop,width=0.98\linewidth,height=0.65\linewidth]{8}{figs/compare_16frames/car_snow/modelscope/frame_}{0}{15}
    & \includegraphics[width=0.98\linewidth,height=0.65\linewidth]{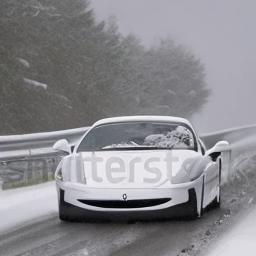}
    & \includegraphics[width=0.98\linewidth,height=0.65\linewidth]{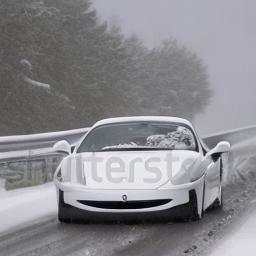} 
    & \includegraphics[width=0.98\linewidth,height=0.65\linewidth]{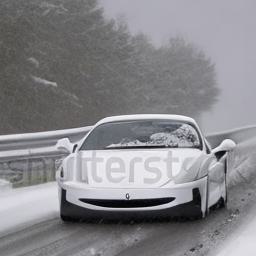} 
    & \includegraphics[width=0.98\linewidth,height=0.65\linewidth]{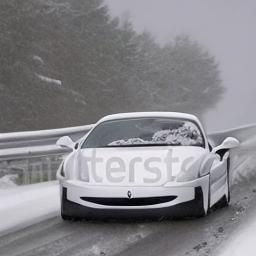} 
    & \includegraphics[width=0.98\linewidth,height=0.65\linewidth]{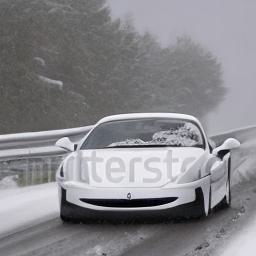}
\tabularnewline
\multirow{1}{*}
    {\rotatebox{90}{
        \begin{tabular}{c}
        \scriptsize LaVie
        \end{tabular}
        \hspace{-1.2\linewidth}
    }} 
    & \animategraphics[autoplay,loop,width=0.98\linewidth,height=0.65\linewidth]{8}{figs/compare_16frames/car_snow/LaVie/frame_}{0}{15}
    & \includegraphics[width=0.98\linewidth,height=0.65\linewidth]{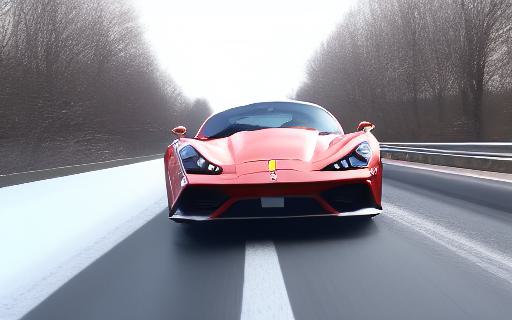}
    & \includegraphics[width=0.98\linewidth,height=0.65\linewidth]{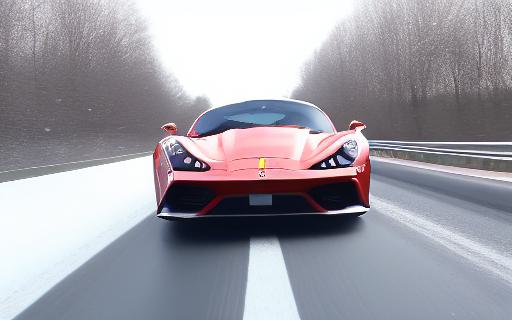} 
    & \includegraphics[width=0.98\linewidth,height=0.65\linewidth]{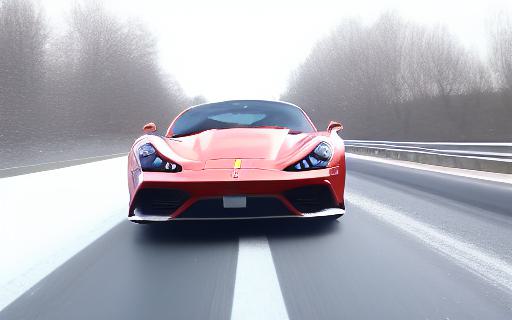} 
    & \includegraphics[width=0.98\linewidth,height=0.65\linewidth]{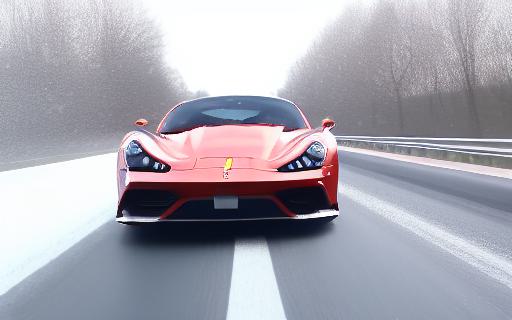} 
    & \includegraphics[width=0.98\linewidth,height=0.65\linewidth]{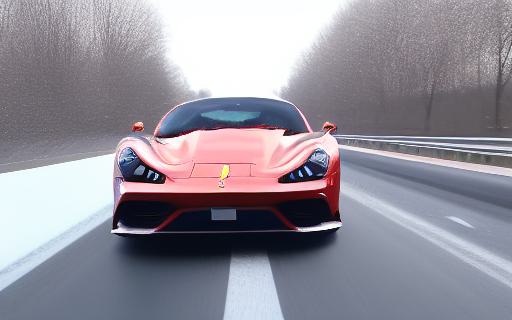}
\tabularnewline
\multirow{1}{*}
    {\rotatebox{90}{
        \begin{tabular}{c}
        \scriptsize AnimateDiff
        \end{tabular}
        \hspace{-2\linewidth}
    }} 
    & \animategraphics[autoplay,loop,width=0.98\linewidth,height=0.65\linewidth]{8}{figs/compare_16frames/car_snow/AnimateDiff/frame_}{0}{15}
    & \includegraphics[width=0.98\linewidth,height=0.65\linewidth]{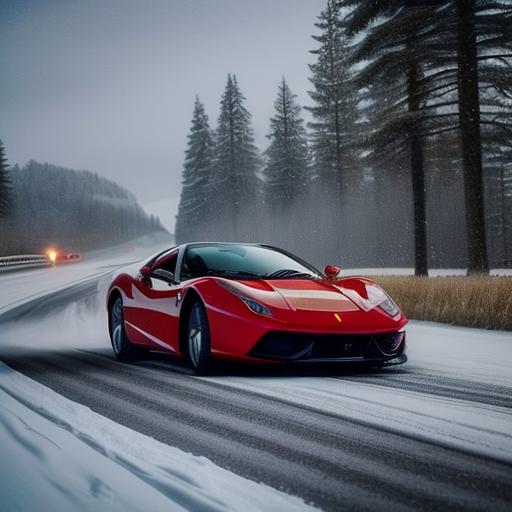}
    & \includegraphics[width=0.98\linewidth,height=0.65\linewidth]{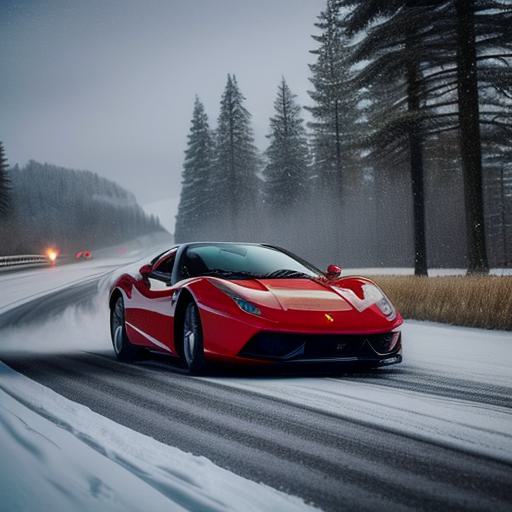} 
    & \includegraphics[width=0.98\linewidth,height=0.65\linewidth]{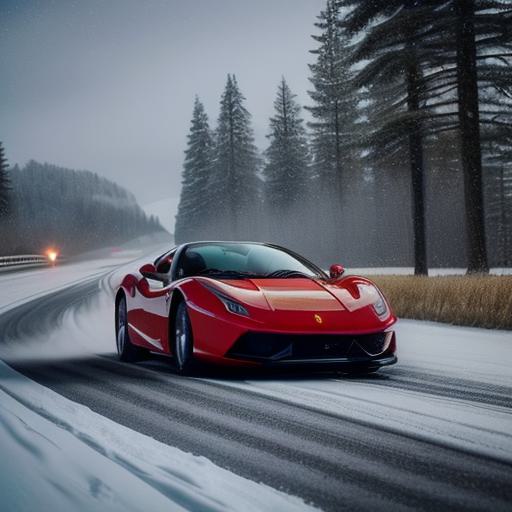} 
    & \includegraphics[width=0.98\linewidth,height=0.65\linewidth]{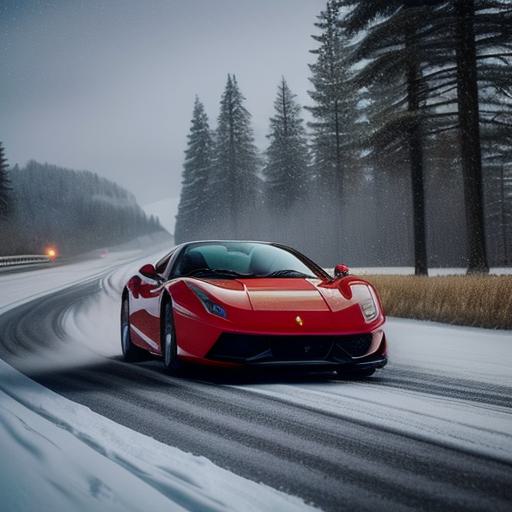} 
    & \includegraphics[width=0.98\linewidth,height=0.65\linewidth]{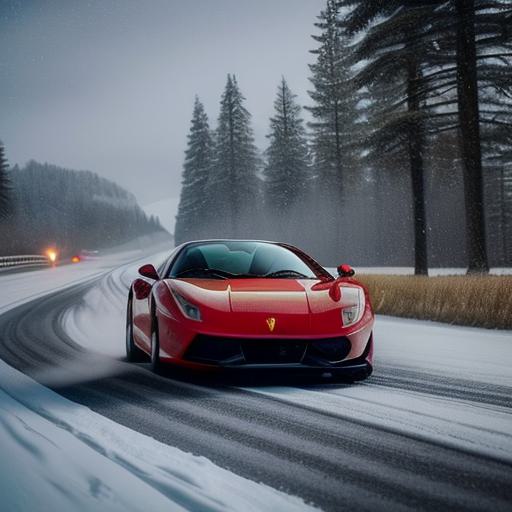}
\tabularnewline
\multirow{1}{*}
    {\rotatebox{90}{
        \begin{tabular}{c}
        \scriptsize V.Crafter2
        \end{tabular}
         \hspace{-1.8\linewidth}
    }} 
    & \animategraphics[autoplay,loop,width=0.98\linewidth,height=0.65\linewidth]{8}{figs/compare_16frames/car_snow/videocrafter/frame_}{0}{15}
    & \includegraphics[width=0.98\linewidth,height=0.65\linewidth]{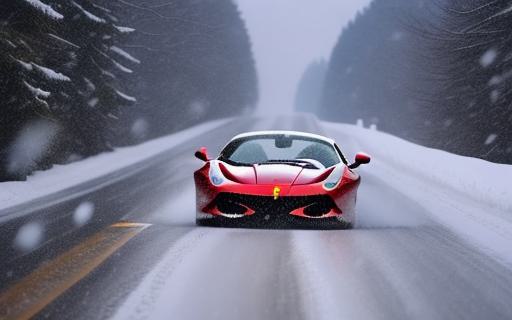}
    & \includegraphics[width=0.98\linewidth,height=0.65\linewidth]{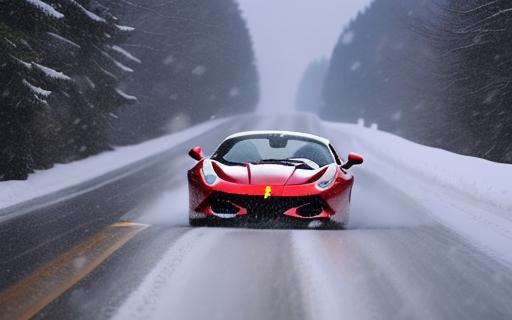} 
    & \includegraphics[width=0.98\linewidth,height=0.65\linewidth]{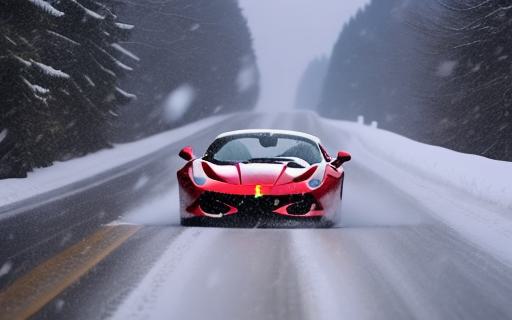} 
    & \includegraphics[width=0.98\linewidth,height=0.65\linewidth]{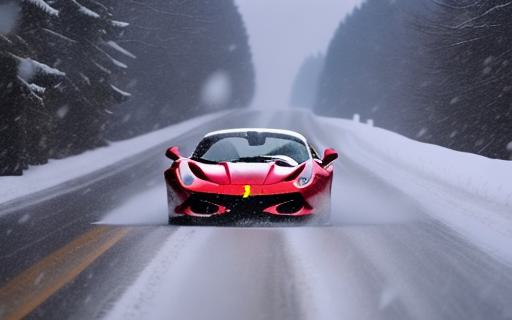} 
    & \includegraphics[width=0.98\linewidth,height=0.65\linewidth]{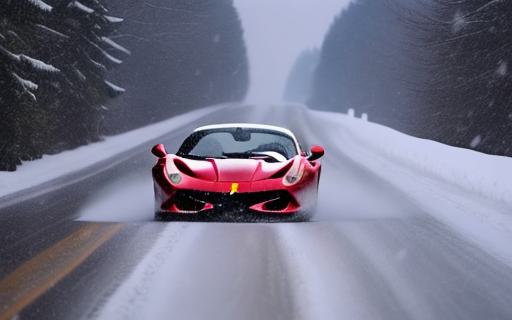}
\tabularnewline
\multirow{1}{*}
    {\rotatebox{90}{
        \begin{tabular}{c}
        \scriptsize \textbf{Ours}
        \end{tabular}
         \hspace{-1.0\linewidth}
    }} 
    & \animategraphics[autoplay,loop,width=0.98\linewidth,height=0.65\linewidth]{8}{figs/compare_16frames/car_snow/ours/frame_}{0}{15}
    & \includegraphics[width=0.98\linewidth,height=0.65\linewidth]{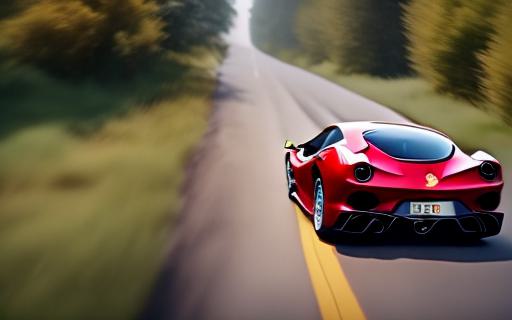}
    & \includegraphics[width=0.98\linewidth,height=0.65\linewidth]{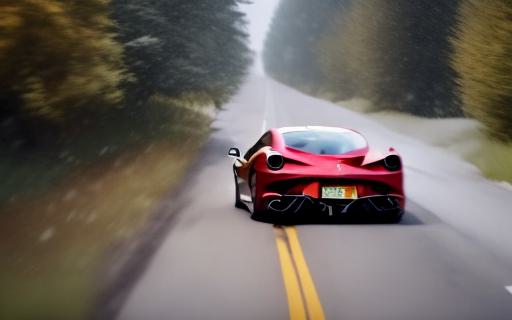} 
    & \includegraphics[width=0.98\linewidth,height=0.65\linewidth]{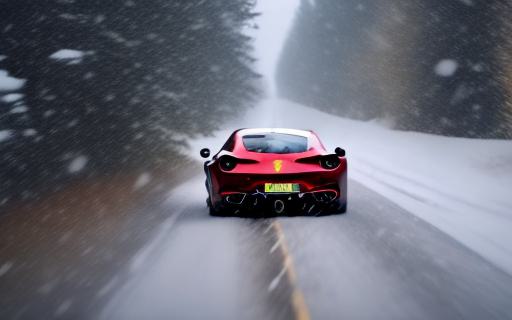} 
    & \includegraphics[width=0.98\linewidth,height=0.65\linewidth]{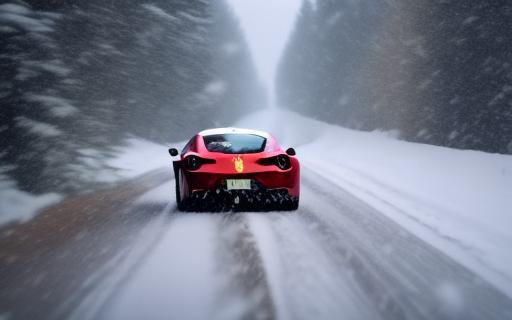} 
    & \includegraphics[width=0.98\linewidth,height=0.65\linewidth]{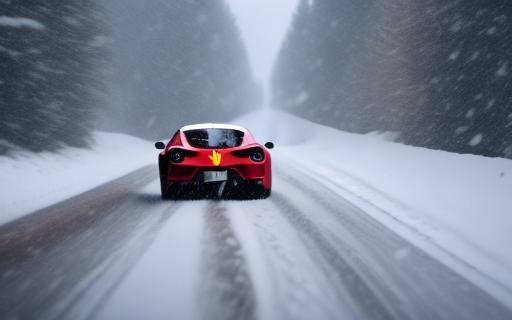}
\tabularnewline
\end{tabular}
\hfill{}
\par\end{centering}
\vspace{-0.5em}
\caption{Comparison with other T2V models on 16 frames generation. Our {\ourdm} can synthesize desired visual development from a clear day to snowy scene, while the others tend to generate the final visual state, \ie, snowy day.
}
\label{fig:compare_t2v_16}
\end{figure}

\begin{figure}[!ht]
\vspace{-0.5em}
\begin{centering}
\setlength{\tabcolsep}{0.0em}
\renewcommand{\arraystretch}{0}
\par\end{centering}
\begin{centering}
\hfill{}%
 \begin{tabular}{
    m{0.03\linewidth}<{\centering} @{}
    m{0.16\linewidth}<{\centering} @{}
    m{0.16\linewidth}<{\centering} @{}
    m{0.16\linewidth}<{\centering} @{} %
    m{0.16\linewidth}<{\centering} @{}
    m{0.16\linewidth}<{\centering} @{}
    m{0.16\linewidth}<{\centering} @{}
    }

\tabularnewline
& \multicolumn{6}{c}{\begin{tabular}{c}
 \myquote{A landscape transitioning from  winter to spring}
\end{tabular}} 
\tabularnewline
    \multirow{1}{*}
    {\rotatebox{90}{
        \scriptsize ModelScope
        \hspace{-1.5\linewidth}
    }} 
    & \animategraphics[autoplay,loop,width=0.98\linewidth,height=0.65\linewidth]{8}{figs/compare_32frames/modelscope/371/frame_}{0}{31}
    & \includegraphics[width=0.98\linewidth,height=0.65\linewidth]{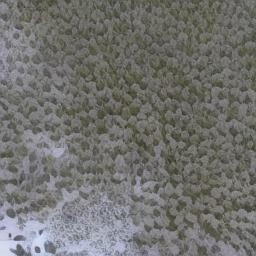}
    & \includegraphics[width=0.98\linewidth,height=0.65\linewidth]{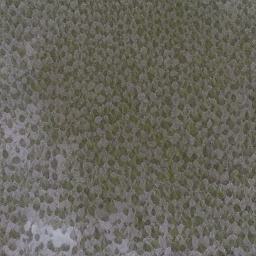} 
    & \includegraphics[width=0.98\linewidth,height=0.65\linewidth]{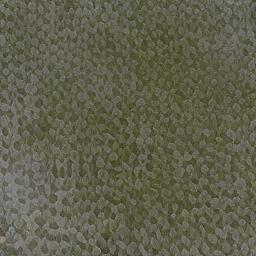} 
    & \includegraphics[width=0.98\linewidth,height=0.65\linewidth]{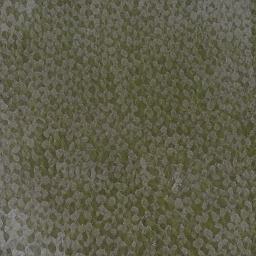} 
    & \includegraphics[width=0.98\linewidth,height=0.65\linewidth]{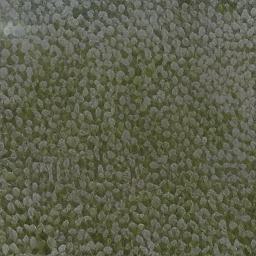}
\tabularnewline
\multirow{1}{*}
    {\rotatebox{90}{
        \begin{tabular}{c}
        \scriptsize LaVie
        \end{tabular}
        \hspace{-1.2\linewidth}
    }} 
    & \animategraphics[autoplay,loop,width=0.98\linewidth,height=0.65\linewidth]{8}{figs/compare_32frames/LaVie/371/frame_}{0}{31}
    & \includegraphics[width=0.98\linewidth,height=0.65\linewidth]{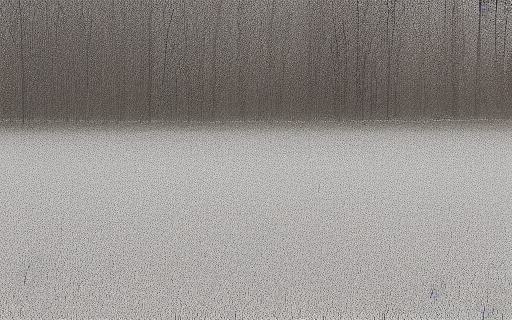}
    & \includegraphics[width=0.98\linewidth,height=0.65\linewidth]{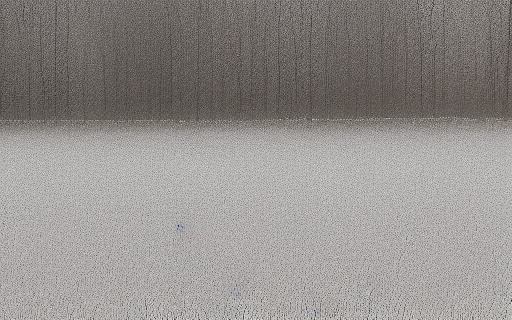} 
    & \includegraphics[width=0.98\linewidth,height=0.65\linewidth]{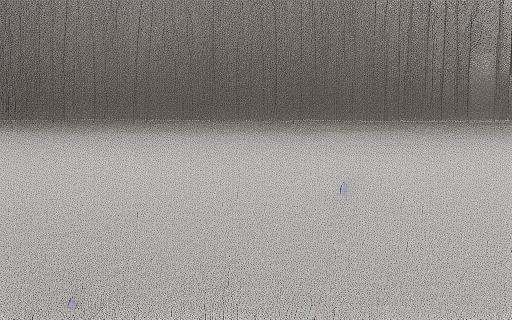} 
    & \includegraphics[width=0.98\linewidth,height=0.65\linewidth]{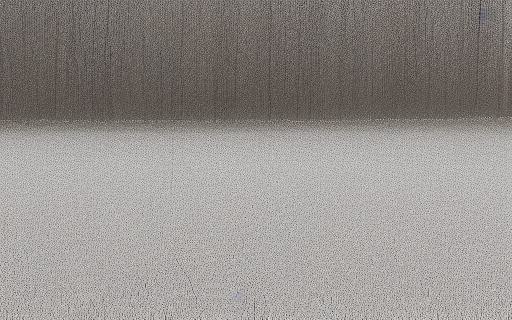} 
    & \includegraphics[width=0.98\linewidth,height=0.65\linewidth]{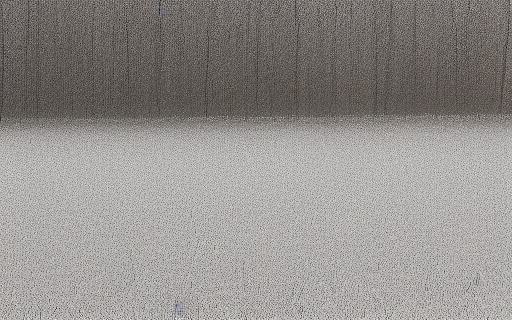}
\tabularnewline
\multirow{1}{*}
    {\rotatebox{90}{
        \begin{tabular}{c}
        \scriptsize AnimateDiff
        \end{tabular}
        \hspace{-2\linewidth}
    }} 
    & \animategraphics[autoplay,loop,width=0.98\linewidth,height=0.65\linewidth]{8}{figs/compare_32frames/animatediff/67/frame_}{0}{31}
    & \includegraphics[width=0.98\linewidth,height=0.65\linewidth]{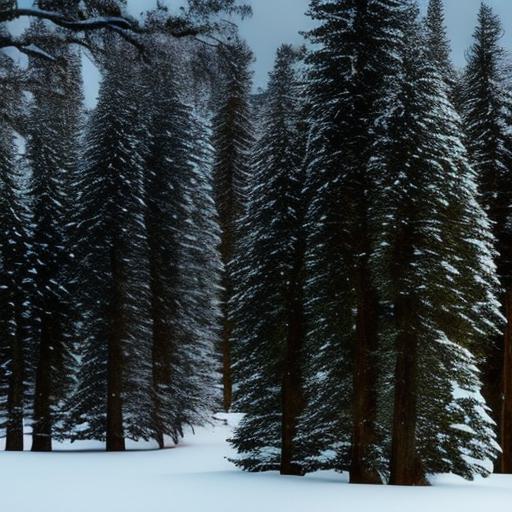}
    & \includegraphics[width=0.98\linewidth,height=0.65\linewidth]{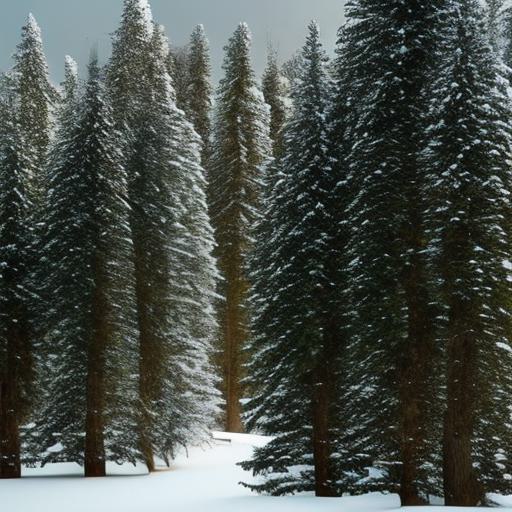} 
    & \includegraphics[width=0.98\linewidth,height=0.65\linewidth]{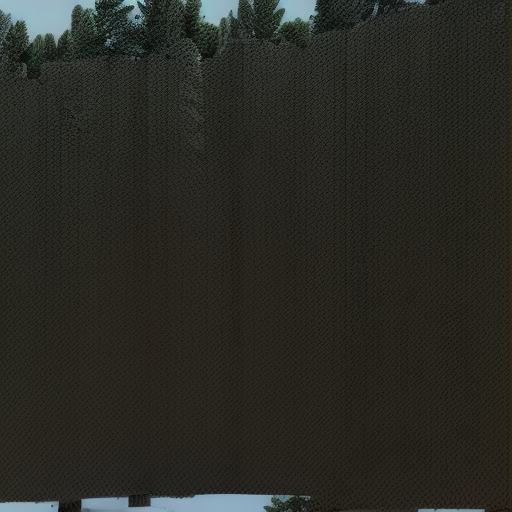} 
    & \includegraphics[width=0.98\linewidth,height=0.65\linewidth]{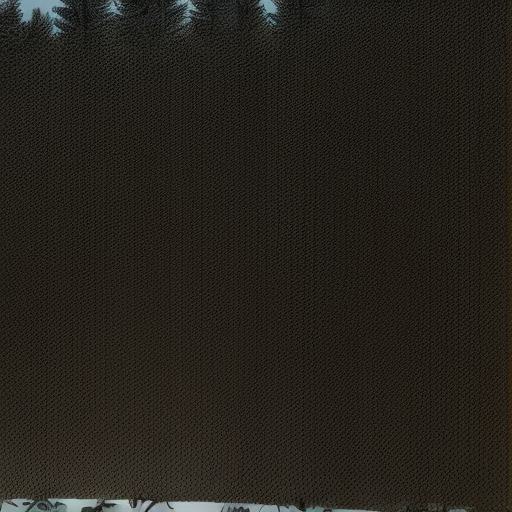} 
    & \includegraphics[width=0.98\linewidth,height=0.65\linewidth]{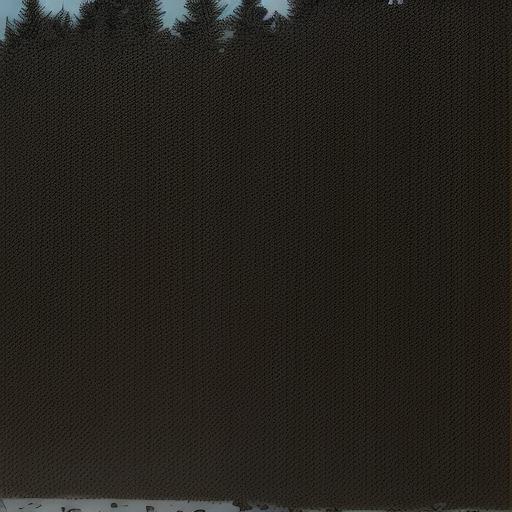}
\tabularnewline
\multirow{1}{*}
    {\rotatebox{90}{
        \begin{tabular}{c}
        \scriptsize V.Crafter2
        \end{tabular}
         \hspace{-1.8\linewidth}
    }} 
    & \animategraphics[autoplay,loop,width=0.98\linewidth,height=0.65\linewidth]{8}{figs/compare_32frames/VideoCrafter/frame_}{0}{31}
    & \includegraphics[width=0.98\linewidth,height=0.65\linewidth]{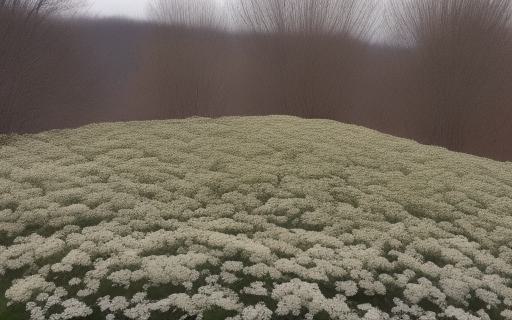}
    & \includegraphics[width=0.98\linewidth,height=0.65\linewidth]{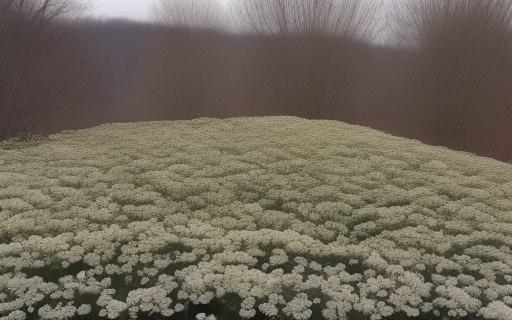} 
    & \includegraphics[width=0.98\linewidth,height=0.65\linewidth]{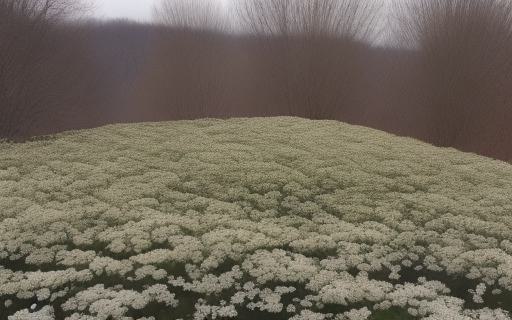} 
    & \includegraphics[width=0.98\linewidth,height=0.65\linewidth]{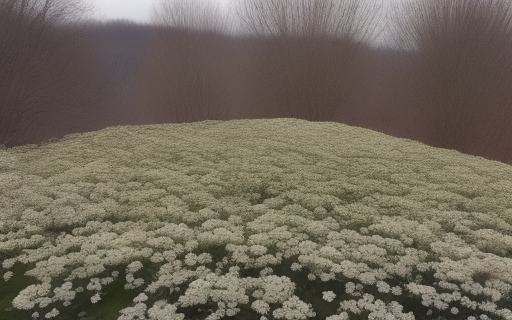} 
    & \includegraphics[width=0.98\linewidth,height=0.65\linewidth]{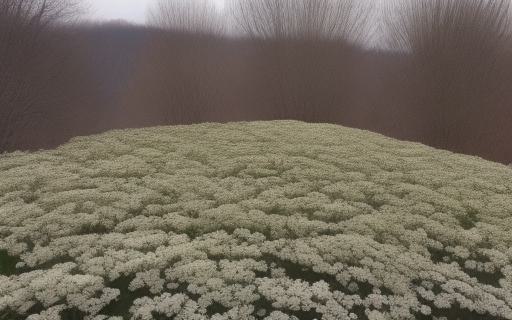}
\tabularnewline
\multirow{1}{*}
    {\rotatebox{90}{
        \begin{tabular}{c}
        \scriptsize \textbf{Ours}
        \end{tabular}
         \hspace{-1.05\linewidth}
    }} 
    & \animategraphics[autoplay,loop,width=0.98\linewidth,height=0.65\linewidth]{8}{figs/compare_32frames/ours/1438/frame_}{0}{31}
    & \includegraphics[width=0.98\linewidth,height=0.65\linewidth]{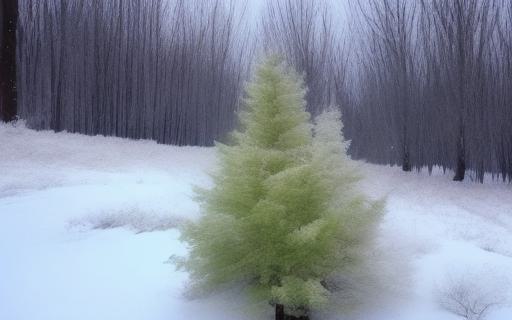}
    & \includegraphics[width=0.98\linewidth,height=0.65\linewidth]{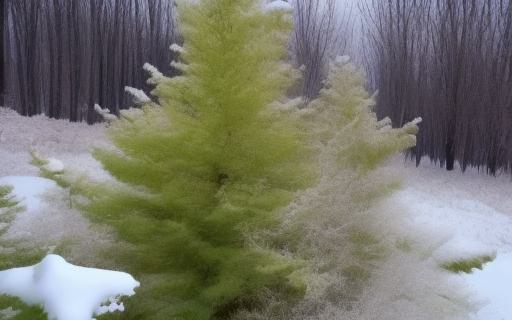} 
    & \includegraphics[width=0.98\linewidth,height=0.65\linewidth]{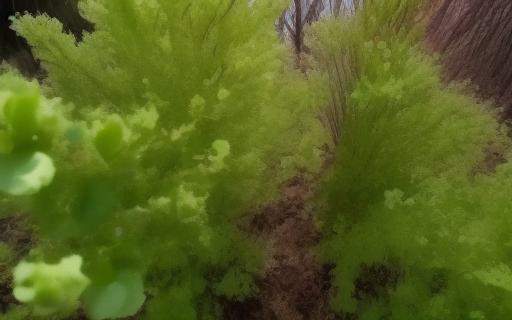} 
    & \includegraphics[width=0.98\linewidth,height=0.65\linewidth]{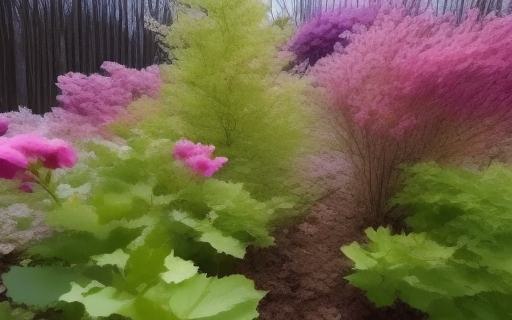} 
    & \includegraphics[width=0.98\linewidth,height=0.65\linewidth]{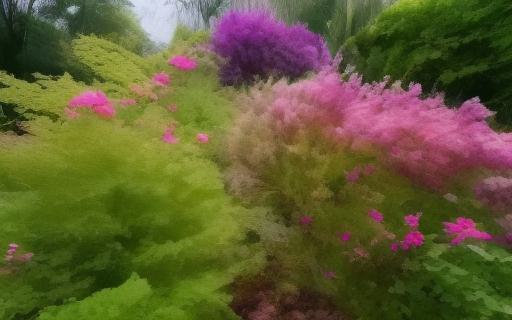}
\tabularnewline
\end{tabular}
\hfill{}
\par\end{centering}
\vspace{-0.5em}
\caption{Comparison with other T2V models on 32 frames generation, which is double the length of the default option. 
Our {\ourdm} can generate long videos with desired dynamics, while the others struggle to synthesize faithful results. 
}
\label{fig:compare_t2v_32}
\end{figure}

\begin{figure*}[t]
\vspace{-0.5em}
\begin{centering}
\setlength{\tabcolsep}{0.0em}
\renewcommand{\arraystretch}{0}
\par\end{centering}
\begin{centering}
\hfill{}%
	\begin{tabular}{
@{\hspace{0.5em}}c
 @{\hspace{1em}}c@{\hspace{0.5em}}c@{\hspace{0.5em}}
 c@{\hspace{0.5em}}c@{\hspace{0.5em}}c@{\hspace{0.5em}}c@{\hspace{0.5em}}c}
	\centering
    & &
    \multicolumn{1}{c}{
    \begin{tabular}{c}{\footnotesize Real}\end{tabular}
    } 
    &
    \multicolumn{1}{c}{
    \begin{tabular}{c}{\footnotesize VideoCrafter2}\end{tabular}
    } 
    & 
    \multicolumn{1}{c}{\begin{tabular}{c}{\footnotesize Ours}\end{tabular}} 
    \tabularnewline
    &
    \includegraphics[height=0.14\linewidth]{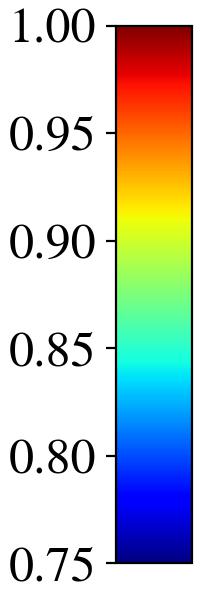}
    &
    \includegraphics[width=0.14\linewidth]{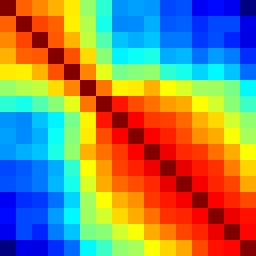}
    &
    \includegraphics[width=0.14\linewidth]{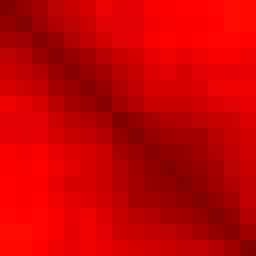}
    & 
    \includegraphics[width=0.14\linewidth]{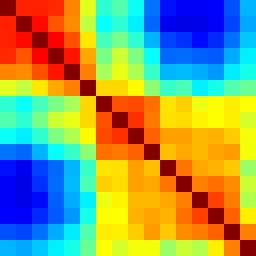}
  \tabularnewline
\end{tabular}
\hfill{}
\par\end{centering}
\vspace{-0.3em}
\caption{ Inter-frame perceptual similarity matrix based on DreamSim~\cite{dreamsim}, where values are normalized across \emph{all} methods. VideoCrafter2 has high similarity across nearly all frames, which is aligned with the visual results lacking variation.
In contrast, our synthesized videos highly resemble the real ones, indicating desired dynamics.
} 
\label{fig:compare_dreamsim}
\vspace{-0.5em}
\end{figure*}

\begin{figure}[t]
\vspace{-0.5em}
\begin{centering}
\setlength{\tabcolsep}{0.0em}
\renewcommand{\arraystretch}{0}
\par\end{centering}
\begin{centering}
\hfill{}%
 \begin{tabular}{
    m{0.03\linewidth}<{\centering} @{}
    m{0.16\linewidth}<{\centering} @{}
    m{0.16\linewidth}<{\centering} @{}
    m{0.16\linewidth}<{\centering} @{} %
    m{0.16\linewidth}<{\centering} @{}
    m{0.16\linewidth}<{\centering} @{}
    m{0.16\linewidth}<{\centering} @{}
    }
\tabularnewline
& \multicolumn{6}{c}{\begin{tabular}{c}
\myquote{Rainbow start to appear after the rainy day}
\end{tabular}} 
\tabularnewline
    \multirow{1}{*}
    {\rotatebox{90}{
        \hspace{-1.2\linewidth}
        \scriptsize Baseline
        \hspace{-1.2\linewidth}
    }} 
    & \animategraphics[autoplay,loop,width=0.98\linewidth,height=0.65\linewidth]{8}{figs/dreamsim_reg/beofre_reg/frame_}{0}{15}
    & \includegraphics[width=0.98\linewidth,height=0.65\linewidth]{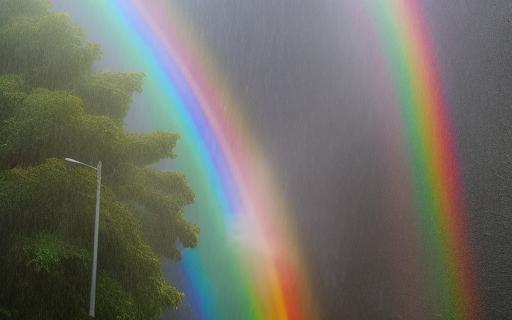}
    & \includegraphics[width=0.98\linewidth,height=0.65\linewidth]{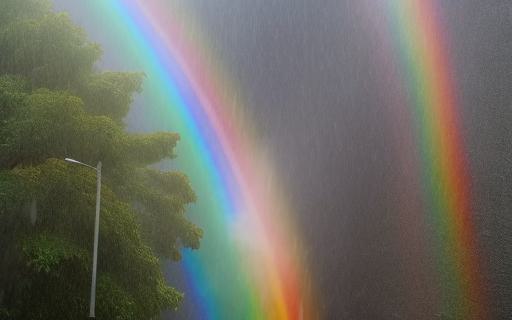} 
    & \includegraphics[width=0.98\linewidth,height=0.65\linewidth]{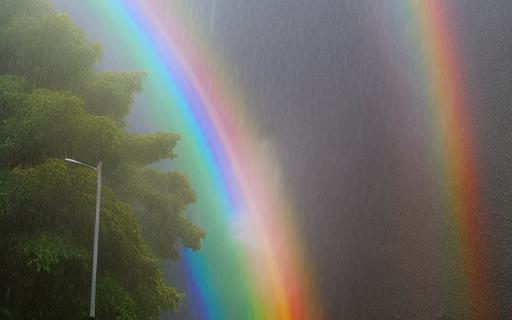} 
    & \includegraphics[width=0.98\linewidth,height=0.65\linewidth]{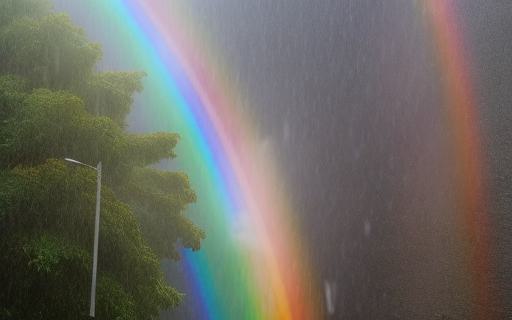} 
    & \includegraphics[width=0.98\linewidth,height=0.65\linewidth]{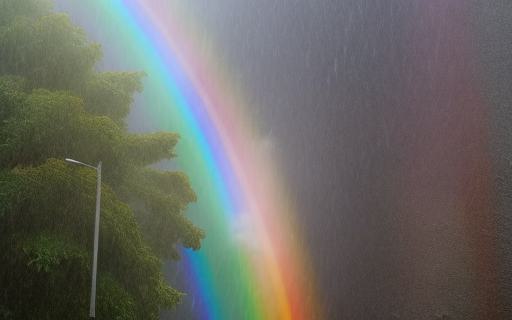}
\tabularnewline
\multirow{1}{*}
    {\rotatebox{90}{
        \begin{tabular}{c}
        \scriptsize + DreamSim
        \end{tabular}
         \hspace{-2.0\linewidth}
    }} 
    & \animategraphics[autoplay,loop,width=0.98\linewidth,height=0.65\linewidth]{8}{figs/dreamsim_reg/after_reg/frame_}{0}{15}
    & \includegraphics[width=0.98\linewidth,height=0.65\linewidth]{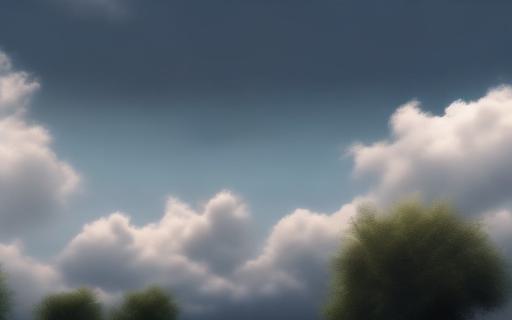}
    & \includegraphics[width=0.98\linewidth,height=0.65\linewidth]{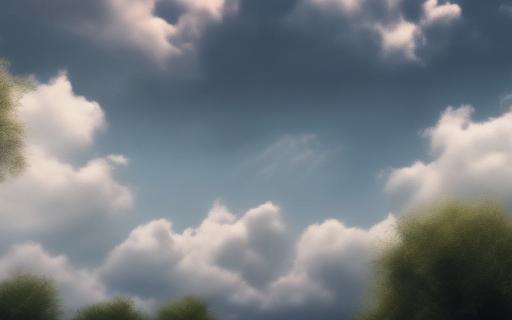} 
    & \includegraphics[width=0.98\linewidth,height=0.65\linewidth]{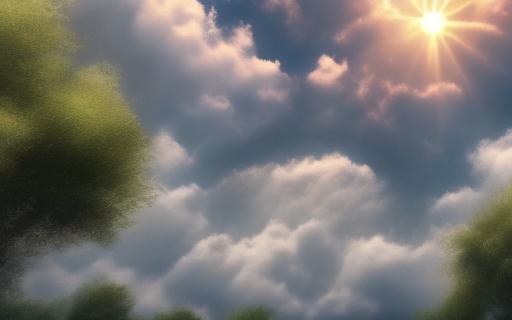} 
    & \includegraphics[width=0.98\linewidth,height=0.65\linewidth]{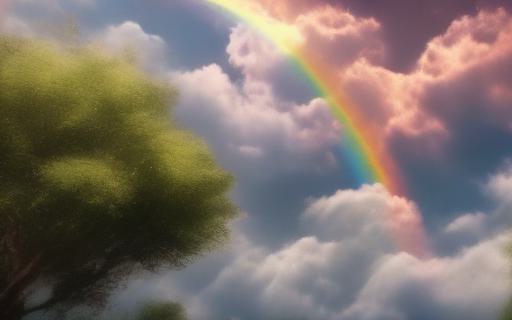} 
    & \includegraphics[width=0.98\linewidth,height=0.65\linewidth]{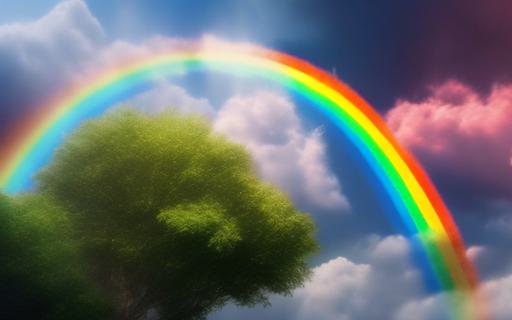}
\tabularnewline
\end{tabular}
\hfill{}
\par\end{centering}
\vspace{-0.7em}
\caption{Regularization with inter-frame DreamSim matrix of one real reference video. 
}
\label{fig:ablation_dreamsim}
\end{figure}

\begin{figure*}[t]
\vspace{-0.5em}
\begin{centering}
\setlength{\tabcolsep}{0.0em}
\renewcommand{\arraystretch}{0}
\par\end{centering}
\begin{centering}
\hfill{}%
	\begin{tabular}{@{\hspace{0.2em}}c@{\hspace{0.7em}}c@{\hspace{0.2em}}
 c@{\hspace{0.7em}}c@{\hspace{0.2em}}c@{\hspace{0.7em}}c@{\hspace{0.2em}}c}
	\centering
    &
    \multicolumn{2}{c}{
    \begin{tabular}{c}{ModelScope\newcite{wang2023modelscope}}\end{tabular}
    } 
    & 
    \multicolumn{2}{c}{
    \begin{tabular}{c}{LaVie\newcite{wang2023lavie}}\end{tabular}
    } 
    & 
    \multicolumn{2}{c}{\begin{tabular}{c}{AnimateDiff\newcite{guo2024animatediff}}\end{tabular}} 
 
    \tabularnewline
    \multirow{1}{*}{ \rotatebox{90}{\hspace{4.0em}  $N=16$ \hspace{-4.0em} }} 
    &
	\includegraphics[width=0.14\linewidth]{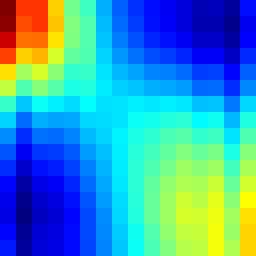}
	      &
	\includegraphics[width=0.14\linewidth]{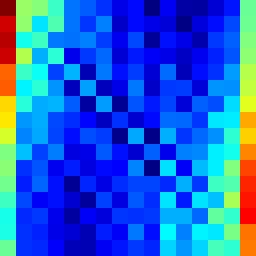}
	    & 
	\includegraphics[width=0.14\linewidth]{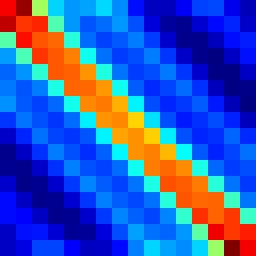}
        & 
    \includegraphics[width=0.14\linewidth]{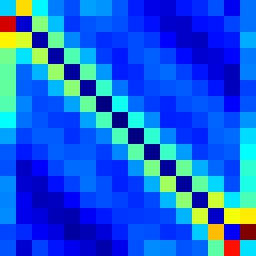}
	      & 
	\includegraphics[width=0.14\linewidth]{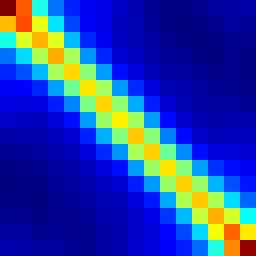}
	    & 
	\includegraphics[width=0.14\linewidth]{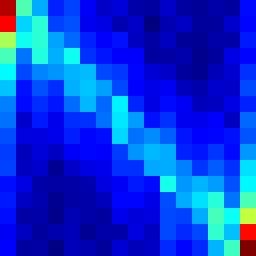}
  \tabularnewline
  \multirow{1}{*}{ \rotatebox{90}{\hspace{4.0em}  $N=32$  \hspace{-4.0em} }} 
    &
	\includegraphics[width=0.14\linewidth]{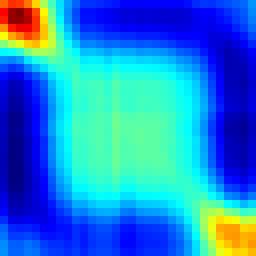}
	      &
	\includegraphics[width=0.14\linewidth]{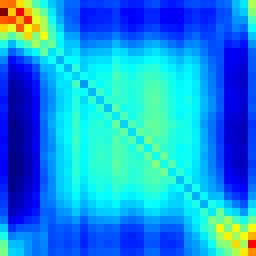}
	    & 
	\includegraphics[width=0.14\linewidth]{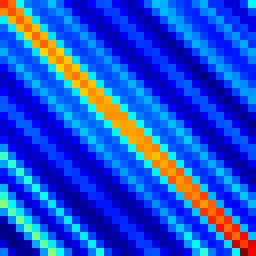}
        & 
    \includegraphics[width=0.14\linewidth]{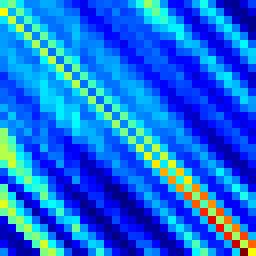}
	      & 
	\includegraphics[width=0.14\linewidth]{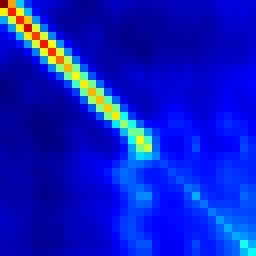}
	    & 
	\includegraphics[width=0.14\linewidth]{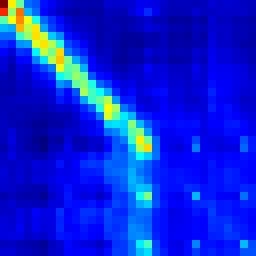}
  \tabularnewline
\end{tabular}
\hfill{}
\par\end{centering}
\vspace{-0.5em}
\caption{Temporal attention visualization of other T2V Models for the default 16 frame and longer 32 frame videos. ModelScope has similar issues to VideoCrafter2 (see \cref{fig:compare_TA_real_fake}), \ie, high correlation spread across many frames, especially for $N=32$. 
LaVie and AnimateDiff incorporate positional encoding of frame indices, thus naturally do not generalize well to long video generation beyond trained 16 frames.
} 
\label{fig:ablation_T2V-TA}
\vspace{-0.5em}
\end{figure*}

\begin{figure}[t]
\vspace{-0.5em}
\begin{centering}
\setlength{\tabcolsep}{0.0em}
\renewcommand{\arraystretch}{0}
\par\end{centering}
\begin{centering}
\hfill{}%
 \begin{tabular}{
    m{0.03\linewidth}<{\centering} @{}
    m{0.16\linewidth}<{\centering} @{}
    m{0.16\linewidth}<{\centering} @{}
    m{0.16\linewidth}<{\centering} @{} %
    m{0.16\linewidth}<{\centering} @{}
    m{0.16\linewidth}<{\centering} @{}
    m{0.16\linewidth}<{\centering} @{}
    }
\tabularnewline
& \multicolumn{6}{c}{\begin{tabular}{c}
\myquote{Spiderman on the beach from morning to evening}
\end{tabular}} 
\tabularnewline
    \multirow{1}{*}
    {\rotatebox{90}{
        \hspace{-1.2\linewidth}
        \scriptsize Baseline
        \hspace{-1.2\linewidth}
    }} 
    & \animategraphics[autoplay,loop,width=0.98\linewidth,height=0.65\linewidth]{8}{figs/ablation_reg_recap/spiderman/baseline/frame_}{0}{47}
    & \includegraphics[width=0.98\linewidth,height=0.65\linewidth]{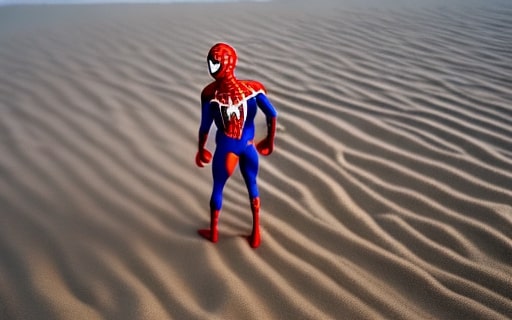}
    & \includegraphics[width=0.98\linewidth,height=0.65\linewidth]{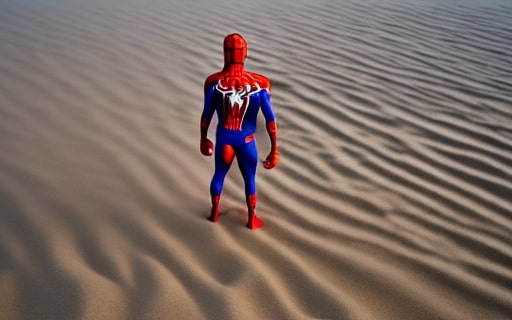} 
    & \includegraphics[width=0.98\linewidth,height=0.65\linewidth]{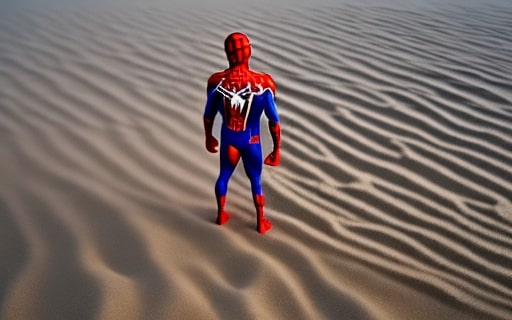} 
    & \includegraphics[width=0.98\linewidth,height=0.65\linewidth]{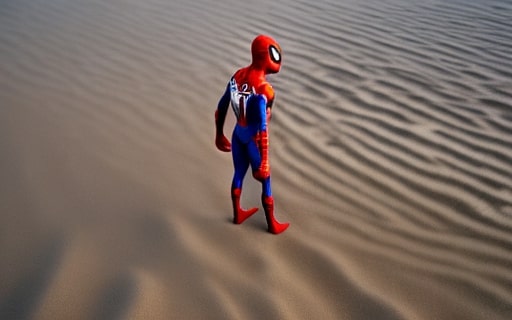} 
    & \includegraphics[width=0.98\linewidth,height=0.65\linewidth]{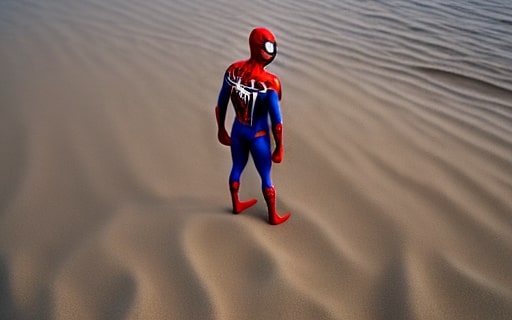}
\tabularnewline
\multirow{1}{*}
    {\rotatebox{90}{
        \hspace{-1.6\linewidth}
        \begin{tabular}{c}
        \scriptsize + TAR  
        \end{tabular}
         \hspace{-1.6\linewidth}
    }} 
    & \animategraphics[autoplay,loop,width=0.98\linewidth,height=0.65\linewidth]{8}{figs/ablation_reg_recap/spiderman/reg_only/frame_}{0}{47}
    & \includegraphics[width=0.98\linewidth,height=0.65\linewidth]{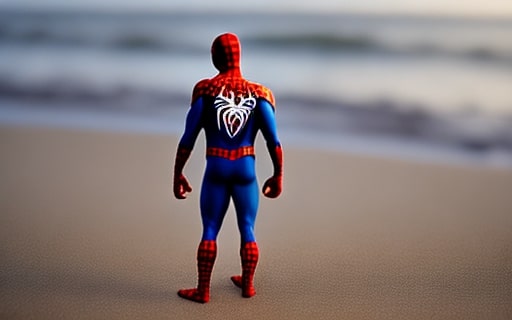}
    & \includegraphics[width=0.98\linewidth,height=0.65\linewidth]{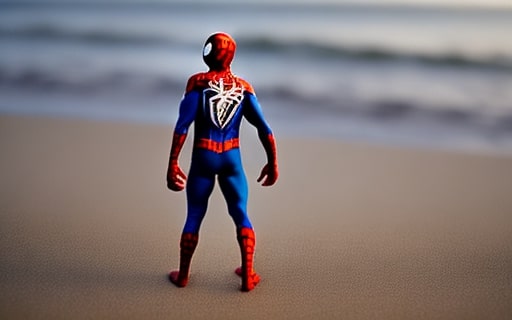} 
    & \includegraphics[width=0.98\linewidth,height=0.65\linewidth]{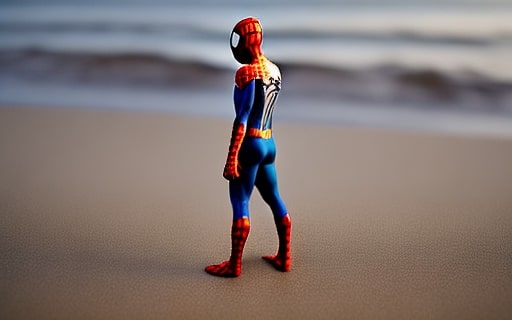} 
    & \includegraphics[width=0.98\linewidth,height=0.65\linewidth]{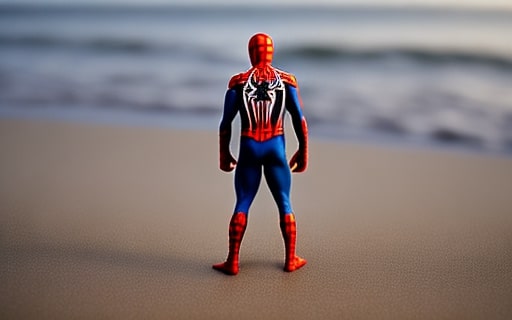} 
    & \includegraphics[width=0.98\linewidth,height=0.65\linewidth]{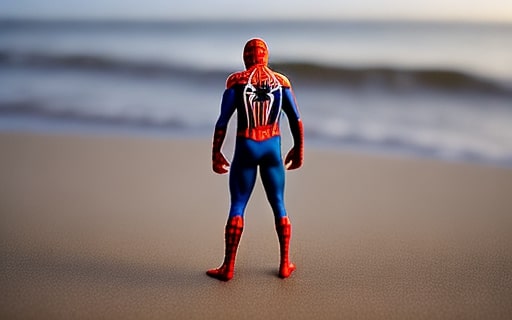}
\tabularnewline
\multirow{1}{*}
    {\rotatebox{90}{
        \hspace{-1.6\linewidth}
        \begin{tabular}{c}
        \scriptsize + VSP
        \end{tabular}
         \hspace{-1.6\linewidth}
    }} 
    & \animategraphics[autoplay,loop,width=0.98\linewidth,height=0.65\linewidth]{8}{figs/ablation_reg_recap/spiderman/recap_only/frame_}{0}{47}
    & \includegraphics[width=0.98\linewidth,height=0.65\linewidth]{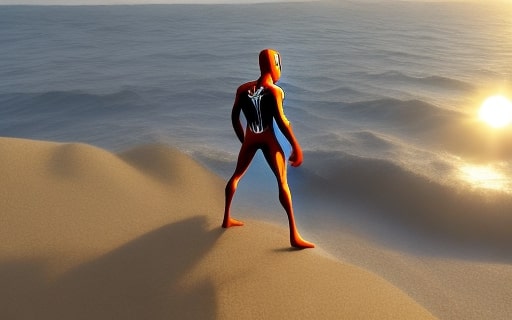}
    & \includegraphics[width=0.98\linewidth,height=0.65\linewidth]{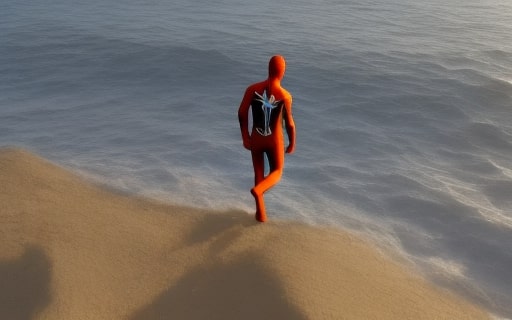} 
    & \includegraphics[width=0.98\linewidth,height=0.65\linewidth]{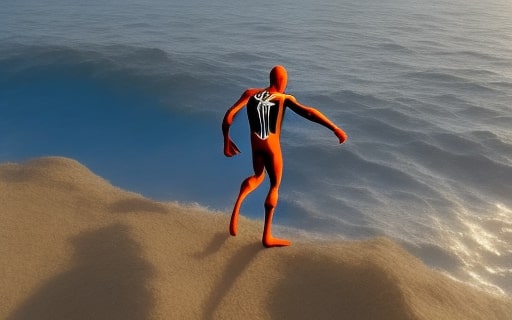} 
    & \includegraphics[width=0.98\linewidth,height=0.65\linewidth]{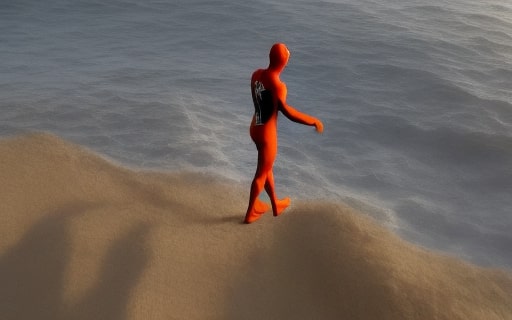} 
    & \includegraphics[width=0.98\linewidth,height=0.65\linewidth]{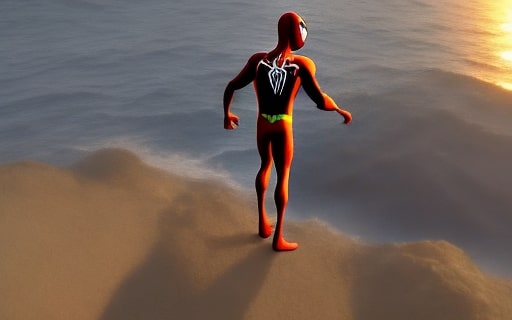}
\tabularnewline
\multirow{1}{*}
    {\rotatebox{90}{
        \hspace{-1.3\linewidth}
        \begin{tabular}{c}
        \scriptsize Ours
        \end{tabular}
         \hspace{-1.3\linewidth}
    }} 
    & \animategraphics[autoplay,loop,width=0.98\linewidth,height=0.65\linewidth]{8}{figs/ablation_reg_recap/spiderman/ours/frame_}{0}{47}
    & \includegraphics[width=0.98\linewidth,height=0.65\linewidth]{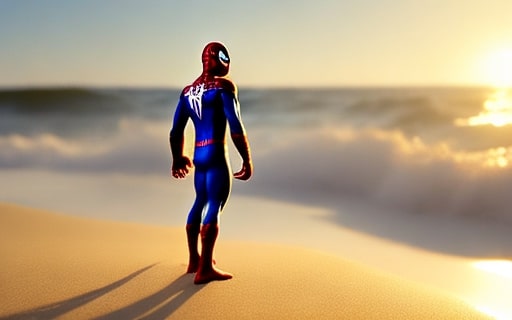}
    & \includegraphics[width=0.98\linewidth,height=0.65\linewidth]{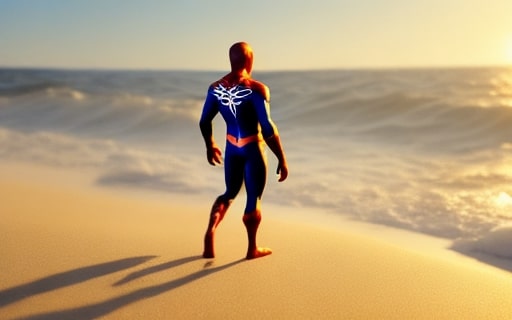} 
    & \includegraphics[width=0.98\linewidth,height=0.65\linewidth]{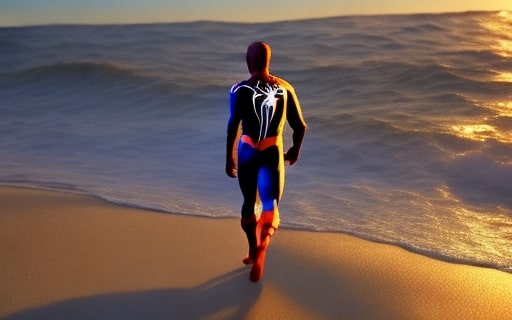} 
    & \includegraphics[width=0.98\linewidth,height=0.65\linewidth]{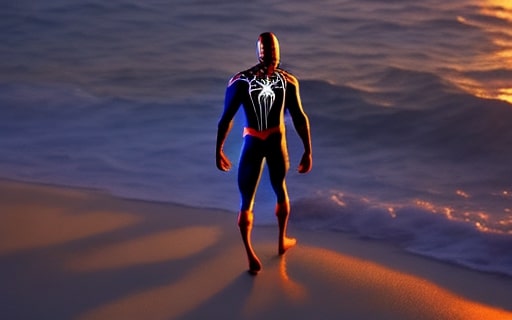}  
    & \includegraphics[width=0.98\linewidth,height=0.65\linewidth]{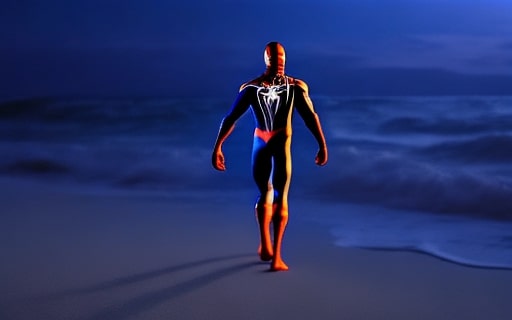}
\tabularnewline

\end{tabular}
\hfill{}
\par\end{centering}
\vspace{-0.5em}
\caption{Ablation on the effect of Video Synopsis Prompting (VSP) and Temporal Attention Regularization (TAR). Subsampled from 48 frames. 
Combination of TAR and VSP effectively enables long video generation with desired visual evolution. While individual strategy improves upon the baseline, there still lacks of desired dynamics. 
}
\label{fig:ablation_recap_reg}
\vspace{-0.5em}
\end{figure}

\paragraph{Experimental setting.}
To demonstrate the effectiveness of {\ourdm} in creating more dynamic videos, we run experiments and ablations on prompts, generated by ChatGPT\newcite{chatgpt}, that describe various visual transitions. All prompts and subprompts generated in the proposed video synopsis prompting step are provided in the Supp. Material. 
By default, we employ the state-of-the-art open-sourced T2V model VideoCrafter2\newcite{chen2024videocrafter2} with 320$\times$512 resolution as our base model, which is combined with the proposed video synopsis prompting and temporal attention regularization. We refer to this combination as our method or {\ourdm} throughout the experiments.

\vspace{-0.5em}
\subsection{Main Results}\label{subsec:main_results}
\vspace{-0.5em}
\paragraph{Comparison with other T2V methods.}
In \cref{fig:compare_t2v_16} and \cref{fig:compare_t2v_32} we compare our {\ourdm} with other commonly used T2V models, namely, ModelScope~\cite{wang2023modelscope}, LaVie~\cite{wang2023lavie} and AnimateDiff~\cite{guo2024animatediff}, for both 16 and 32 frame generation. For a fair comparison, we use the base model of LaVie without its cascaded components, \eg, the video super-resolution model.
Although all methods are able to generate meaningful results for 16-frame videos (see \cref{fig:compare_t2v_16}), the videos created by the other T2V models do not properly reflect the visual content specified by the input prompt. Given \myquote{A Ferrari driving on the road, starts to snow}, the other methods tend to focus on one particular state, \eg, the snowy scene, lacking dynamic progression throughout the video. In contrast, our {\ourdm} appropriately captures the weather transition from a clear day to a snowy one.

When generating 32 frames in one pass, as shown in \cref{fig:compare_t2v_32}, our method exhibits even greater advantages. The comparison methods yet again fail to generate content corresponding to the given prompt, but this time to the extent that the visual quality of the individual frames is also greatly compromised.  In contrast, our {\ourdm} is able to generate long videos with dynamic visual evolution. 
Based on these results, with a desire to further understand why other T2V models generalize poorly to long video generation, we analyze the temporal attention of these models, as detailed in the following paragraph.

\paragraph{Comparison on inter-frame similarity with real videos.}
To quantitatively assess inter-frame similarity in a video, we calculate the perceptual similarity between every pair of frames using the recently proposed metric DreamSim~\cite{dreamsim}, which has been demonstrated to align closely with human judgment.
In \cref{fig:compare_dreamsim}, we plot the similarity matrices of real videos and those synthesized by VideoCrafter2 and our {\ourdm}; the values in the matrix are normalized across all methods. VideoCrafter2 exhibits very high similarity across all frames, suggesting minimal visual dynamics, which is aligned with qualitative results. Our {\ourdm} on the other hand mimics the perceptual similarity correlation of real videos, affirming the effectiveness of our proposal for nursing the video dynamics.

Observing the resemblance between the temporal attention maps of the real videos and their similarity matrices, we attempt to directly employ a DreamSim-based similarity matrix as $\Delta A$ for regularization. As shown in \cref{fig:ablation_dreamsim}, this improves the temporal dynamics, leading to a gradual appearance of the rainbow.

\paragraph{Temporal attention analysis of other T2V models.}
In \cref{fig:ablation_T2V-TA}, we visualize the temporal attention layers of ModelScope~\cite{wang2023modelscope}, LaVie~\cite{wang2023lavie} and AnimateDiff~\cite{guo2024animatediff}.
It can be seen that ModelScope exhibits similar attention behavior to VideoCrafter (see \cref{fig:compare_TA_real_fake}), in that the temporal correlation significantly deteriorates when generating longer videos.  This is noticeable even for videos of 32 frames, twice the length of the standard option, and aligns with the qualitative comparison in \cref{fig:compare_t2v_32}.
In the Supp. Material we show that our {\ourdm} can also improve ModelScope's long video generation.
AnimateDiff\newcite{guo2024animatediff} and LaVie\newcite{wang2023lavie} demonstrate different temporal attention behavior, due to the incorporation of Rotary Positional Encoding\newcite{touvron2023llama} in the former and Sinusoidal Positional Encoding in the latter.
With the positional encoding, the models learn better temporal correlation among neighboring frames for 16 frames, showing a band-matrix structure more closely resembles that of real videos.
However, when generating videos longer than its training capacity,
the model faces considerable difficulty in preserving the desired temporal dynamics,
resulting in inferior synthesis quality, as depicted in \cref{fig:compare_t2v_32}.
The Rotary Positional Encoding employed in LaVie is a form of relative positional encoding, \ie, it depends on the relative offsets of frames, which could explain the periodic pattern seen in the attention maps.
While the Sinusoidal Positional Encoding used in AnimateDiff is based on the absolute frame index, leading to the model failing completely for indices unseen during training (past 16).
These observations concerning T2V models are interestingly aligned with prior studies regarding Positional Encoding on length generalization in Transformers\newcite{kazemnejad2023PELength} in the context of LLMs.

This comparison offers valuable insights into improving the training of the next generation of T2V models. For instance, omitting positional encoding can improve generalization capability, and incorporating a regularization loss on the temporal attention maps can help to enforce the desired temporal dynamics. 
Alternatively, one can employ a better combination of data format and positional encodings, as explored in the recent work\newcite{zhou2024transformers}, which achieves improved length generalization.
For instance, Randomized Positional Encoding\newcite{ruoss2023randomizedPE} can help to avoid overfitting on the position indices, and mixing up subsampled video sequences can further strengthen local correlations. Combining such techniques may improve the generalization to long video generation.

\vspace{-0.5em}
\subsection{Ablation Study}\label{subsec:ablation}
\vspace{-0.7em}
\paragraph{Ablation on the effect of TAR and {\ourrecaption}.}
We investigate the effects of the proposed Temporal Attention Regularization and Video Synopsis Prompting individually in \cref{fig:ablation_recap_reg}, where we generate videos of 48 frames in one pass based on the prompt \myquote{Spiderman on the beach from morning
to evening}, using the same initial noise. The synthesized video clips are presented in the first column as GIFs; the other images are subsampled from the full sequence.
The baseline model VideoCrafter2 struggles to synthesize a video faithful to the input prompt, generating a sequence of highly similar frames, with a stride-like texture in the background, that fail to depict the time-lapse video.
When employing the TAR, the model generates a more realistic sequence, however without the desired visual evolution; the single plain prompt is insufficient to describe the scene changes.   
Interestingly, while VSP provides a more descriptive summary of different visual states, without TAR, the temporal attention remains strongly correlated.
The model then attempts to depict the provided textual description, however with limited visual variation. 
When combining both strategies, our {\ourdm} can effectively synthesize the desired visual content, exhibiting improved dynamics with a more appealing time-lapse effect.

\paragraph{Ablation on regularization matrix.}
We further ablate by investigating the effect of using a different standard deviation $\sigma$ in the regularization matrix $\Delta A$, shown in \cref{fig:ablate_reg_matrix}.  
We start from applying regularization at the highest temporal resolution \ie, 64, since high-resolution temporal attention more greatly influences the video dynamics, as demonstrated in the temporal attention analysis in \cref{sec:attention-analysis}. 
The results show that decreasing $\sigma$ results in a stronger regularization effect, inducing more pronounced visual changes throughout the video (\eg compare row 2 to row 4, and notice the extent of the blooming of the flower). 
Going one step further, applying regularization also at a resolution of 32 results in the peony reaching its fullest bloom. However, when equally strong regularization is applied at both a resolution of 64 and 32, \ie, $\sigma_{64}=\sigma_{32}=1$, the visual changes can be too excessive, leaving the impression of poor temporal coherency across frames.
Empirically, we find that applying $\sigma_{64}=1$ strikes a good balance between dynamic changes and temporal coherency.

\vspace{-0.7em}
\subsection{Discussion}
\vspace{-0.5em}
\paragraph{Limitations.}
{\ourdm} offers a simple yet effective solution for improving pretrained T2V models, however, there are fundamental issues of pretrained models that may not be completely resolved via generative nursing at inference time only. 
Although our {\ourdm} has eased the process of reasoning prompts that involve dynamic evolution,
the model can still struggle with responding to the decomposed open-world prompts, resulting in visuals that are not aligned with the prompt, potentially due to limited capability of the text encoder\newcite{podell2023sdxl,liu2023evalcrafter}.
Nevertheless, several recent works\cite{chefer2023attendandexcite,li2023divide} have employed on-the-fly latent optimization to improve the textual alignment of a frozen T2I model. One may explore the combination of {\ourdm} with such techniques for further improvement.

\paragraph{Potential negative societal impact.}
Given the imbalanced nature of large-scale datasets, pretrained T2V models may inherit certain data biases, inaccurately representing the diversity of the overall population. 
These biases can potentially reinforce existing societal stereotypes and inequalities. Therefore, it is advisable to undertake proactive steps to identify and mitigate such biases, which may include the involvement of human reviewers in sensitive contexts.

\vspace{-0.5em}
\section{Conclusion}
\label{sec:conclusion} 
\vspace{-0.5em}
In this paper, we contribute two simple concepts, Video Synopsis Prompting (VSP) and Temporal Attention Regularization (TAR), that, when employed together, facilitate the generation of longer (\eg 64 frames), temporally coherent videos with improved dynamics. We show the benefit of both VSP and TAR on diverse prompts and in comparison to the state of the art, and ablate on the employed TAR regularization matrix. Besides motivating TAR, our analysis of temporal correlation in real videos may offer valuable insights towards improving design and training of the next generation of T2V models. For example, some form of positional encoding appears to be hampering generalization capability, while the incorporation of a regularization loss on temporal attention maps can help to enforce temporal dynamics. 
While {\ourdm} is readily applied to pretrained T2V models, future work may incorporate it during training for improved procedural dynamics, such as complex activities on respective data.

\clearpage
\small
\bibliographystyle{ieeetr_fullname}
\bibliography{reference}

\clearpage
\appendix

\renewcommand*{\thesection}{\Alph{section}}
\renewcommand{\thetable}{S.\arabic{table}}  
\renewcommand{\thefigure}{S.\arabic{figure}} 

\refstepcounter{figure} 
\refstepcounter{table} 
\refstepcounter{equation}
\setcounter{figure}{0}  
\setcounter{table}{0}  

\newcommand{\suppred}[1]{\textcolor{red}{#1}}

\begin{center}
    \Large \textbf{VSTAR: Generative Temporal Nursing for Longer} \\
    \Large \textbf{Dynamic Video Synthesis}\\
    \Large \textbf{Supplementary Material}\\
    \vspace{1.5em}
\end{center}

This supplementary material to the main paper is structured as follows:

\begin{itemize}[label={$\bullet$}]
    \item In \cref{sec:supp-more-results}, we provide more experimental results, including quantitative comparisons, a user study, additional visual results, and the combination of {\ourdm} and ModelScope. 
    \item In \cref{sec:supp-optimization}, we discuss our attempts and insights into optimization-based temporal generative nursing, which might spark interest for subsequent studies.
    \item In \cref{sec:supp-prompt}, we elaborate further  on Video Synopsis Prompting., \eg, how to instruct LLMs to generate the video synopsis. 
    \item In \cref{sec:supp-other-vis}, we provide additional visualizations of the regularization matrix and samples of collected real dynamic videos.
\end{itemize}

\section{More Experimental Results}\label{sec:supp-more-results}

\subsection{Quantitative Comparison}
\begin{figure*}[t]
\centering
\includegraphics[width=1.0\linewidth]{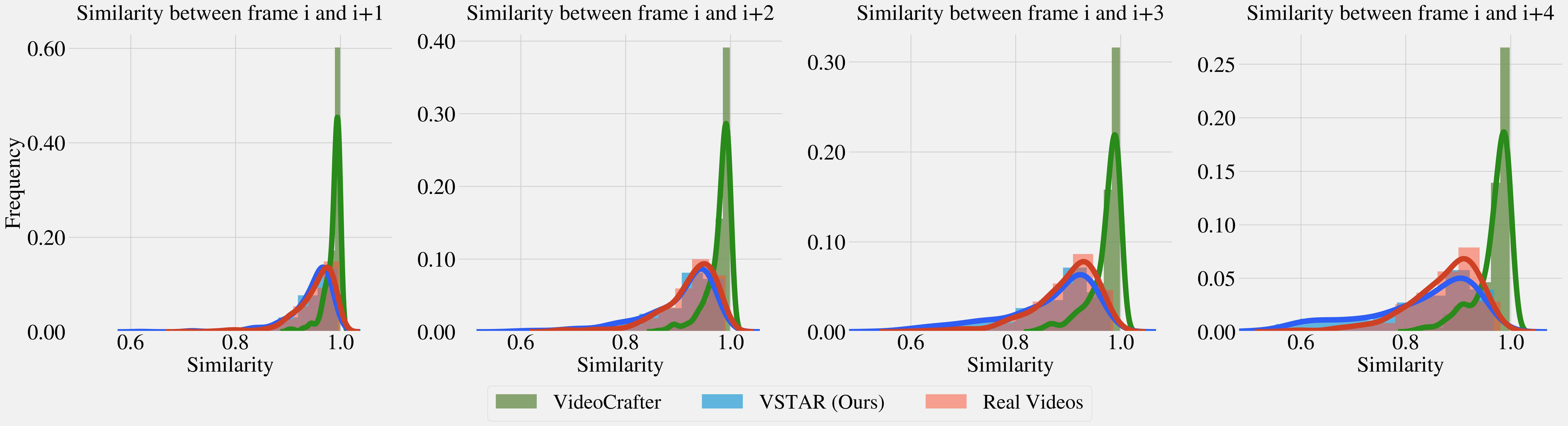}
\caption{
Comparison of DreamSim Similarity Distribution between real videos, VideoCrafter and our {\ourdm}. 
For preferred video dynamics, the similarity should decrease as the interval between frames increases, signaling a steady visual evolution. Meanwhile frames that are closer together should exhibit higher similarity compared to those further apart, indicating sustained temporal coherence.
For VideoCrafter2, the majority is located in the high similarity region regardless of the interval, indicating that there is limited visual variation within the video. 
The distribution of ours overlaps significantly with that of real videos, suggesting improved video dynamics without compromising coherency.
}
\label{fig:supp-dreamsim-hist}
\end{figure*}

We quantitatively compute the similarity of two frames a certain interval apart using the recent perceptual similarity metric DreamSim\newcite{dreamsim}. The similarity distribution of real dynamic videos, VideoCrafter2\newcite{chen2024videocrafter2} and our {\ourdm} are presented in \cref{fig:supp-dreamsim-hist}. 
For desired video dynamics, the similarity should decrease as the interval between frames increases, signaling a steady visual evolution. Meanwhile, frames that are closer together should exhibit higher similarity compared to those further apart, indicating preserved temporal coherence.
The distribution exhibited by VideoCrafter2 is highly concentrated at the high similarity region, even with large intervals. This can be explained by the fact that it often generates videos with limited visual variation over time, which is aligned with the qualitative results and analysis in Sec. \suppred{4.1} of main paper.
In contrast, with an increasing interval between frames, the distribution for both our {\ourdm} and real dynamic videos slightly shifts towards a region of lower similarity, indicating that more visual variation has been introduced within the video.
Our distribution extensively overlaps with that of real videos, suggesting that our results not only exhibit improved temporal dynamics, but also maintain the continuity.

\subsection{User Study}
For further evaluation, we conducted a user study to compare our {\ourdm} with the SOTA T2V model VideoCrafter2\newcite{chen2024videocrafter2}.
110 individuals with diverse backgrounds participated in the user study, working in fields such as computer vision, reinforcement learning, natural language processing, art design, medical engineering, mechanical engineering, and administrative management, among others.
We assess the videos across four dimensions: text alignment, video dynamics, visual quality and temporal coherency. 
Text alignment concerns whether the synthesized results properly reflect the input text prompt.
Video dynamics examines the dynamic visual changes within the progression of the video. 
A higher visual quality indicates fewer artifacts and distortions, leading to a more visually pleasing result.  
Temporal coherency evaluates if the result is temporally smooth, \ie, there are no abrupt or unexplained changes that could disrupt the viewing experience. For the first three aspects, participants are presented with paired results to evaluate, selecting one over the other or deeming them equivalent. Regarding temporal coherency, we pose a simple yes-or-no question, asking whether the participants perceive the video as being temporally smooth.

\begin{figure*}[t]
    \begin{centering}
    \setlength{\tabcolsep}{0.0em}
    \renewcommand{\arraystretch}{0}
    \par\end{centering}
    \begin{centering}
    \hfill{}%
	\begin{tabular}{@{}c}
        \centering
    \includegraphics[width=0.99\linewidth]{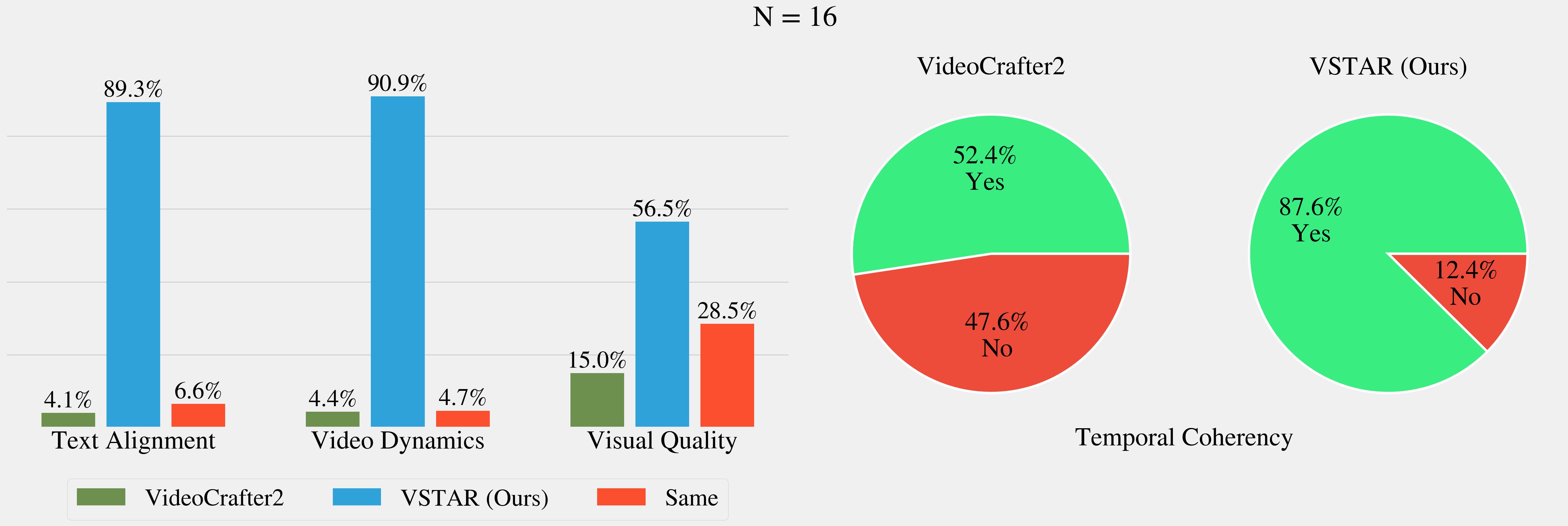} 
\tabularnewline
    \includegraphics[width=0.99\linewidth]{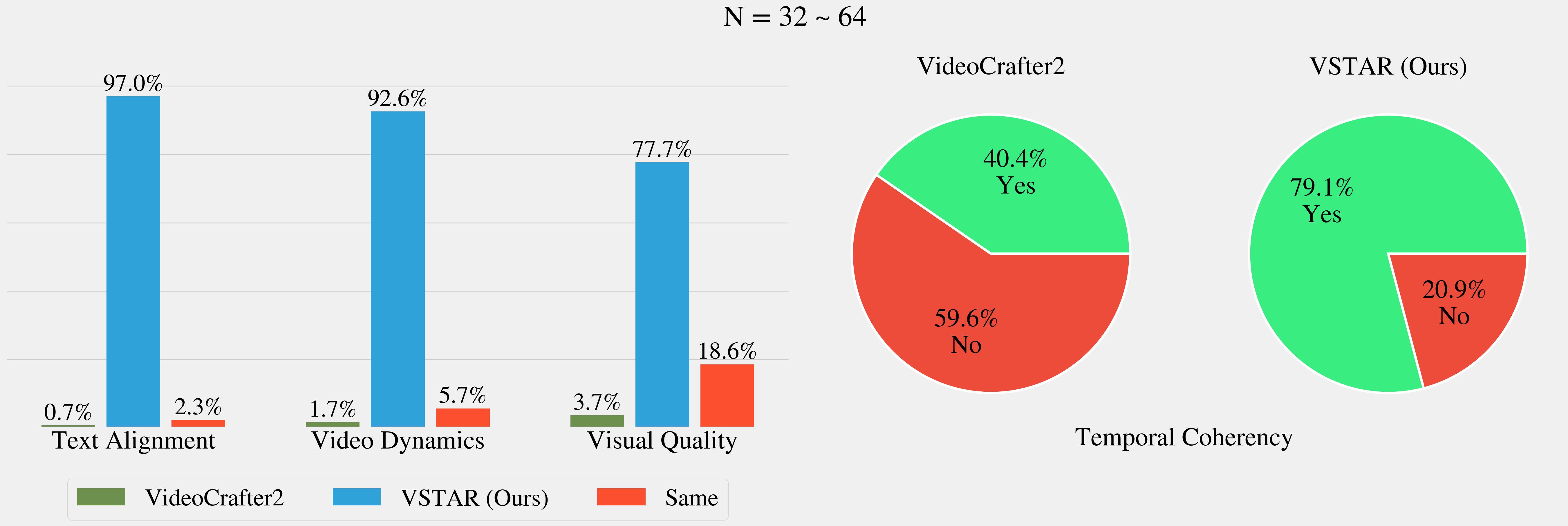} 
\end{tabular}
\hfill{}
\par\end{centering}
\caption{
User study on both standard $16$ frames and longer videos with $32\sim64$  frames.  For the first three aspects, participants review pairs of videos, choosing between them or rating them as the same. For temporal coherency, the numbers are the absolute probability that a participant perceives the video from the respective method as having smooth temporal progression.
}
\label{fig:supp-user-study}
\end{figure*}

The outcome is summarized in \cref{fig:supp-user-study}. Our {\ourdm} emerges as the preferred choice across various frame lengths from all aspects, with its advantages becoming more pronounced in the generation of longer videos with $N=32\sim64$.
Importantly, our method not only enhances video dynamics but also preserves temporal coherency. A majority of participants confirmed that our results exhibit smooth temporal transitions, with $87.6\%$ for standard-length videos and $79.1\%$ for longer videos agreeing to this assessment. This favorable reception surpasses the baseline VideoCrafter2, possibly as a result of its less engaging content.

\begin{figure*}[ht!]
    \begin{centering}
    \setlength{\tabcolsep}{0.0em}
    \renewcommand{\arraystretch}{0}
    \par\end{centering}
    \begin{centering}
    \hfill{}%
	\begin{tabular}{@{}c}
        \centering
    \includegraphics[width=0.45\linewidth]{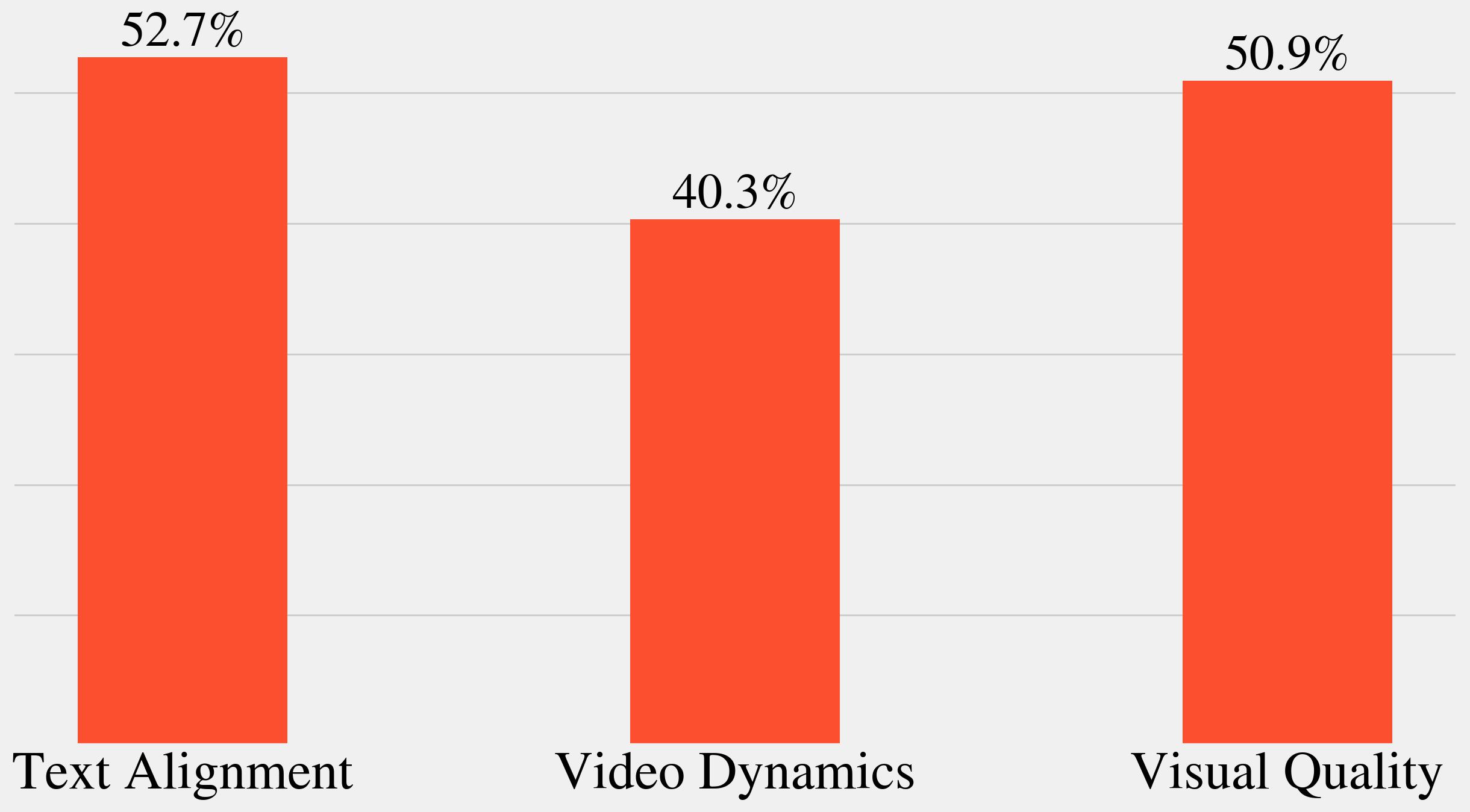} 
\tabularnewline
\end{tabular}
\hfill{}
\par\end{centering}
\vspace{-0.5em}
\caption{
User study on paired of videos, both generated by our {\ourdm}, to verify the consistency of our method's improvement, making it challenging for users to make a clear choice. Indeed, a large number of participants perceived both videos as identical across all three aspects. The rest had diverse preferences between the two videos. This demonstrates the consistency of our synthesis results and their closely matched quality.
}
\label{fig:supp-user-study-ours}
\end{figure*}

Additionally, we included pairs of videos, both generated by {\ourdm}, to verify the consistency of our method's improvement, making it challenging for users to make a clear choice.
As shown in \cref{fig:supp-user-study-ours}, participants indeed often found it difficult to differentiate, with $52.7\%$, $40.3\%$ and $50.9\%$ of them rating both videos as equal in terms of text alignment, visual dynamics, and visual quality, respectively. The remaining participants were divided in their preference between the two videos.
This indicates that our synthesis results are consistent and display a narrow gap between them.


\begin{figure}[ht!]
\begin{centering}
\setlength{\tabcolsep}{0.0em}
\renewcommand{\arraystretch}{0}
\par\end{centering}
\begin{centering}
\hfill{}%
 \begin{tabular}{
    m{0.0\linewidth}<{\centering} @{}
    m{0.16\linewidth}<{\centering} @{\hspace{0.015\linewidth}}
    m{0.16\linewidth}<{\centering} @{}
    m{0.16\linewidth}<{\centering} @{} %
    m{0.16\linewidth}<{\centering} @{}
    m{0.16\linewidth}<{\centering} @{}
    m{0.16\linewidth}<{\centering} @{}
    }
\tabularnewline
& & \multicolumn{5}{c}{\begin{tabular}{c}
   \myquote{Superman flying in the sky, sunny day becomes a dark rainy day}
\end{tabular}} 
\tabularnewline
    & \multirow{2}{*}{\begin{tabular}{c}
    $N = 48$ \\
    \animategraphics[autoplay,loop,width=0.98\linewidth,height=0.65\linewidth]{6}{appendix/figs/f48/superman_fly_8/frame_}{0}{47}
    \end{tabular}}
    & \includegraphics[width=0.98\linewidth,height=0.65\linewidth]{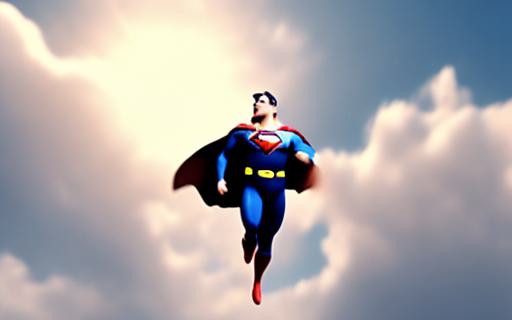}
    & \includegraphics[width=0.98\linewidth,height=0.65\linewidth]{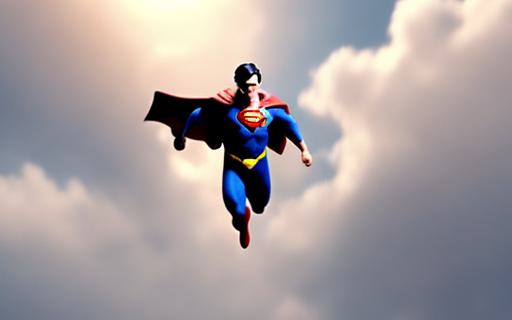} 
    & \includegraphics[width=0.98\linewidth,height=0.65\linewidth]{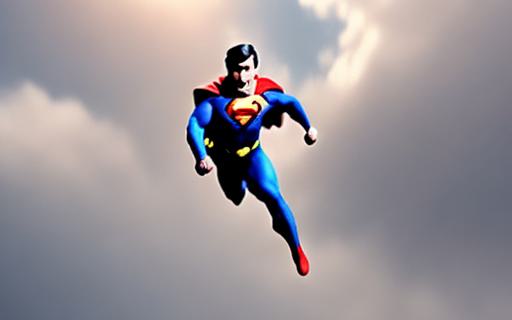} 
    & \includegraphics[width=0.98\linewidth,height=0.65\linewidth]{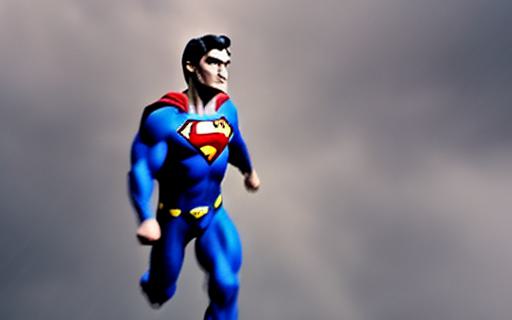}  
    & \includegraphics[width=0.98\linewidth,height=0.65\linewidth]{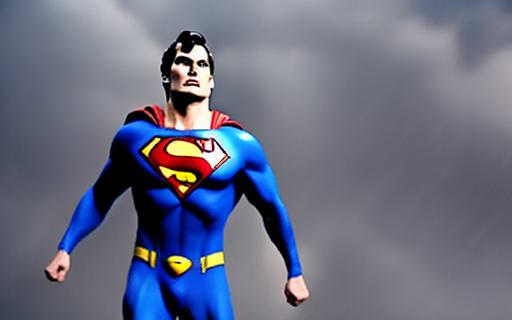}
\tabularnewline
    & %
    & \includegraphics[width=0.98\linewidth,height=0.65\linewidth]{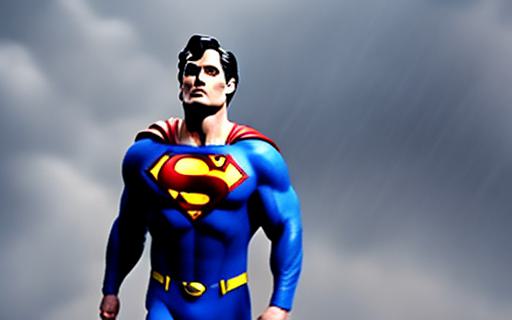}
    & \includegraphics[width=0.98\linewidth,height=0.65\linewidth]{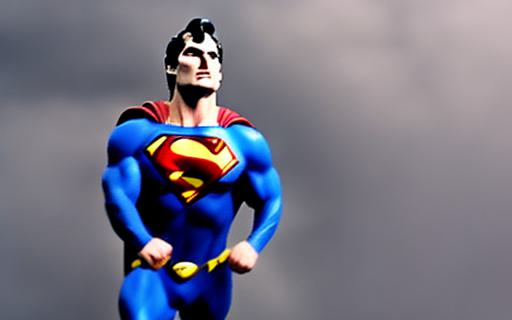} 
    & \includegraphics[width=0.98\linewidth,height=0.65\linewidth]{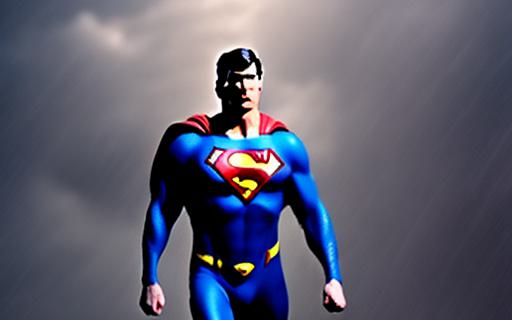} 
    & \includegraphics[width=0.98\linewidth,height=0.65\linewidth]{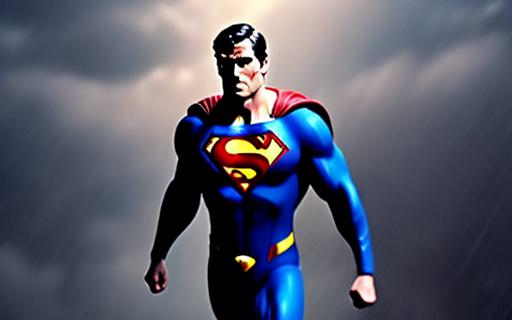} 
    & \includegraphics[width=0.98\linewidth,height=0.65\linewidth]{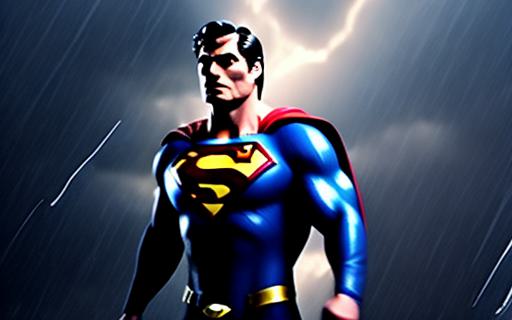}
\tabularnewline
& & \multicolumn{5}{c}{\begin{tabular}{c}
   \myquote{A young boy becomes an old man, and turns into a young girl}
\end{tabular}} 
\tabularnewline
    & \multirow{2}{*}{\begin{tabular}{c}
    $N = 64$ \\
    \animategraphics[autoplay,loop,width=0.98\linewidth,height=0.65\linewidth]{6}{appendix/figs/f64/loop_boy_girl_892/frame_}{0}{63}
    \end{tabular}}
    & \includegraphics[width=0.98\linewidth,height=0.65\linewidth]{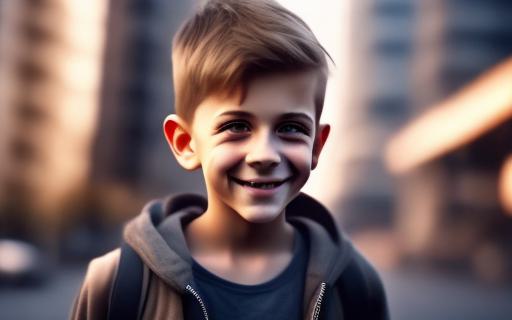}
    & \includegraphics[width=0.98\linewidth,height=0.65\linewidth]{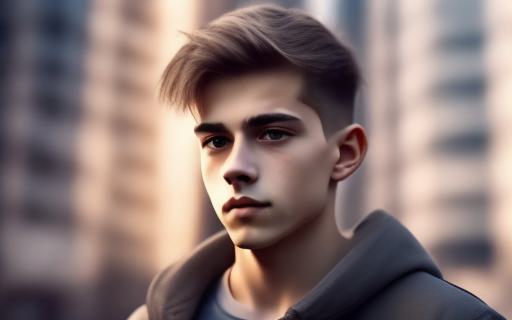} 
    & \includegraphics[width=0.98\linewidth,height=0.65\linewidth]{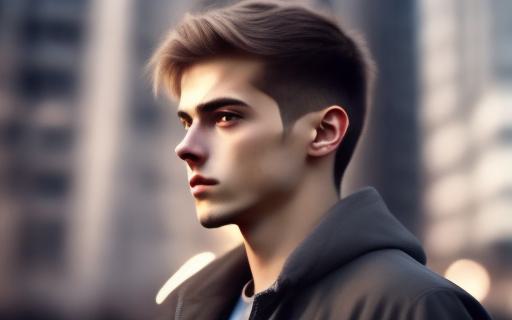} 
    & \includegraphics[width=0.98\linewidth,height=0.65\linewidth]{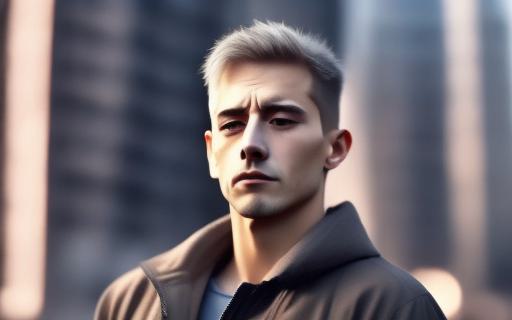}  
    & \includegraphics[width=0.98\linewidth,height=0.65\linewidth]{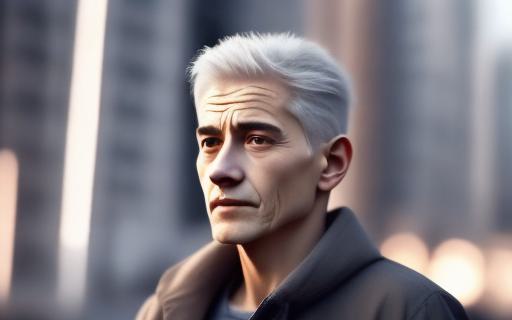}
\tabularnewline
    & %
    & \includegraphics[width=0.98\linewidth,height=0.65\linewidth]{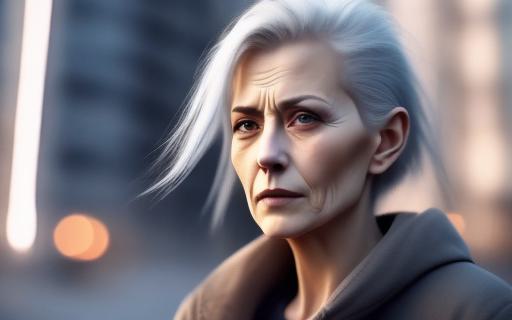}
    & \includegraphics[width=0.98\linewidth,height=0.65\linewidth]{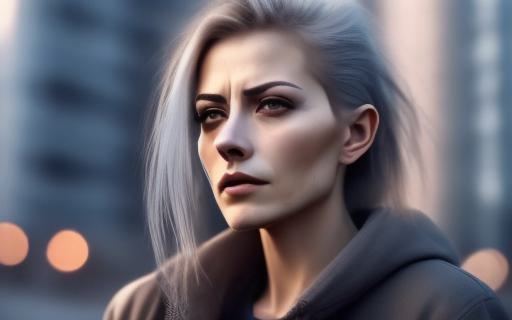} 
    & \includegraphics[width=0.98\linewidth,height=0.65\linewidth]{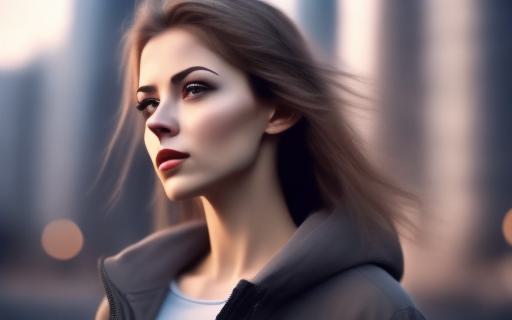} 
    & \includegraphics[width=0.98\linewidth,height=0.65\linewidth]{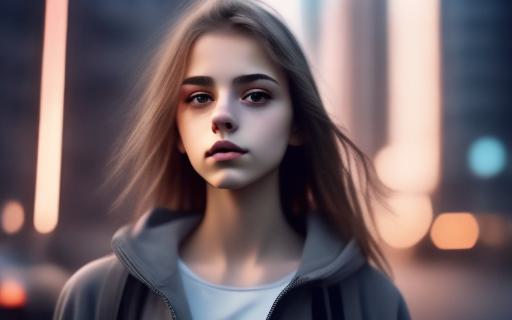} 
    & \includegraphics[width=0.98\linewidth,height=0.65\linewidth]{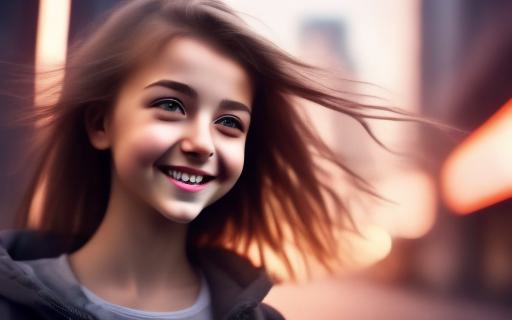}
\tabularnewline
& & \multicolumn{5}{c}{\begin{tabular}{c}
   \myquote{A day at the beach from dawn till dusk, bird's-eye view}
\end{tabular}} 
\tabularnewline
    & \multirow{2}{*}{\begin{tabular}{c}
    $N = 64$ \\
    \animategraphics[autoplay,loop,width=0.98\linewidth,height=0.65\linewidth]{6}{appendix/figs/f64/beach_birdeyeview_816/frame_}{0}{63}
    \end{tabular}}
    & \includegraphics[width=0.98\linewidth,height=0.65\linewidth]{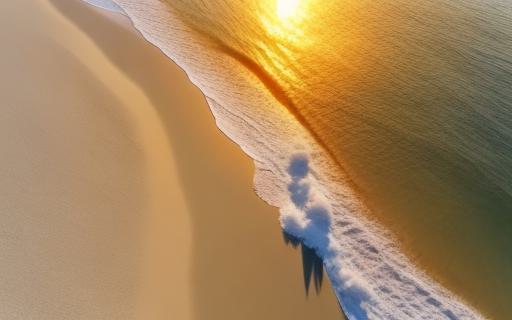}
    & \includegraphics[width=0.98\linewidth,height=0.65\linewidth]{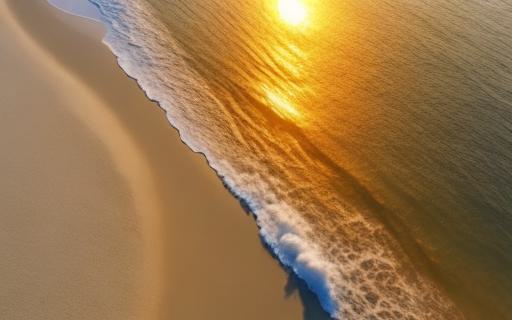} 
    & \includegraphics[width=0.98\linewidth,height=0.65\linewidth]{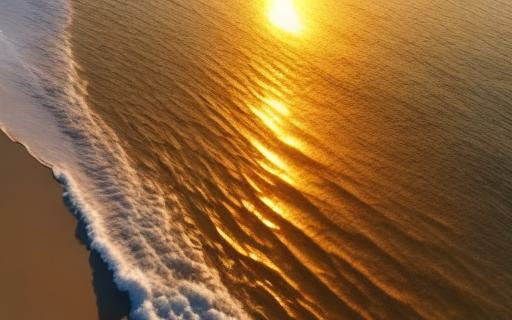} 
    & \includegraphics[width=0.98\linewidth,height=0.65\linewidth]{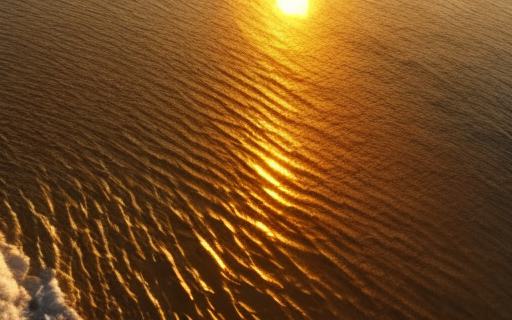}  
    & \includegraphics[width=0.98\linewidth,height=0.65\linewidth]{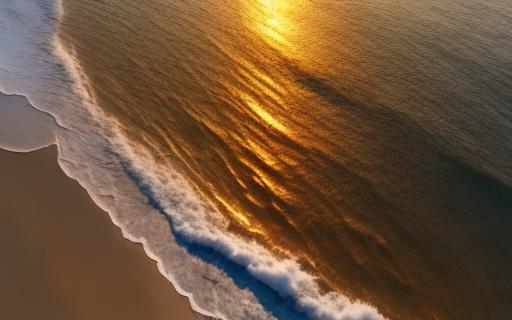}
\tabularnewline
    & %
    & \includegraphics[width=0.98\linewidth,height=0.65\linewidth]{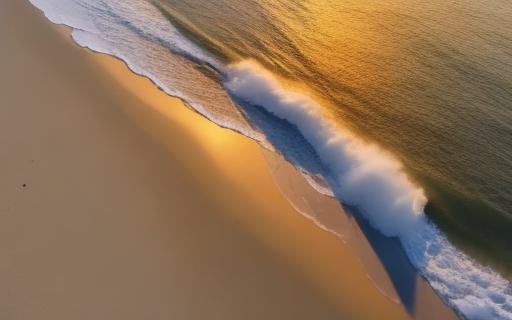}
    & \includegraphics[width=0.98\linewidth,height=0.65\linewidth]{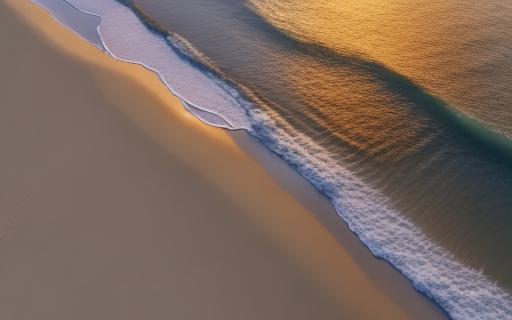} 
    & \includegraphics[width=0.98\linewidth,height=0.65\linewidth]{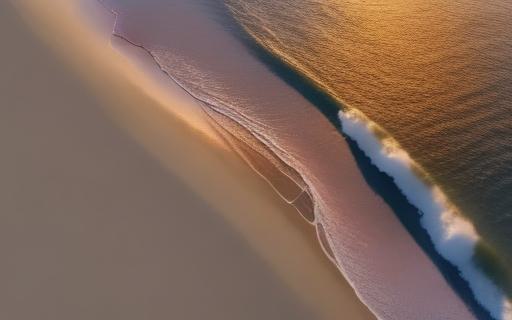} 
    & \includegraphics[width=0.98\linewidth,height=0.65\linewidth]{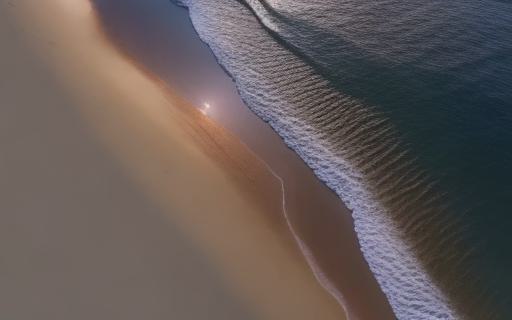} 
    & \includegraphics[width=0.98\linewidth,height=0.65\linewidth]{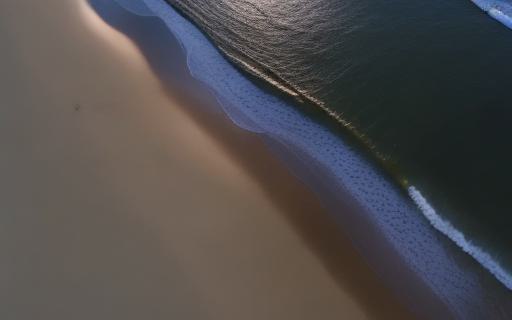}
\tabularnewline
& & \multicolumn{5}{c}{\begin{tabular}{c}
   \myquote{The seasonal cycle of a lake from frozen winter to autumn}
\end{tabular}} 
\tabularnewline
    & \multirow{2}{*}{\begin{tabular}{c}
    $N = 64$ \\
    \animategraphics[autoplay,loop,width=0.98\linewidth,height=0.65\linewidth]{6}{appendix/figs/f64/lake_winter_autumn_713/frame_}{0}{63}
    \end{tabular}}
    & \includegraphics[width=0.98\linewidth,height=0.65\linewidth]{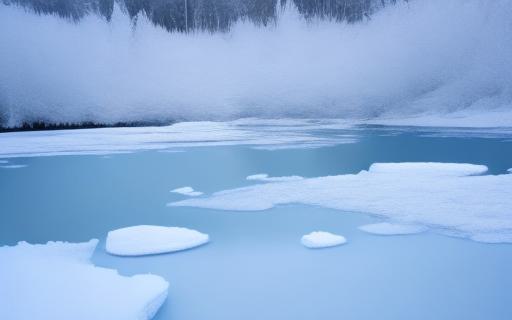}
    & \includegraphics[width=0.98\linewidth,height=0.65\linewidth]{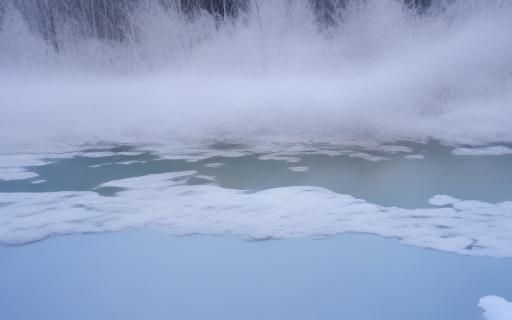} 
    & \includegraphics[width=0.98\linewidth,height=0.65\linewidth]{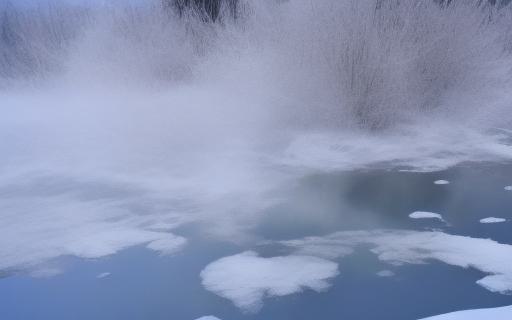} 
    & \includegraphics[width=0.98\linewidth,height=0.65\linewidth]{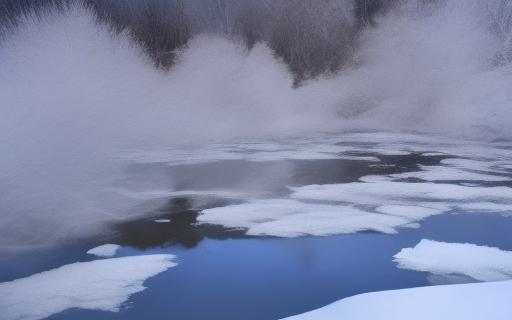}  
    & \includegraphics[width=0.98\linewidth,height=0.65\linewidth]{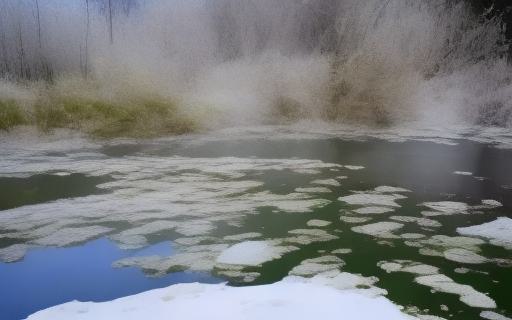}
\tabularnewline
    & %
    & \includegraphics[width=0.98\linewidth,height=0.65\linewidth]{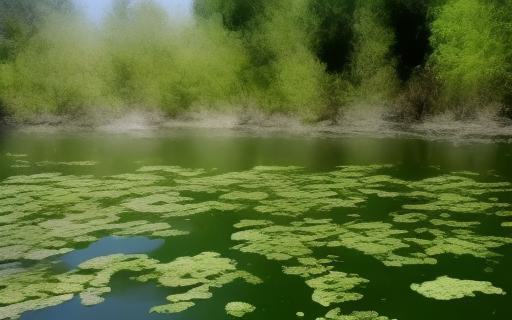}
    & \includegraphics[width=0.98\linewidth,height=0.65\linewidth]{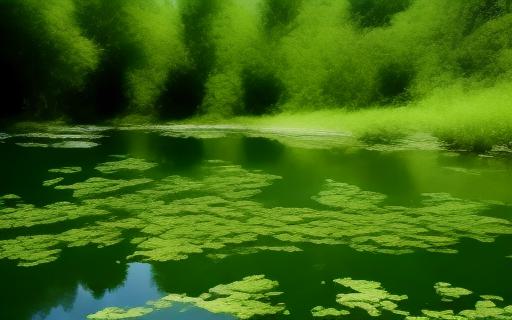} 
    & \includegraphics[width=0.98\linewidth,height=0.65\linewidth]{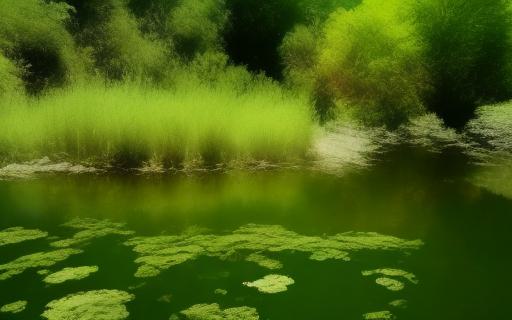} 
    & \includegraphics[width=0.98\linewidth,height=0.65\linewidth]{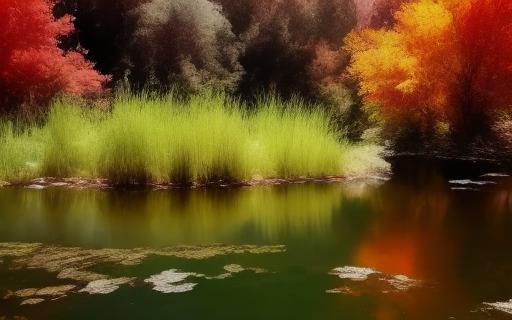} 
    & \includegraphics[width=0.98\linewidth,height=0.65\linewidth]{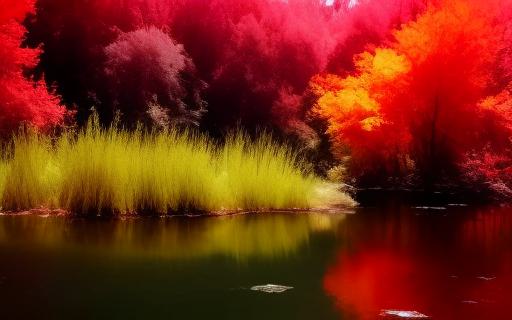}
\tabularnewline
\end{tabular}
\hfill{}
\par\end{centering}
\caption{Qualitative results of videos with 48 and 64 frames synthesized by {\ourdm}. Images are sub-sampled from the sequence.
Note that the first column is a GIF, best viewed in \emph{Acrobat Reader}.
}
\label{fig:supp_ours_f64}
\end{figure}

\begin{figure}[ht!]
\begin{centering}
\setlength{\tabcolsep}{0.0em}
\renewcommand{\arraystretch}{0}
\par\end{centering}
\begin{centering}
\hfill{}%
 \begin{tabular}{
    m{0.0\linewidth}<{\centering} @{}
    m{0.16\linewidth}<{\centering} @{\hspace{0.005\linewidth}}
    m{0.16\linewidth}<{\centering} @{}
    m{0.16\linewidth}<{\centering} @{} %
    m{0.16\linewidth}<{\centering} @{}
    m{0.16\linewidth}<{\centering} @{}
    m{0.16\linewidth}<{\centering} @{}
    }
\tabularnewline
& \multicolumn{6}{c}{\begin{tabular}{c}
  \myquote{A makeup transformation}
\end{tabular}} 
\tabularnewline
    & \animategraphics[autoplay,loop,width=0.98\linewidth,height=0.65\linewidth]{6}{appendix/figs/f32/makeup_391/frame_}{0}{31}
    & \includegraphics[width=0.98\linewidth,height=0.65\linewidth]{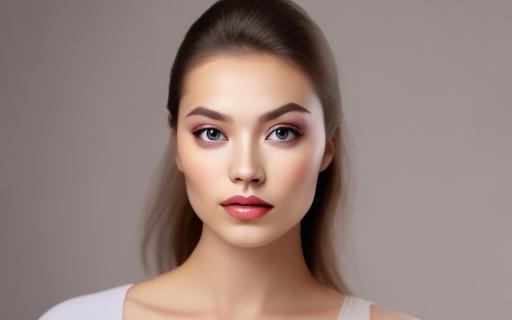}
    & \includegraphics[width=0.98\linewidth,height=0.65\linewidth]{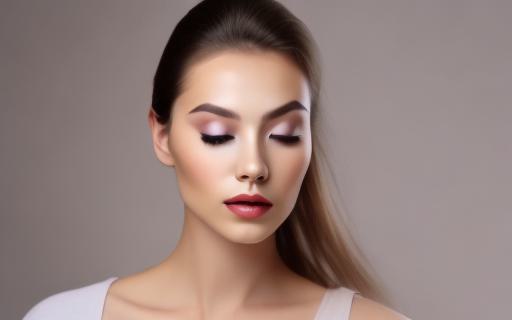} 
    & \includegraphics[width=0.98\linewidth,height=0.65\linewidth]{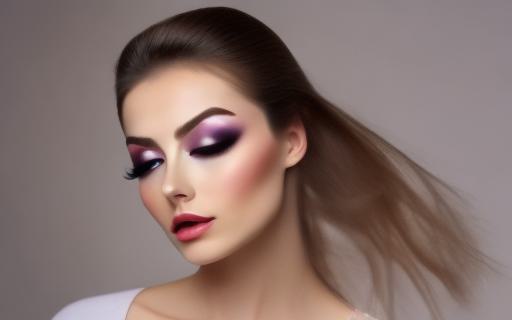} 
    & \includegraphics[width=0.98\linewidth,height=0.65\linewidth]{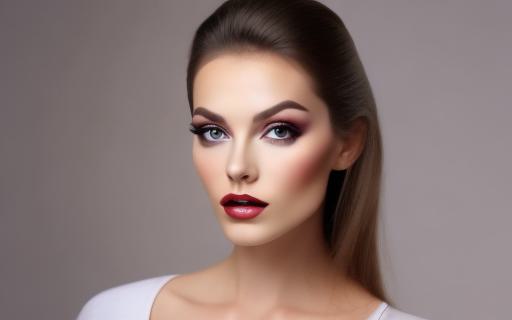}  
    & \includegraphics[width=0.98\linewidth,height=0.65\linewidth]{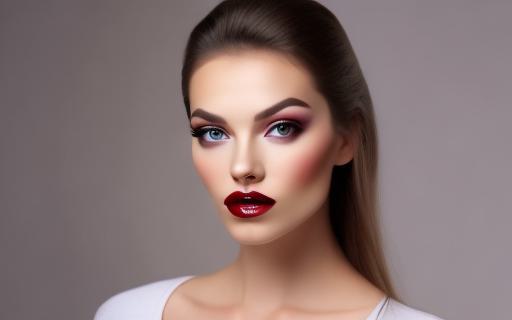}
\tabularnewline
& \multicolumn{6}{c}{\begin{tabular}{c}
  \myquote{A landscape transitioning from winter to spring }
\end{tabular}} 
\tabularnewline
    & \animategraphics[autoplay,loop,width=0.98\linewidth,height=0.65\linewidth]{6}{appendix/figs/f32/winter_spring_569/frame_}{0}{31}
    & \includegraphics[width=0.98\linewidth,height=0.65\linewidth]{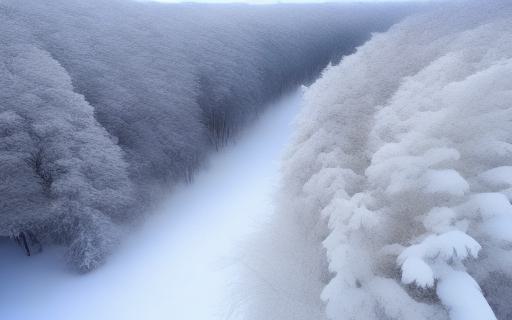}
    & \includegraphics[width=0.98\linewidth,height=0.65\linewidth]{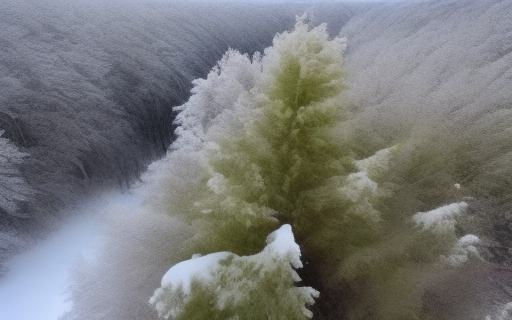} 
    & \includegraphics[width=0.98\linewidth,height=0.65\linewidth]{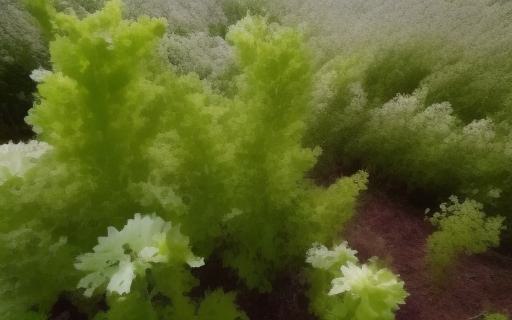} 
    & \includegraphics[width=0.98\linewidth,height=0.65\linewidth]{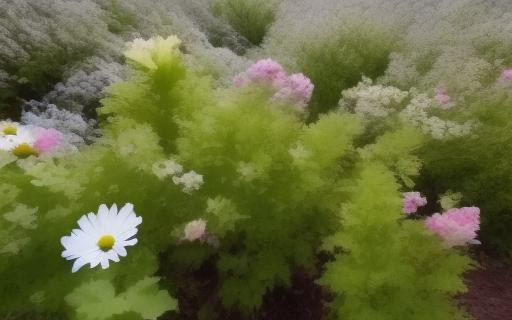}  
    & \includegraphics[width=0.98\linewidth,height=0.65\linewidth]{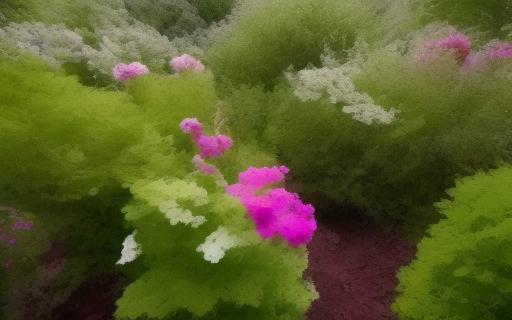}
\tabularnewline
& \multicolumn{6}{c}{\begin{tabular}{c}
  \myquote{A boy is getting old }
\end{tabular}} 
\tabularnewline
    & \animategraphics[autoplay,loop,width=0.98\linewidth,height=0.65\linewidth]{6}{appendix/figs/f32/boy_getting_old_1276/frame_}{0}{31}
    & \includegraphics[width=0.98\linewidth,height=0.65\linewidth]{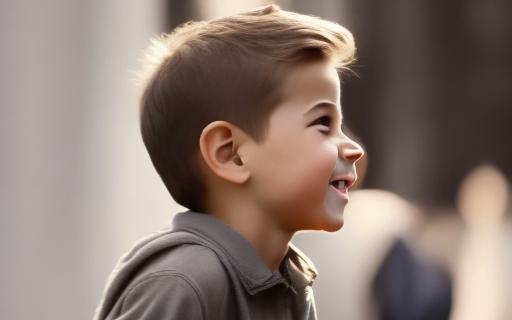}
    & \includegraphics[width=0.98\linewidth,height=0.65\linewidth]{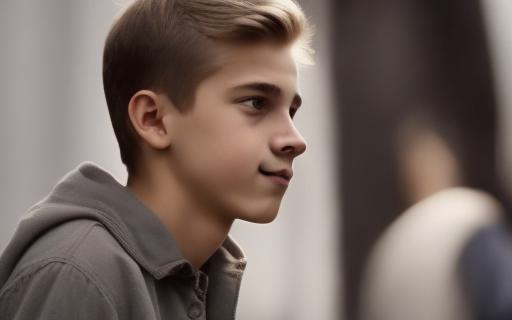} 
    & \includegraphics[width=0.98\linewidth,height=0.65\linewidth]{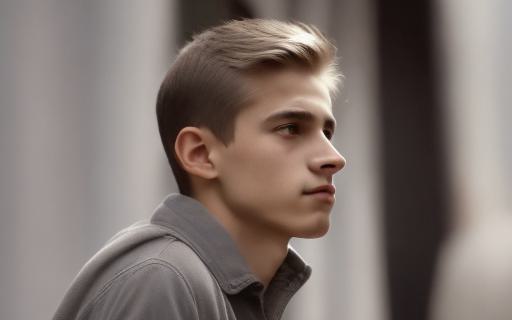} 
    & \includegraphics[width=0.98\linewidth,height=0.65\linewidth]{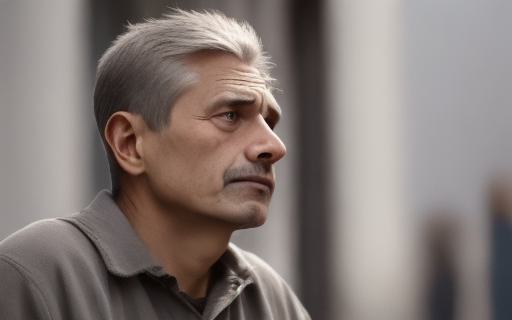}  
    & \includegraphics[width=0.98\linewidth,height=0.65\linewidth]{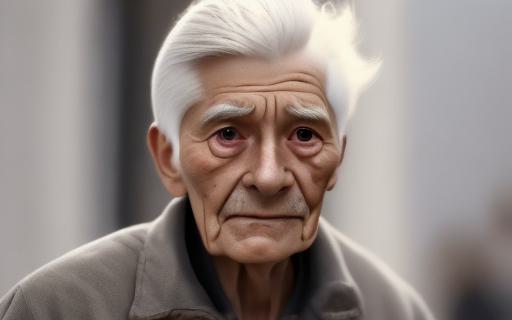}
\tabularnewline
\end{tabular}
\hfill{}
\par\end{centering}
\caption{ Qualitative results of videos with 32 frames synthesized by {\ourdm}. Images are sub-sampled from the sequence.
Note that the first column is a GIF, best viewed in \emph{Acrobat Reader}.
}
\label{fig:supp_ours_f32}
\end{figure}

\begin{figure}[ht!]
\begin{centering}
\setlength{\tabcolsep}{0.0em}
\renewcommand{\arraystretch}{0}
\par\end{centering}
\begin{centering}
\hfill{}%
 \begin{tabular}{
    m{0.00\linewidth}<{\centering} @{}
    m{0.16\linewidth}<{\centering} @{}
    m{0.16\linewidth}<{\centering} @{}
    m{0.16\linewidth}<{\centering} @{} %
    m{0.16\linewidth}<{\centering} @{}
    m{0.16\linewidth}<{\centering} @{}
    m{0.16\linewidth}<{\centering} @{}
    }
    &  \multicolumn{6}{c}{
    \begin{tabular}{c}
    \myquote{A mural being painted on a city wall}
    \end{tabular}} 
\tabularnewline
    & \animategraphics[autoplay,loop,width=0.98\linewidth,height=0.65\linewidth]{8}{appendix/figs/f16/mural_painted_1603/frame_}{0}{15}
    & \includegraphics[width=0.98\linewidth,height=0.65\linewidth]{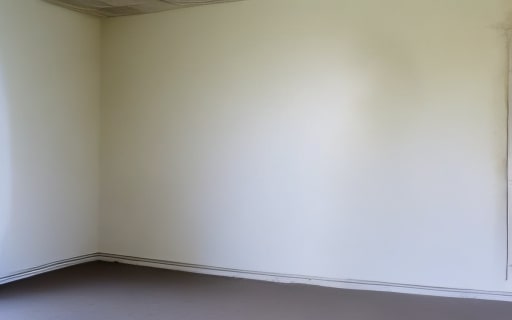}
    & \includegraphics[width=0.98\linewidth,height=0.65\linewidth]{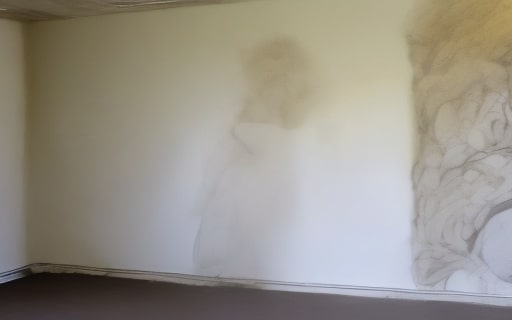} 
    & \includegraphics[width=0.98\linewidth,height=0.65\linewidth]{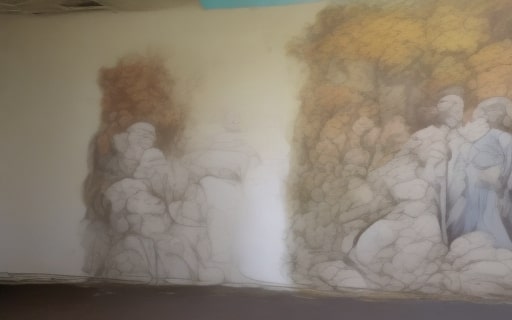} 
    & \includegraphics[width=0.98\linewidth,height=0.65\linewidth]{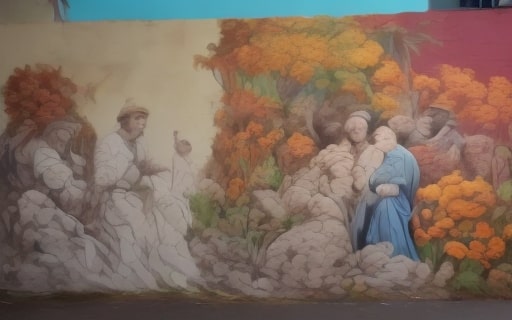} 
    & \includegraphics[width=0.98\linewidth,height=0.65\linewidth]{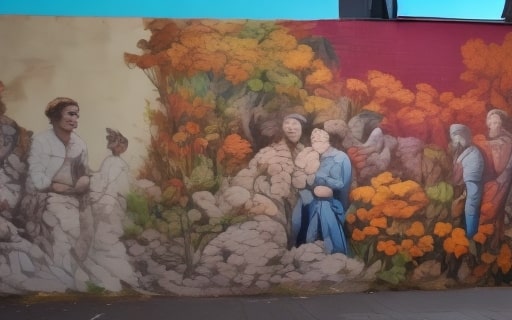}
\tabularnewline
    &  \multicolumn{6}{c}{
    \begin{tabular}{c}
    \myquote{The process of a woman losing 50kg of weight}
    \end{tabular}} 
\tabularnewline
    & \animategraphics[autoplay,loop,width=0.98\linewidth,height=0.65\linewidth]{8}{appendix/figs/f16/lose_weight_224/frame_}{0}{15}
    & \includegraphics[width=0.98\linewidth,height=0.65\linewidth]{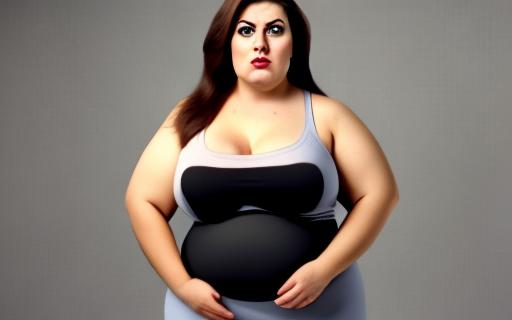}
    & \includegraphics[width=0.98\linewidth,height=0.65\linewidth]{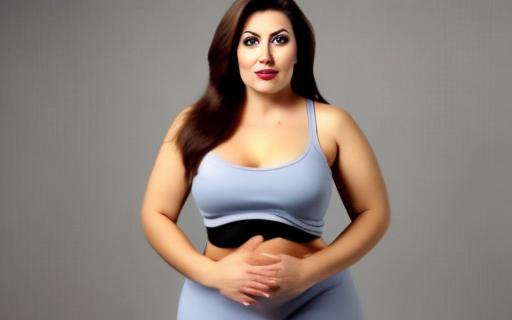} 
    & \includegraphics[width=0.98\linewidth,height=0.65\linewidth]{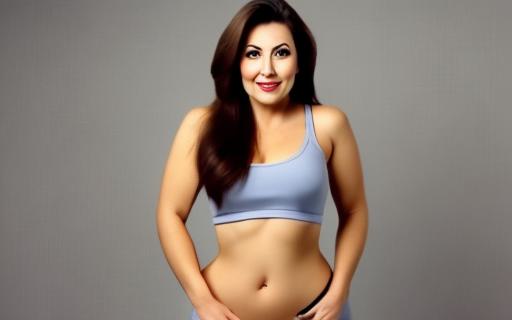} 
    & \includegraphics[width=0.98\linewidth,height=0.65\linewidth]{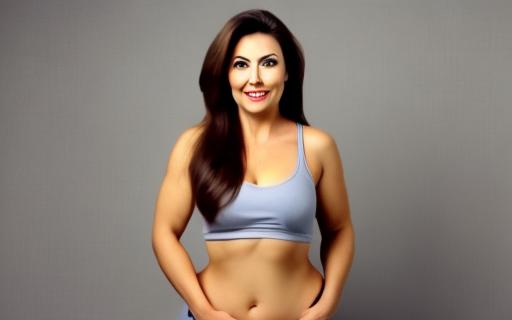} 
    & \includegraphics[width=0.98\linewidth,height=0.65\linewidth]{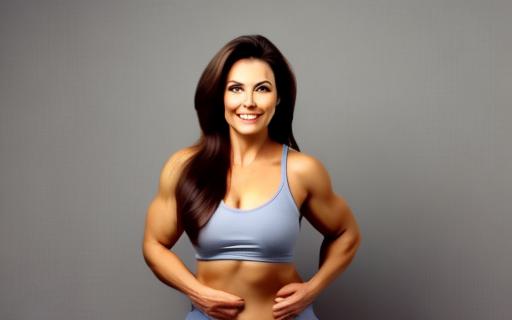}
\tabularnewline
    &  \multicolumn{6}{c}{
    \begin{tabular}{c}
    \myquote{A caterpillar transforming into a butterfly}
    \end{tabular}} 
\tabularnewline
    & \animategraphics[autoplay,loop,width=0.98\linewidth,height=0.65\linewidth]{8}{appendix/figs/f16/butterfly_transform_755/frame_}{0}{15}
    & \includegraphics[width=0.98\linewidth,height=0.65\linewidth]{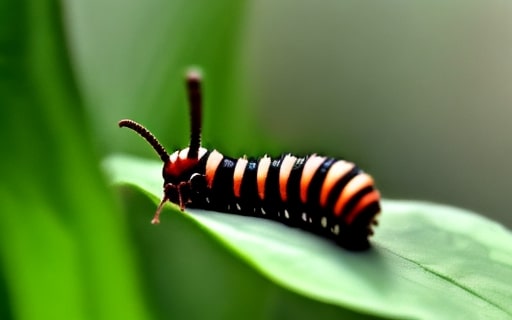}
    & \includegraphics[width=0.98\linewidth,height=0.65\linewidth]{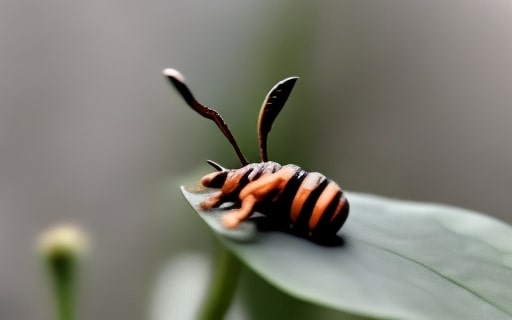} 
    & \includegraphics[width=0.98\linewidth,height=0.65\linewidth]{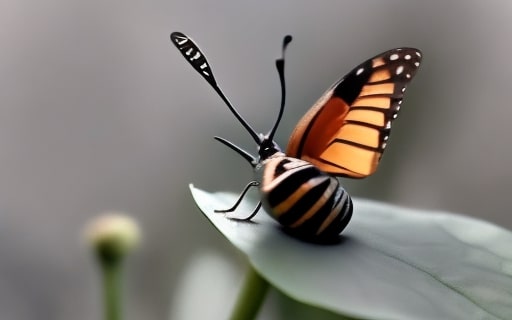} 
    & \includegraphics[width=0.98\linewidth,height=0.65\linewidth]{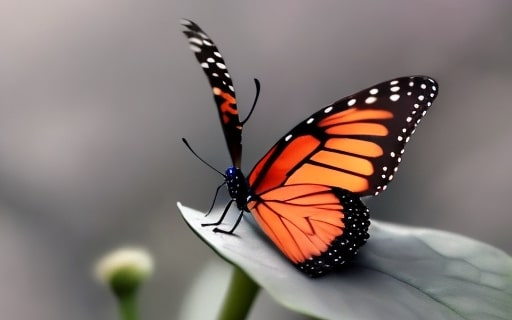} 
    & \includegraphics[width=0.98\linewidth,height=0.65\linewidth]{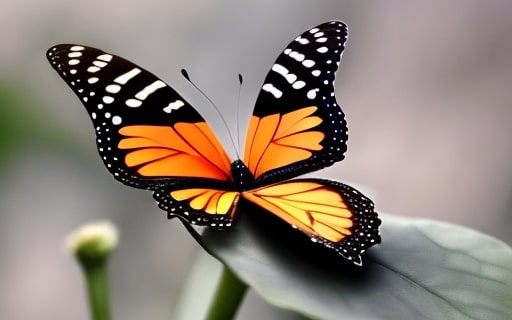}
\tabularnewline
    &  \multicolumn{6}{c}{
    \begin{tabular}{c}
    \myquote{A female person's hairstyle changing through the years}
    \end{tabular}} 
\tabularnewline
    & \animategraphics[autoplay,loop,width=0.98\linewidth,height=0.65\linewidth]{8}{appendix/figs/f16/female_hairstyle_578/frame_}{0}{15}
    & \includegraphics[width=0.98\linewidth,height=0.65\linewidth]{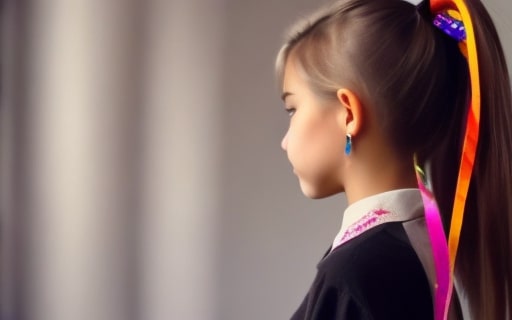}
    & \includegraphics[width=0.98\linewidth,height=0.65\linewidth]{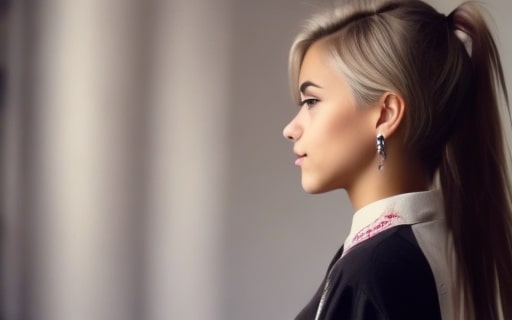} 
    & \includegraphics[width=0.98\linewidth,height=0.65\linewidth]{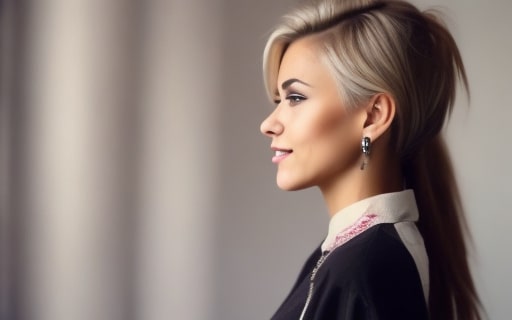} 
    & \includegraphics[width=0.98\linewidth,height=0.65\linewidth]{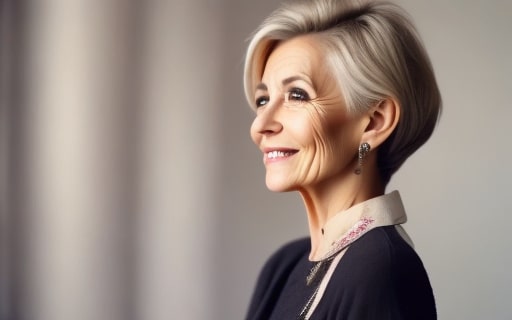} 
    & \includegraphics[width=0.98\linewidth,height=0.65\linewidth]{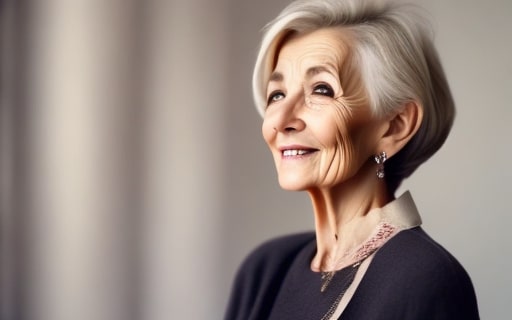}
\tabularnewline
    &  \multicolumn{6}{c}{
    \begin{tabular}{c}
    \myquote{A pizza is being made}
    \end{tabular}} 
\tabularnewline
    & \animategraphics[autoplay,loop,width=0.98\linewidth,height=0.65\linewidth]{8}{appendix/figs/f16/pizza_47/frame_}{0}{15}
    & \includegraphics[width=0.98\linewidth,height=0.65\linewidth]{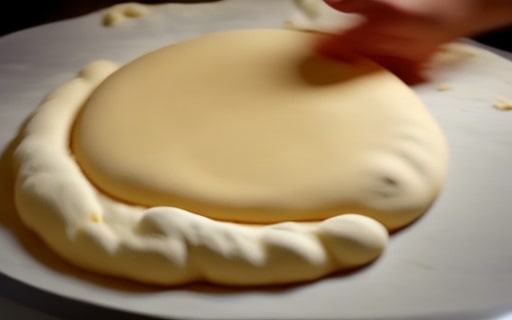}
    & \includegraphics[width=0.98\linewidth,height=0.65\linewidth]{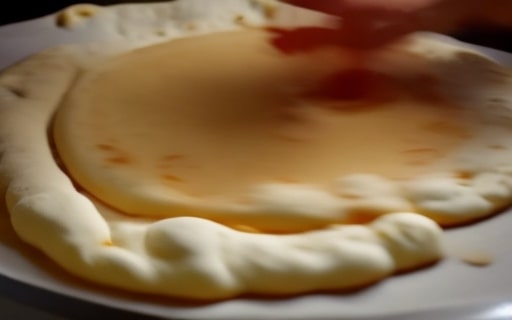} 
    & \includegraphics[width=0.98\linewidth,height=0.65\linewidth]{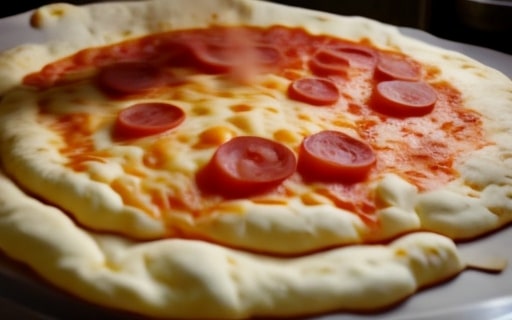} 
    & \includegraphics[width=0.98\linewidth,height=0.65\linewidth]{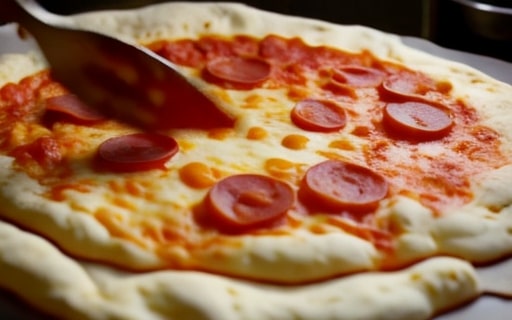} 
    & \includegraphics[width=0.98\linewidth,height=0.65\linewidth]{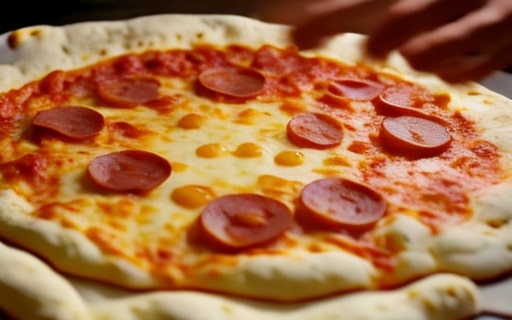}
\tabularnewline
\end{tabular}
\hfill{}
\par\end{centering}
\caption{Qualitative results of videos with 16 frames synthesized by {\ourdm}. Images are sub-sampled from the sequence.
Note that the first column is a GIF, best viewed in \emph{Acrobat Reader}.
}
\label{fig:supp_ours_f16}
\end{figure}

\begin{figure}[ht!]
\begin{centering}
\setlength{\tabcolsep}{0.0em}
\renewcommand{\arraystretch}{0}
\par\end{centering}
\begin{centering}
\hfill{}%
 \begin{tabular}{
    m{0.03\linewidth}<{\centering} @{}
    m{0.16\linewidth}<{\centering} @{}
    m{0.16\linewidth}<{\centering} @{}
    m{0.16\linewidth}<{\centering} @{} %
    m{0.16\linewidth}<{\centering} @{}
    m{0.16\linewidth}<{\centering} @{}
    m{0.16\linewidth}<{\centering} @{}
    }
\tabularnewline
& \multicolumn{6}{c}{\begin{tabular}{c}
\myquote{A flower starts to bloom}
\end{tabular}} 
\tabularnewline
    \multirow{1}{*}
    {\rotatebox{90}{
          \scriptsize ModelScope
        \hspace{-1.5\linewidth}
    }} 
    & \animategraphics[autoplay,loop,width=0.98\linewidth,height=0.65\linewidth]{8}{appendix/figs/f16/flower_bloom_187/ModelScope/frame_}{0}{15}
    & \includegraphics[width=0.98\linewidth,height=0.65\linewidth]{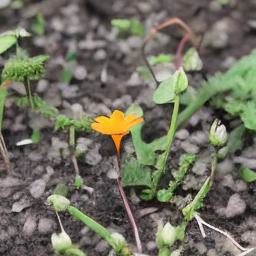}
    & \includegraphics[width=0.98\linewidth,height=0.65\linewidth]{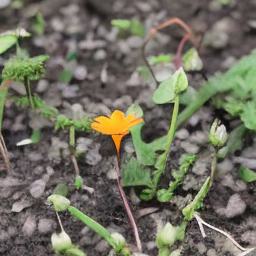} 
    & \includegraphics[width=0.98\linewidth,height=0.65\linewidth]{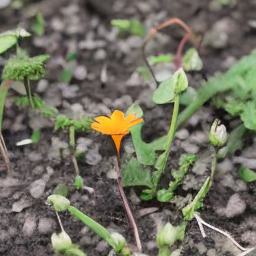} 
    & \includegraphics[width=0.98\linewidth,height=0.65\linewidth]{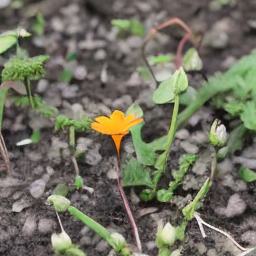} 
    & \includegraphics[width=0.98\linewidth,height=0.65\linewidth]{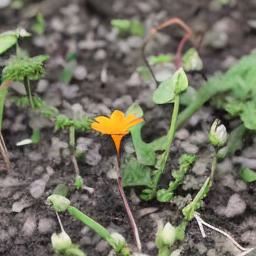}
\tabularnewline
\multirow{1}{*}
    {\rotatebox{90}{
        \begin{tabular}{c}
          \scriptsize LaVie
        \end{tabular}
        \hspace{-1.2\linewidth}
    }} 
    & \animategraphics[autoplay,loop,width=0.98\linewidth,height=0.65\linewidth]{8}{appendix/figs/f16/flower_bloom_187/LaVie/frame_}{0}{15}
    & \includegraphics[width=0.98\linewidth,height=0.65\linewidth]{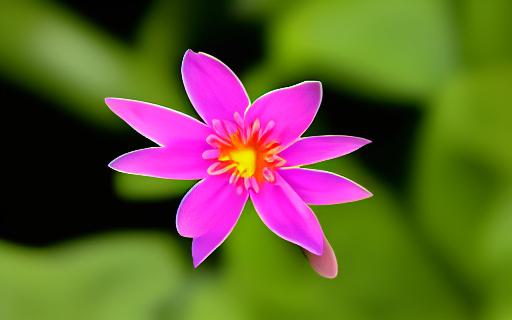}
    & \includegraphics[width=0.98\linewidth,height=0.65\linewidth]{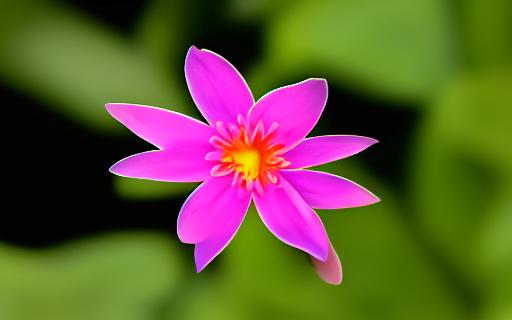} 
    & \includegraphics[width=0.98\linewidth,height=0.65\linewidth]{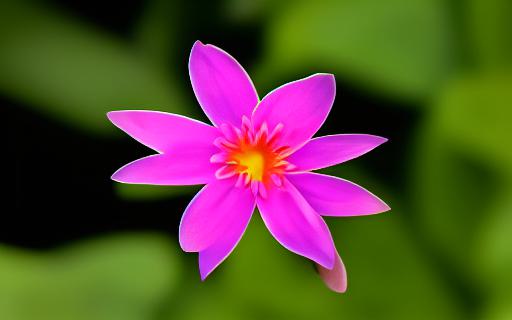} 
    & \includegraphics[width=0.98\linewidth,height=0.65\linewidth]{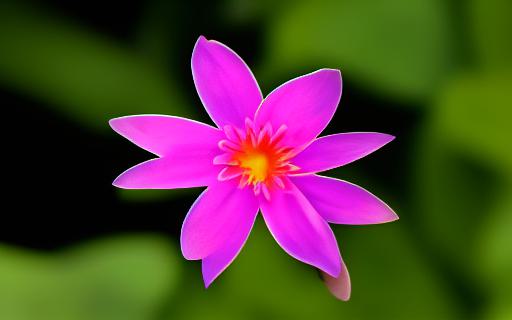} 
    & \includegraphics[width=0.98\linewidth,height=0.65\linewidth]{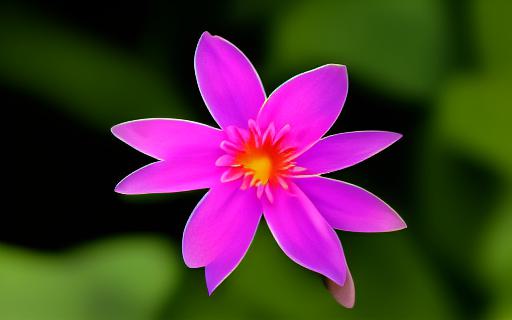}
\tabularnewline
\multirow{1}{*}
    {\rotatebox{90}{
        \begin{tabular}{c}
          \scriptsize AnimateDiff
        \end{tabular}
        \hspace{-2\linewidth}
    }} 
    & \animategraphics[autoplay,loop,width=0.98\linewidth,height=0.65\linewidth]{8}{appendix/figs/f16/flower_bloom_187/AnimateDiff/frame_}{0}{15}
    & \includegraphics[width=0.98\linewidth,height=0.65\linewidth]{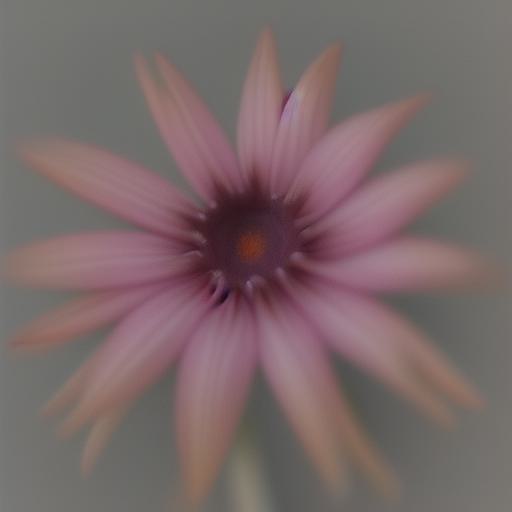}
    & \includegraphics[width=0.98\linewidth,height=0.65\linewidth]{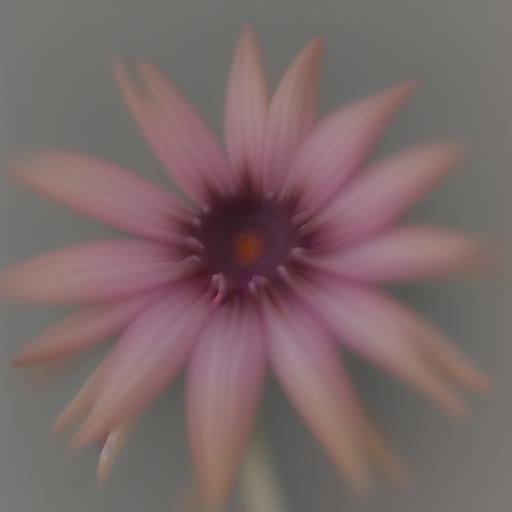} 
    & \includegraphics[width=0.98\linewidth,height=0.65\linewidth]{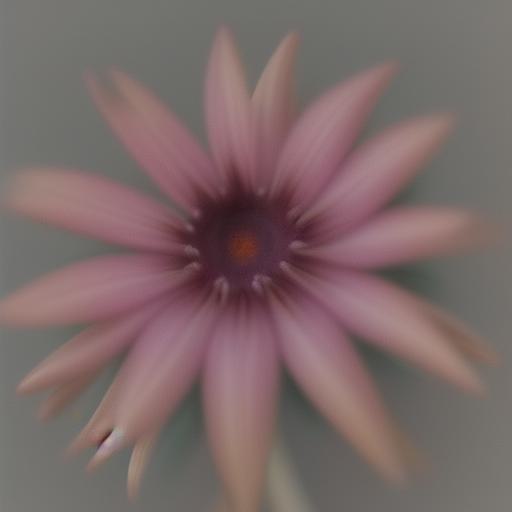} 
    & \includegraphics[width=0.98\linewidth,height=0.65\linewidth]{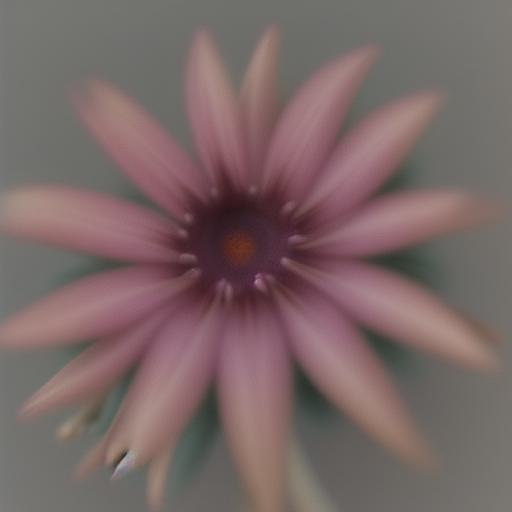} 
    & \includegraphics[width=0.98\linewidth,height=0.65\linewidth]{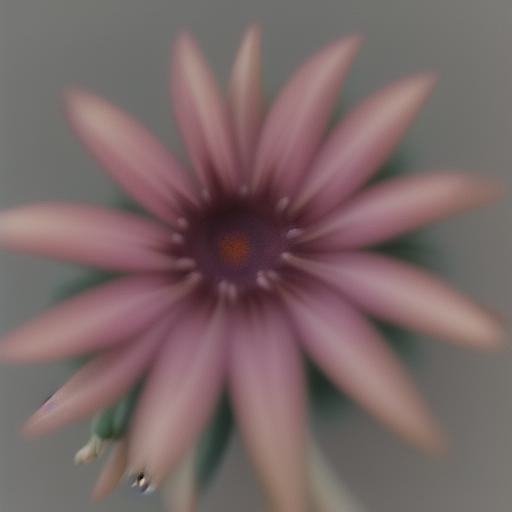}
\tabularnewline
\multirow{1}{*}
    {\rotatebox{90}{
        \begin{tabular}{c}
          \scriptsize V.Crafter2
        \end{tabular}
         \hspace{-1.8\linewidth}
    }} 
    & \animategraphics[autoplay,loop,width=0.98\linewidth,height=0.65\linewidth]{8}{appendix/figs/f16/flower_bloom_187/VideoCrafter/frame_}{0}{15}
    & \includegraphics[width=0.98\linewidth,height=0.65\linewidth]{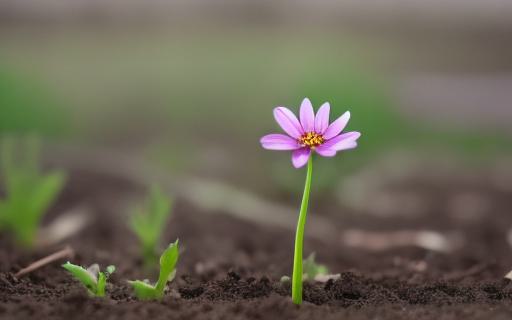}
    & \includegraphics[width=0.98\linewidth,height=0.65\linewidth]{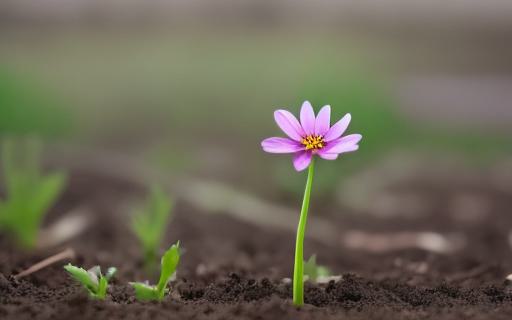} 
    & \includegraphics[width=0.98\linewidth,height=0.65\linewidth]{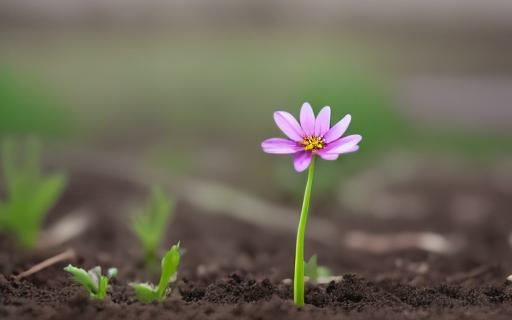} 
    & \includegraphics[width=0.98\linewidth,height=0.65\linewidth]{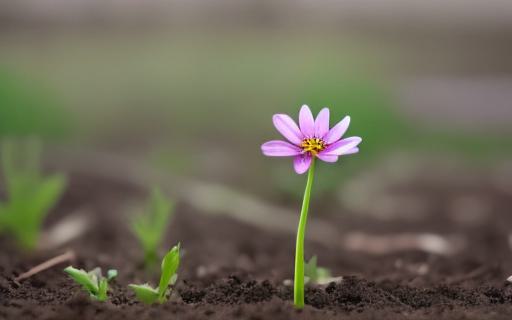} 
    & \includegraphics[width=0.98\linewidth,height=0.65\linewidth]{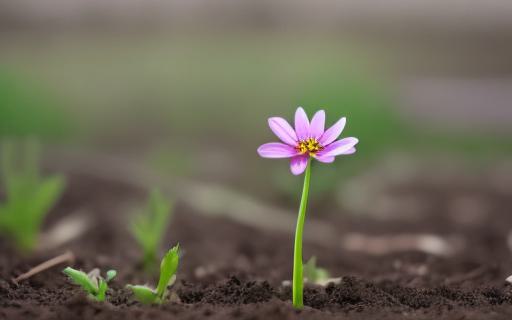}
\tabularnewline
\multirow{1}{*}
    {\rotatebox{90}{
        \begin{tabular}{c}
          \scriptsize \textbf{Ours}
        \end{tabular}
         \hspace{-1.0\linewidth}
    }} 
    & \animategraphics[autoplay,loop,width=0.98\linewidth,height=0.65\linewidth]{8}{appendix/figs/f16/flower_bloom_187/ours/frame_}{0}{15}
    & \includegraphics[width=0.98\linewidth,height=0.65\linewidth]{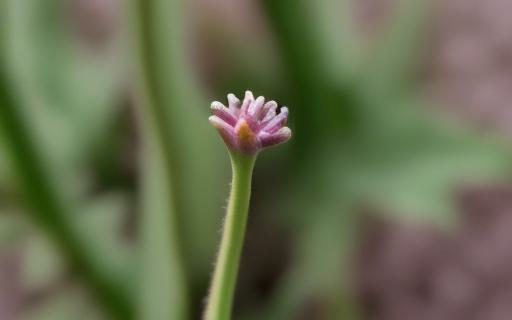}
    & \includegraphics[width=0.98\linewidth,height=0.65\linewidth]{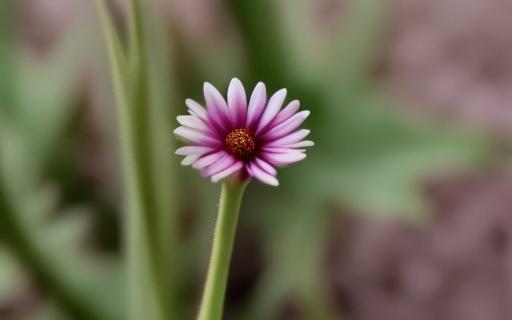} 
    & \includegraphics[width=0.98\linewidth,height=0.65\linewidth]{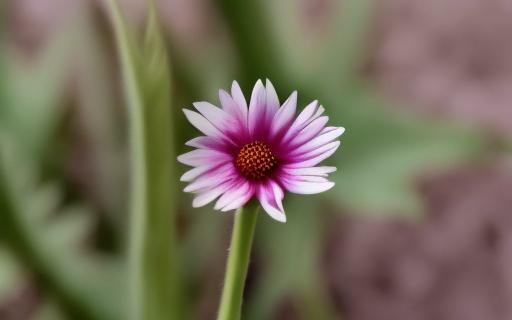} 
    & \includegraphics[width=0.98\linewidth,height=0.65\linewidth]{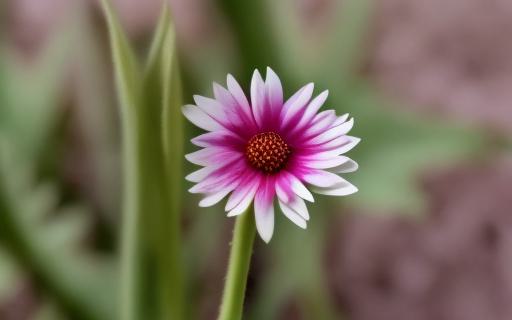} 
    & \includegraphics[width=0.98\linewidth,height=0.65\linewidth]{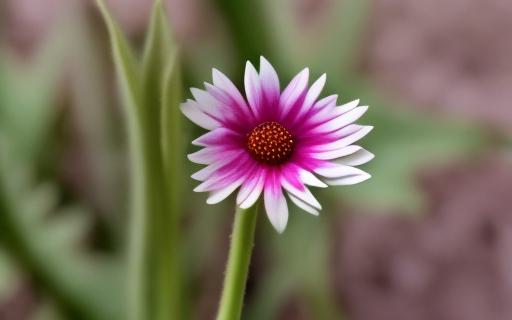}
\tabularnewline
\\
& \multicolumn{6}{c}{\begin{tabular}{c}
\myquote{A young girl is aging}
\end{tabular}} 
\tabularnewline
    \multirow{1}{*}
    {\rotatebox{90}{
          \scriptsize ModelScope
        \hspace{-1.5\linewidth}
    }} 
    & \animategraphics[autoplay,loop,width=0.98\linewidth,height=0.65\linewidth]{8}{appendix/figs/f16/aging_girl_717/ModelScope/frame_}{0}{15}
    & \includegraphics[width=0.98\linewidth,height=0.65\linewidth]{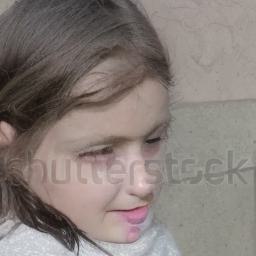}
    & \includegraphics[width=0.98\linewidth,height=0.65\linewidth]{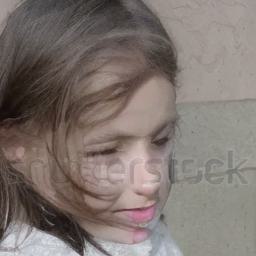} 
    & \includegraphics[width=0.98\linewidth,height=0.65\linewidth]{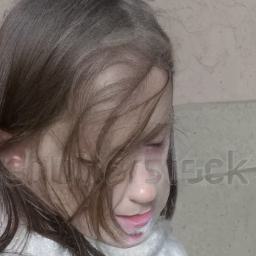} 
    & \includegraphics[width=0.98\linewidth,height=0.65\linewidth]{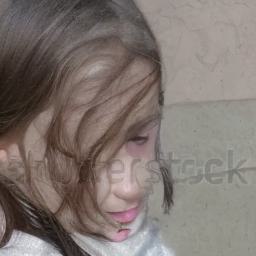} 
    & \includegraphics[width=0.98\linewidth,height=0.65\linewidth]{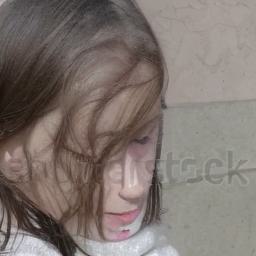}
\tabularnewline
\multirow{1}{*}
    {\rotatebox{90}{
        \begin{tabular}{c}
          \scriptsize LaVie
        \end{tabular}
        \hspace{-1.2\linewidth}
    }} 
    & \animategraphics[autoplay,loop,width=0.98\linewidth,height=0.65\linewidth]{8}{appendix/figs/f16/aging_girl_717/LaVie/frame_}{0}{15}
    & \includegraphics[width=0.98\linewidth,height=0.65\linewidth]{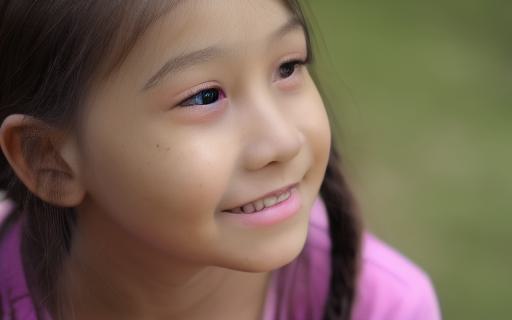}
    & \includegraphics[width=0.98\linewidth,height=0.65\linewidth]{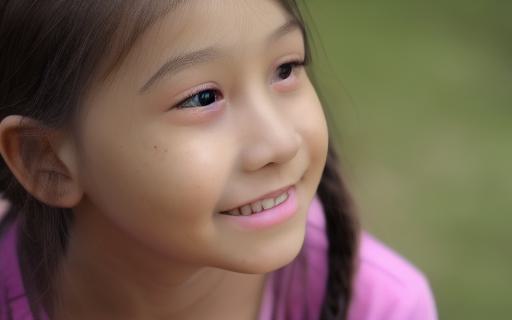} 
    & \includegraphics[width=0.98\linewidth,height=0.65\linewidth]{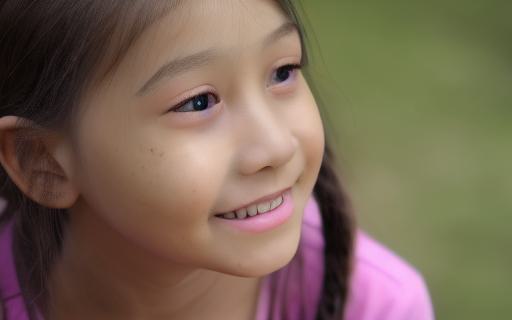} 
    & \includegraphics[width=0.98\linewidth,height=0.65\linewidth]{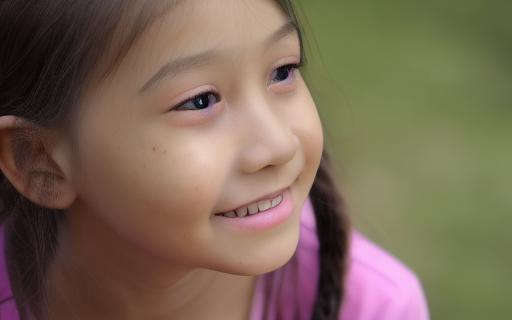} 
    & \includegraphics[width=0.98\linewidth,height=0.65\linewidth]{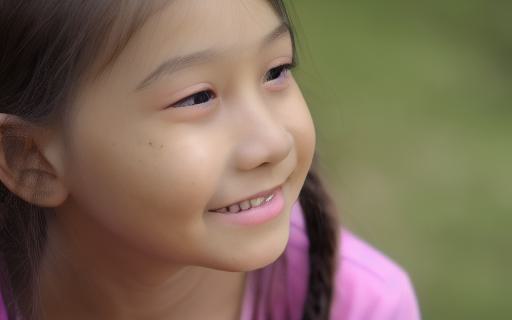}
\tabularnewline
\multirow{1}{*}
    {\rotatebox{90}{
        \begin{tabular}{c}
          \scriptsize AnimateDiff
        \end{tabular}
        \hspace{-2\linewidth}
    }} 
    & \animategraphics[autoplay,loop,width=0.98\linewidth,height=0.65\linewidth]{8}{appendix/figs/f16/aging_girl_717/AnimateDiff/frame_}{0}{15}
    & \includegraphics[width=0.98\linewidth,height=0.65\linewidth]{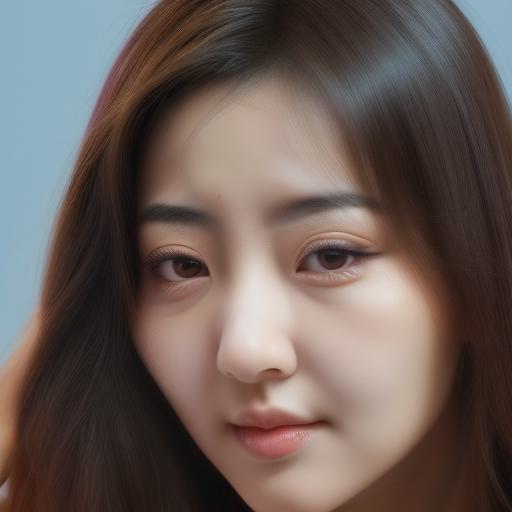}
    & \includegraphics[width=0.98\linewidth,height=0.65\linewidth]{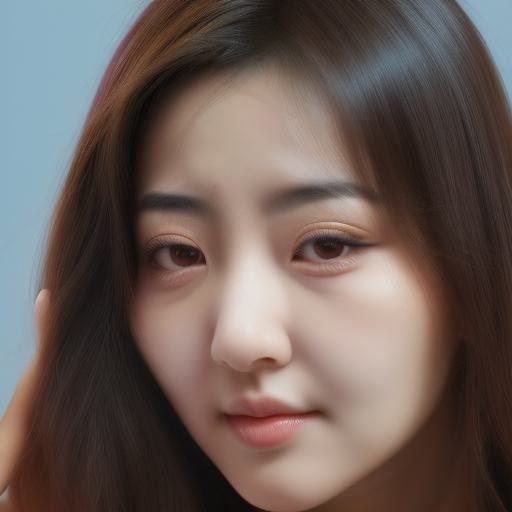} 
    & \includegraphics[width=0.98\linewidth,height=0.65\linewidth]{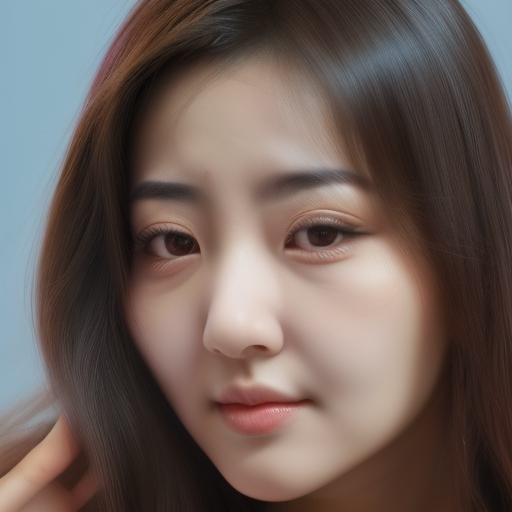} 
    & \includegraphics[width=0.98\linewidth,height=0.65\linewidth]{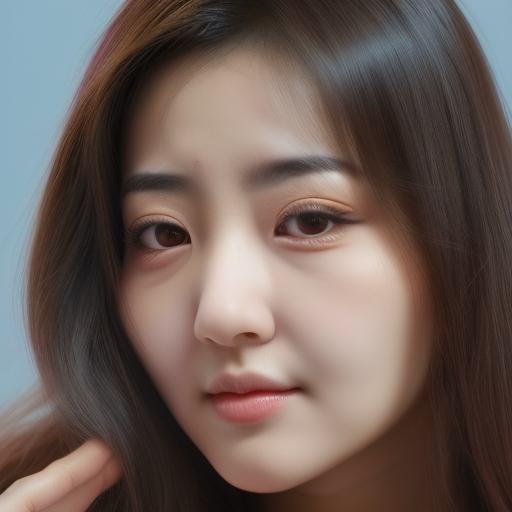} 
    & \includegraphics[width=0.98\linewidth,height=0.65\linewidth]{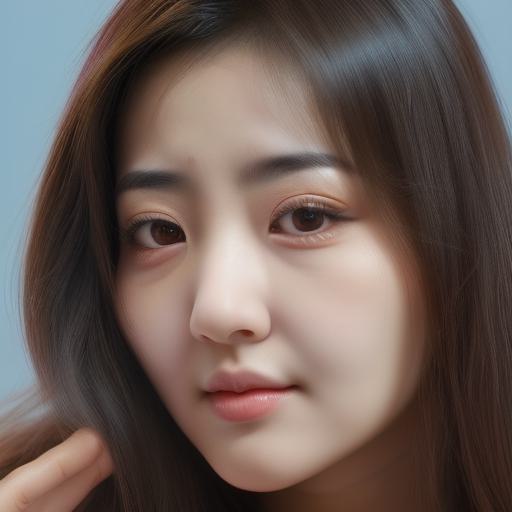}
\tabularnewline
\multirow{1}{*}
    {\rotatebox{90}{
        \begin{tabular}{c}
          \scriptsize V.Crafter2
        \end{tabular}
         \hspace{-1.8\linewidth}
    }} 
    & \animategraphics[autoplay,loop,width=0.98\linewidth,height=0.65\linewidth]{8}{appendix/figs/f16/aging_girl_717/VideoCrafter/frame_}{0}{15}
    & \includegraphics[width=0.98\linewidth,height=0.65\linewidth]{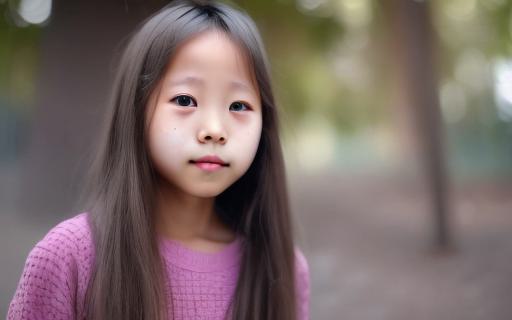}
    & \includegraphics[width=0.98\linewidth,height=0.65\linewidth]{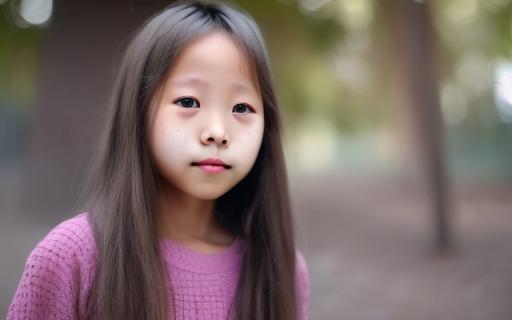} 
    & \includegraphics[width=0.98\linewidth,height=0.65\linewidth]{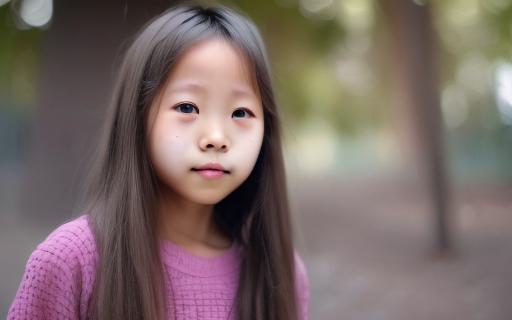} 
    & \includegraphics[width=0.98\linewidth,height=0.65\linewidth]{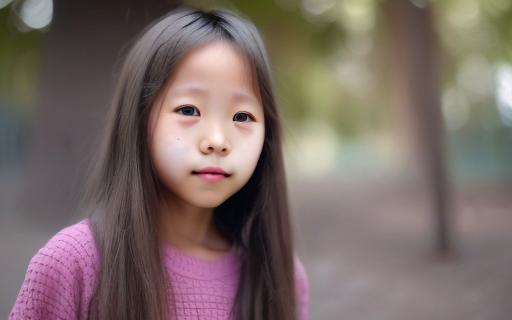} 
    & \includegraphics[width=0.98\linewidth,height=0.65\linewidth]{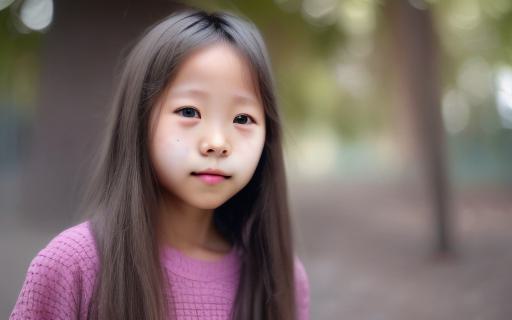}
\tabularnewline
\multirow{1}{*}
    {\rotatebox{90}{
        \begin{tabular}{c}
          \scriptsize \textbf{Ours}
        \end{tabular}
         \hspace{-1.0\linewidth}
    }} 
    & \animategraphics[autoplay,loop,width=0.98\linewidth,height=0.65\linewidth]{8}{appendix/figs/f16/aging_girl_717/ours/frame_}{0}{15}
    & \includegraphics[width=0.98\linewidth,height=0.65\linewidth]{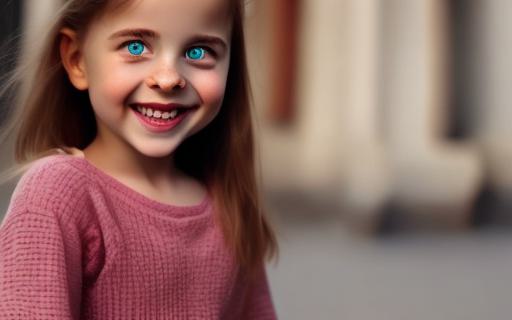}
    & \includegraphics[width=0.98\linewidth,height=0.65\linewidth]{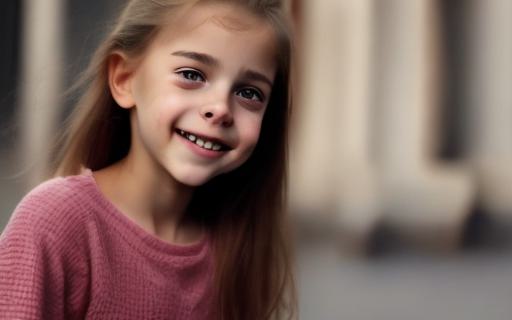} 
    & \includegraphics[width=0.98\linewidth,height=0.65\linewidth]{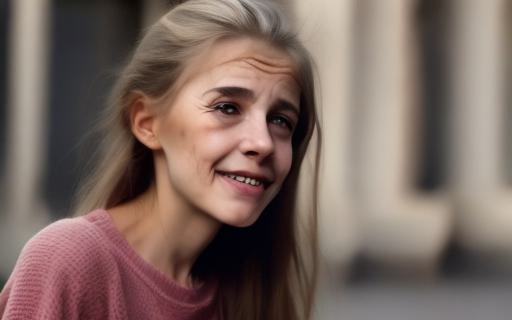} 
    & \includegraphics[width=0.98\linewidth,height=0.65\linewidth]{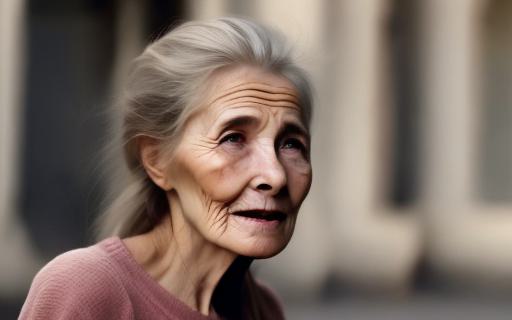} 
    & \includegraphics[width=0.98\linewidth,height=0.65\linewidth]{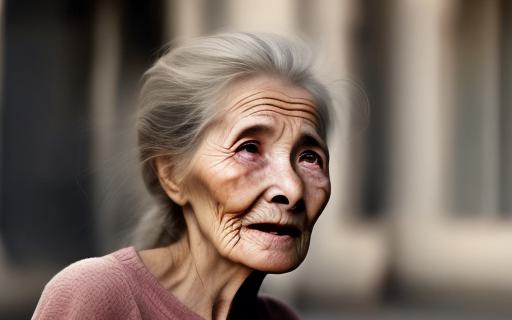}
\tabularnewline
\end{tabular}
\hfill{}
\par\end{centering}
\vspace{-0.5em}
\caption{Comparison with other T2V models on 16 frames generation. Our {\ourdm} consistently demonstrate improved video dynamics, resulting in more visually appealing content compared to the other methods. 
Note that the first column is a GIF, best viewed in \emph{Acrobat Reader}.
}
\label{fig:supp_compare_t2v_f16}
\end{figure}

\begin{figure}[t]
\begin{centering}
\setlength{\tabcolsep}{0.0em}
\renewcommand{\arraystretch}{0}
\par\end{centering}
\begin{centering}
\hfill{}%
 \begin{tabular}{
    m{0.03\linewidth}<{\centering} @{}
    m{0.16\linewidth}<{\centering} @{}
    m{0.16\linewidth}<{\centering} @{}
    m{0.16\linewidth}<{\centering} @{} %
    m{0.16\linewidth}<{\centering} @{}
    m{0.16\linewidth}<{\centering} @{}
    m{0.16\linewidth}<{\centering} @{}
    }

\tabularnewline
& \multicolumn{6}{c}{\begin{tabular}{c}
\myquote{The flower starts to bloom}
\end{tabular}} 
\tabularnewline
    \multirow{1}{*}
    {\rotatebox{90}{
          \scriptsize ModelScope
        \hspace{-1.5\linewidth}
    }} 
    & \animategraphics[autoplay,loop,width=0.98\linewidth,height=0.65\linewidth]{8}{appendix/figs/modelscope_vstar/flower_4088/orig/frame_}{0}{31}
    & \includegraphics[width=0.98\linewidth,height=0.65\linewidth]{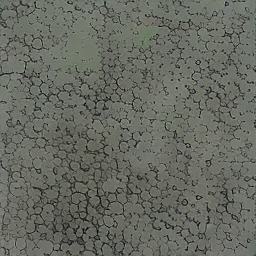}
    & \includegraphics[width=0.98\linewidth,height=0.65\linewidth]{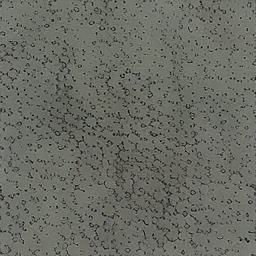} 
    & \includegraphics[width=0.98\linewidth,height=0.65\linewidth]{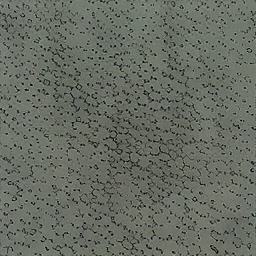} 
    & \includegraphics[width=0.98\linewidth,height=0.65\linewidth]{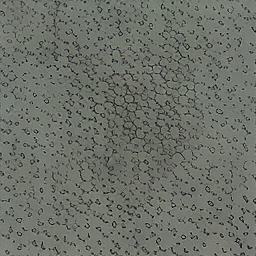} 
    & \includegraphics[width=0.98\linewidth,height=0.65\linewidth]{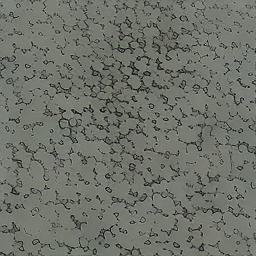}
\tabularnewline
\multirow{1}{*}
    {\rotatebox{90}{
        \begin{tabular}{c}
          \scriptsize +VSTAR
        \end{tabular}
        \hspace{-1.6\linewidth}
    }} 
    & \animategraphics[autoplay,loop,width=0.98\linewidth,height=0.65\linewidth]{8}{appendix/figs/modelscope_vstar/flower_4088/ours/frame_}{0}{31}
    & \includegraphics[width=0.98\linewidth,height=0.65\linewidth]{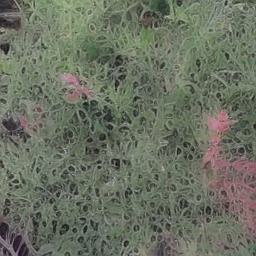}
    & \includegraphics[width=0.98\linewidth,height=0.65\linewidth]{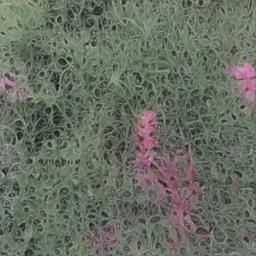} 
    & \includegraphics[width=0.98\linewidth,height=0.65\linewidth]{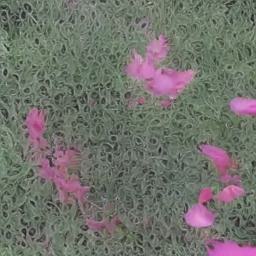} 
    & \includegraphics[width=0.98\linewidth,height=0.65\linewidth]{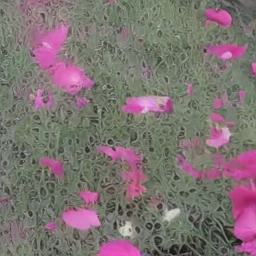} 
    & \includegraphics[width=0.98\linewidth,height=0.65\linewidth]{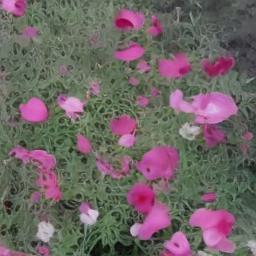}
\tabularnewline
\\

& \multicolumn{6}{c}{\begin{tabular}{c}
\myquote{A landscape transitioning from winter to spring}
\end{tabular}} 
\tabularnewline
    \multirow{1}{*}
    {\rotatebox{90}{
          \scriptsize ModelScope
        \hspace{-1.5\linewidth}
    }} 
    & \animategraphics[autoplay,loop,width=0.98\linewidth,height=0.65\linewidth]{8}{appendix/figs/modelscope_vstar/winter_spring_3027/orig/frame_}{0}{31}
    & \includegraphics[width=0.98\linewidth,height=0.65\linewidth]{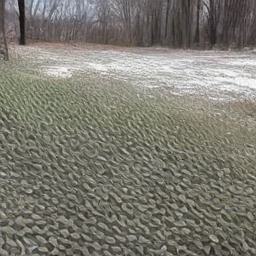}
    & \includegraphics[width=0.98\linewidth,height=0.65\linewidth]{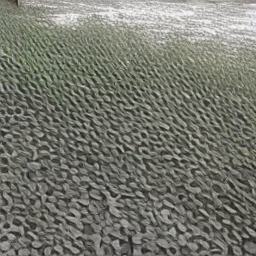} 
    & \includegraphics[width=0.98\linewidth,height=0.65\linewidth]{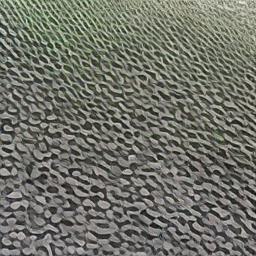} 
    & \includegraphics[width=0.98\linewidth,height=0.65\linewidth]{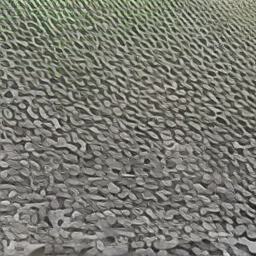} 
    & \includegraphics[width=0.98\linewidth,height=0.65\linewidth]{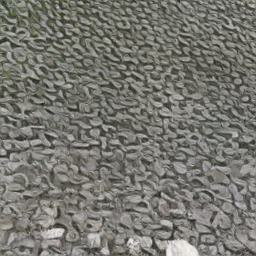}
\tabularnewline
\multirow{1}{*}
    {\rotatebox{90}{
        \begin{tabular}{c}
          \scriptsize +VSTAR
        \end{tabular}
        \hspace{-1.6\linewidth}
    }} 
    & \animategraphics[autoplay,loop,width=0.98\linewidth,height=0.65\linewidth]{8}{appendix/figs/modelscope_vstar/winter_spring_3027/ours/frame_}{0}{31}
    & \includegraphics[width=0.98\linewidth,height=0.65\linewidth]{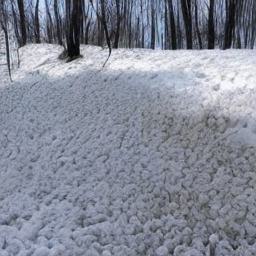}
    & \includegraphics[width=0.98\linewidth,height=0.65\linewidth]{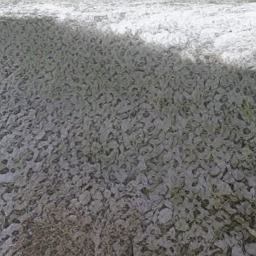} 
    & \includegraphics[width=0.98\linewidth,height=0.65\linewidth]{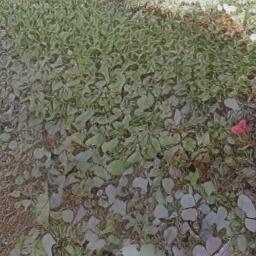} 
    & \includegraphics[width=0.98\linewidth,height=0.65\linewidth]{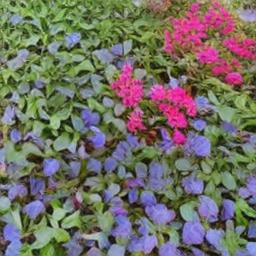} 
    & \includegraphics[width=0.98\linewidth,height=0.65\linewidth]{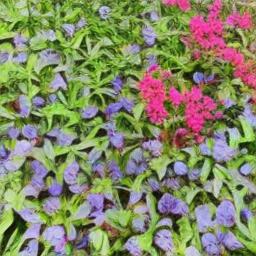}
\tabularnewline
\end{tabular}
\hfill{}
\par\end{centering}
\vspace{-0.5em}
\caption{Combination of ModelScope with {\ourdm} on 32 frames generation, which is double the length of the default option. The same random seed is used. 
ModelScope cannot generalize well to unseen frames, as discussed in Sec. \suppred{4.1}. Applying our {\ourdm} can significantly boost its generalization ability without fine-tuning required.
Note that the first column is a GIF, best viewed in \emph{Acrobat Reader}.
}
\label{fig:supp_modelscope_vstar}
\end{figure}

\begin{figure}[t]
\begin{centering}
\setlength{\tabcolsep}{0.0em}
\renewcommand{\arraystretch}{0}
\par\end{centering}
\begin{centering}
\hfill{}%
 \begin{tabular}{
    m{0.03\linewidth}<{\centering} @{}
    m{0.16\linewidth}<{\centering} @{}
    m{0.16\linewidth}<{\centering} @{}
    m{0.16\linewidth}<{\centering} @{} %
    m{0.16\linewidth}<{\centering} @{}
    m{0.16\linewidth}<{\centering} @{}
    m{0.16\linewidth}<{\centering} @{}
    }

\tabularnewline
& \multicolumn{6}{c}{\begin{tabular}{c}
\myquote{A day from sunrise to sunset on the beach}
\end{tabular}} 
\tabularnewline
    \multirow{1}{*}
    {\rotatebox{90}{
        \footnotesize Before
        \hspace{-1.0\linewidth}
    }} 
    & \animategraphics[autoplay,loop,width=0.98\linewidth,height=0.65\linewidth]{8}{appendix/figs/latent_opt/beach_sunrise_sunset_19/original/frame_}{0}{15}
    & \includegraphics[width=0.98\linewidth,height=0.65\linewidth]{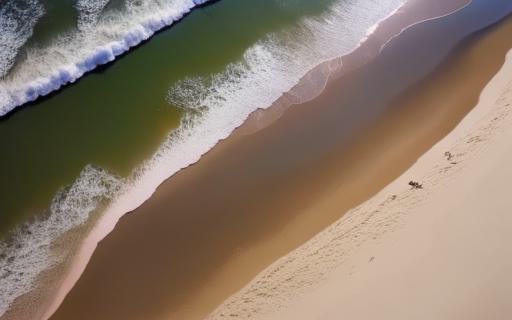}
    & \includegraphics[width=0.98\linewidth,height=0.65\linewidth]{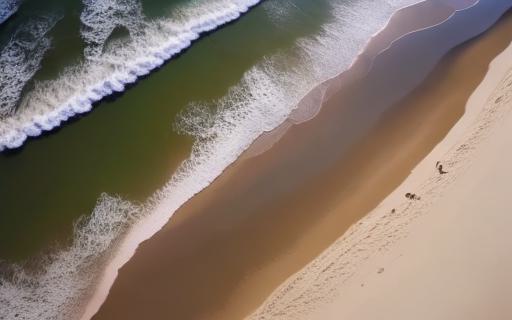} 
    & \includegraphics[width=0.98\linewidth,height=0.65\linewidth]{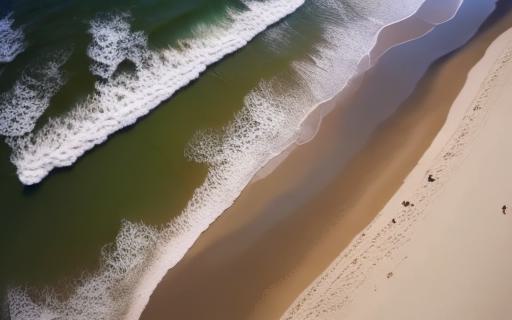} 
    & \includegraphics[width=0.98\linewidth,height=0.65\linewidth]{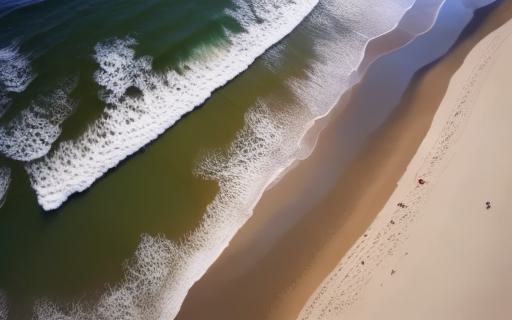} 
    & \includegraphics[width=0.98\linewidth,height=0.65\linewidth]{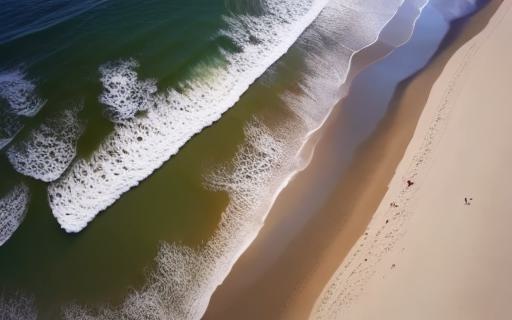}
\tabularnewline
\multirow{1}{*}
    {\rotatebox{90}{
        \begin{tabular}{c}
        \footnotesize After
        \end{tabular}
        \hspace{-1.2\linewidth}
    }} 
    & \animategraphics[autoplay,loop,width=0.98\linewidth,height=0.65\linewidth]{8}{appendix/figs/latent_opt/beach_sunrise_sunset_19/ours/frame_}{0}{15}
    & \includegraphics[width=0.98\linewidth,height=0.65\linewidth]{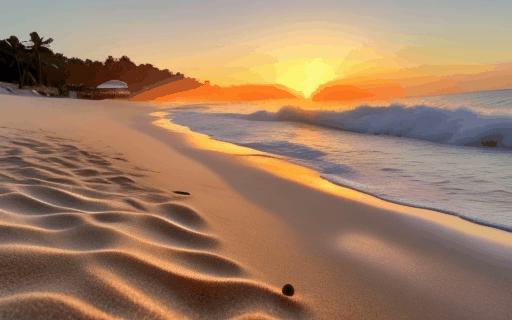}
    & \includegraphics[width=0.98\linewidth,height=0.65\linewidth]{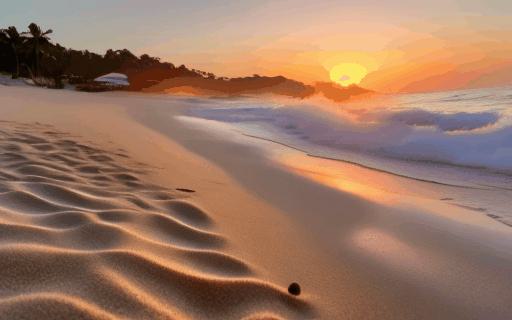} 
    & \includegraphics[width=0.98\linewidth,height=0.65\linewidth]{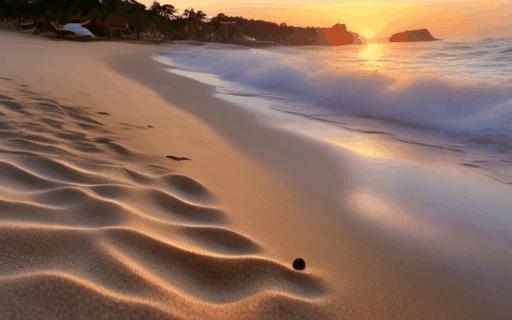} 
    & \includegraphics[width=0.98\linewidth,height=0.65\linewidth]{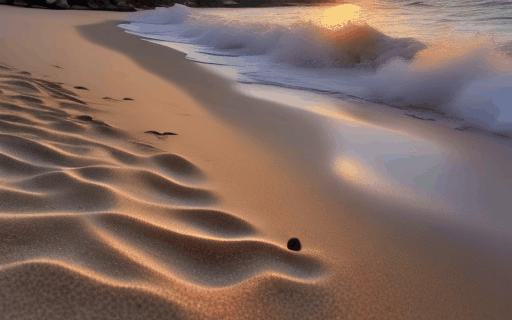} 
    & \includegraphics[width=0.98\linewidth,height=0.65\linewidth]{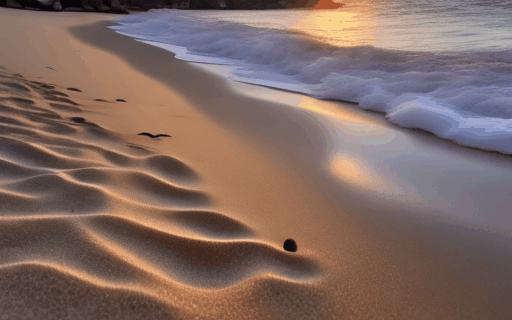}
\tabularnewline
\\

& \multicolumn{6}{c}{\begin{tabular}{c}
\myquote{The flower is blooming}
\end{tabular}} 
\tabularnewline
    \multirow{1}{*}
    {\rotatebox{90}{
        \footnotesize Before
        \hspace{-1.0\linewidth}
    }} 
    & \animategraphics[autoplay,loop,width=0.98\linewidth,height=0.65\linewidth]{8}{appendix/figs/latent_opt/flower_bloom_99/original/frame_}{0}{15}
    & \includegraphics[width=0.98\linewidth,height=0.65\linewidth]{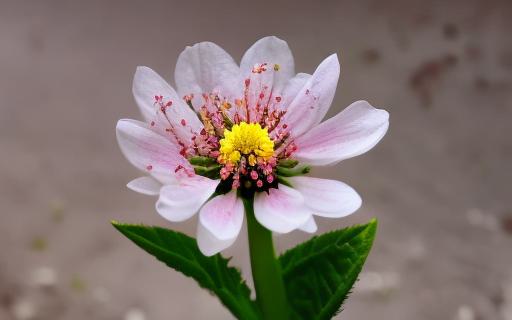}
    & \includegraphics[width=0.98\linewidth,height=0.65\linewidth]{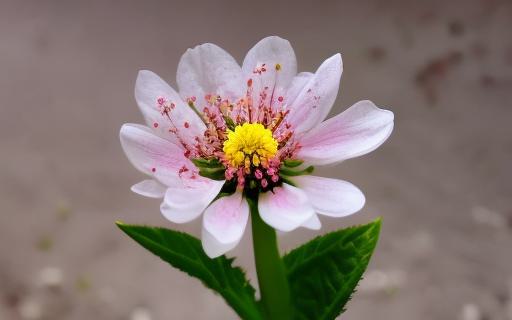} 
    & \includegraphics[width=0.98\linewidth,height=0.65\linewidth]{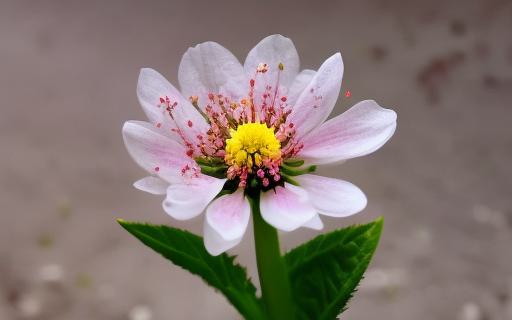} 
    & \includegraphics[width=0.98\linewidth,height=0.65\linewidth]{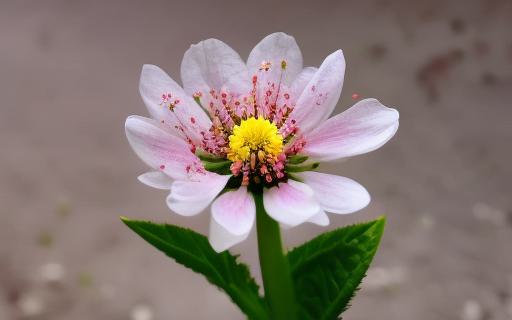} 
    & \includegraphics[width=0.98\linewidth,height=0.65\linewidth]{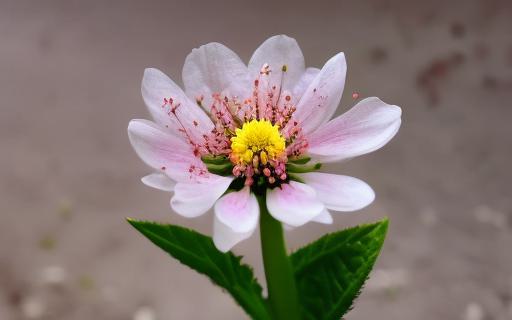}
\tabularnewline
\multirow{1}{*}
    {\rotatebox{90}{
        \begin{tabular}{c}
        \footnotesize After
        \end{tabular}
        \hspace{-1.2\linewidth}
    }} 
    & \animategraphics[autoplay,loop,width=0.98\linewidth,height=0.65\linewidth]{8}{appendix/figs/latent_opt/flower_bloom_99/ours/frame_}{0}{15}
    & \includegraphics[width=0.98\linewidth,height=0.65\linewidth]{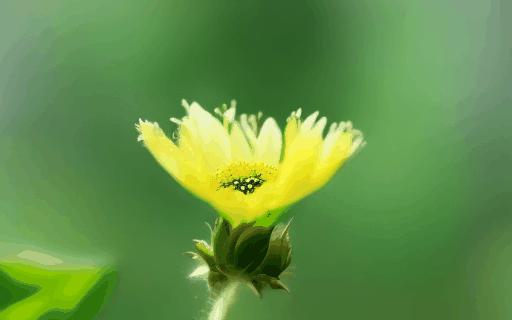}
    & \includegraphics[width=0.98\linewidth,height=0.65\linewidth]{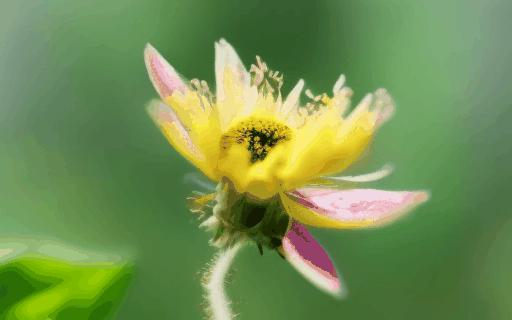} 
    & \includegraphics[width=0.98\linewidth,height=0.65\linewidth]{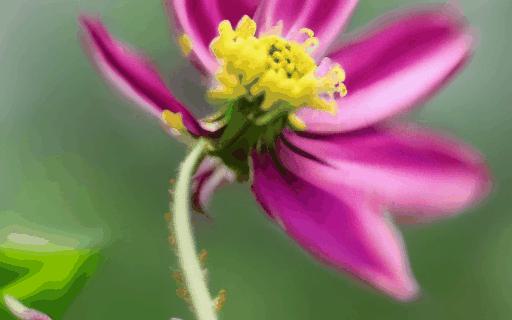} 
    & \includegraphics[width=0.98\linewidth,height=0.65\linewidth]{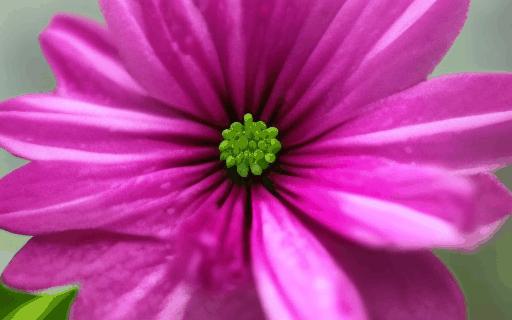} 
    & \includegraphics[width=0.98\linewidth,height=0.65\linewidth]{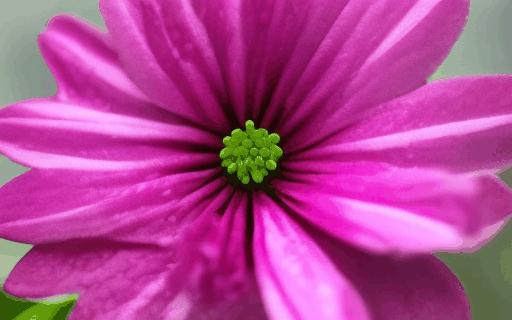}
\tabularnewline
\end{tabular}
\hfill{}
\par\end{centering}
\vspace{-0.5em}
\caption{Comparison of before and after applying initial noise optimization for 16 frames generation using the same single prompt and random seed. After optimization, the video dynamics has been enhanced, \ie, more visual variation has been introduced over time.
Note that the first column is a GIF, best viewed in \emph{Acrobat Reader}.
}
\label{fig:supp_noise_opt}
\end{figure}

\subsection{More visual results}
More qualitative visual results are provided in \cref{fig:supp_ours_f64,fig:supp_ours_f32,fig:supp_ours_f16}, in which the length of videos varies from the default 16 frames to longer ones with 64 frames.  
Intriguingly, all videos are generated in one \emph{single} pass using our {\ourdm}.
Additional results on the comparison with other T2V methods can be found in \cref{fig:supp_compare_t2v_f16}. It can be seen that our {\ourdm} consistently outperforms the other T2V models, demonstrating better dynamics with more visual changes over time complying with the text prompt.

\subsection{Combination of VSTAR and ModelScope}
In the main paper, we by default apply proposed {\ourdm} with state-of-the-art open-sourced T2V model VideoCrafter2\newcite{chen2024videocrafter2}. Nonetheless, {\ourdm} can also be combined with other pretrained T2V models to enhance their video dynamics. In \cref{fig:supp_modelscope_vstar}, we showcase that {\ourdm} can boost the long video generation ability of pretrained ModelScope\newcite{wang2023modelscope}, resulting in better visual quality and video dynamics.
However, due to the constrained capacity of the base model ModelScope, the overall synthesis results underperform those achieved by combining {\ourdm} with VideoCrafter2 as shown in the main paper.

\section{Optimization-based Generative Temporal Nursing}\label{sec:supp-optimization}
As a method of generative temporal nursing (GTN), our {\ourdm} is completely training- and optimization-free, and can be readily applied to frozen pretrained T2V models without introducing inference time overhead.
Additionally, we explored optimization-based GTN, assuming that a real reference video is available to guide the learning of desired dynamics.
Inspired by the temporal attention analysis detailed in Sec. \suppred{3.3}, we attempt to optimize the initial noise latents at inference time to align the attention maps of the given real video and the synthesized one, as outlined below.

Following \cite{ma2024magicme}, we parameterize the initial video latents of $N$ frames with a Multivariate Gaussian distribution $\epsilon \sim N(\mu,\mathbf{\Sigma}_N(\beta, \gamma)))$, where $\mathbf{\Sigma}_N(\beta, \gamma)$ denotes the covariance matrix:
\begin{equation}  \label{eq:supp-cov-matrix}
\mathbf{\Sigma}_N(\gamma) =
\begin{pmatrix}
\beta & \gamma & \gamma^2 & \cdots & \gamma^{N-1} \\
\gamma & \beta & \gamma & \cdots & \gamma^{N-2} \\
\gamma^2 & \gamma & \beta & \cdots & \gamma^{N-3} \\
\vdots & \vdots & \vdots & \ddots & \vdots \\
\gamma^{N-1} & \gamma^{N-2} & \gamma^{N-3} & \cdots & \beta \\
\end{pmatrix}.
\end{equation}
Given a real reference video, we can add noise to its clean latent and extract temporal attention maps $A_{t}^{ref}$ from its denoising process at the timestep $t$.
Then, we perform an initial noise optimization to match the temporal attention maps $A_t$ during the synthesis process with that of the reference ones:
\begin{align} \label{eq:supp-att-loss}
L_{Attn}  = \norm{A_{t}^{ref} - A_t}.
\end{align}
Furthermore, to prevent the initial noise from deviating significantly from the Gaussian Distribution, we  minimize the Kullback-Leibler divergence between the optimized latents and the standard Gaussian Distribution:
\begin{align} \label{eq:supp-KL-loss}
L_{KL}  = KL(N(\mu,\mathbf{\Sigma} || N(0,I)).
\end{align}
The joint optimization loss is a weighted sum of both loss terms:  
\begin{align} \label{eq:supp-all-loss}
\min_{\epsilon} L_{joint} = \min_{\mu, \beta, \gamma} L_{all}  = \min_{\mu, \beta, \gamma} L_{attn} + \lambda L_{KL},
\end{align}
where $\lambda$ is a weighting factor.

As shown in \cref{fig:supp_noise_opt}, after applying the initial noise optimization, the temporal dynamics of synthesized results from the same single prompt have noticeably improved, with more visual changes occurring throughout the video's progression.
However, this optimization-based technique increases the inference time and demands more memory, making it challenging to scale for longer videos.
In this regard, {\ourdm} stands out as more scalable and efficient, demonstrating its capability for facilitating long video generation.
Overall, we can see both approaches highlight that regularizing the temporal attention is an effective solution, 
suggesting that further exploration in this area could present an intriguing direction for future research.

\section{More details on Video Synopsis Prompting}\label{sec:supp-prompt}
Leveraging the in-context learning capability\newcite{brown2020GPT,hu2022context} of LLMs, we can guide them to perform the video synopsis prompting task automatically through prompting with a single concrete example.
For instance, we can instruct ChatGPT\newcite{chatgpt} with the following prompt:
\begin{tcolorbox}[]
I have a prompt "A landscape transitioning from winter to spring" for video generation. Can you split the process and describe the states separately? Each state is described in only one sentence and please consider the coherency between sub-prompts. Please be straightforward and do not use a narrative style. 

For example, for prompt "a boy is getting old", it can be divided into two states, e.g., "a young boy" and "an old man". 

Based on this example, can you provide the description? The number of states is not limited to two.
\end{tcolorbox}
Subsequently, ChatGPT can provide a detailed video synopsis that includes multiple visual states.
Once the LLM has learned such a task, we can then simply prompt it to execute the task  without reiterating the examples:
\begin{tcolorbox}[]
I have a prompt "A peony starts to bloom, in the field". Can you split the process and describe the states separately?
\end{tcolorbox}

Original prompts and the ChatGPT generated video synopsis are available in the \emph{prompt\_list.json} file included in the Supp. Material.

\section{Other Visualization}\label{sec:supp-other-vis}
\subsection{Visualization of Attention Regularization Matrix}
\begin{figure*}[t]
    \begin{centering}
    \setlength{\tabcolsep}{0.0em}
    \renewcommand{\arraystretch}{0}
    \par\end{centering}
    \begin{centering}
    \hfill{}%
	\begin{tabular}{@{}c@{}c@{}c}
        \centering
	$\sigma = 1$  
        & {\hspace{1em}} $\sigma = 4$ 
        & {\hspace{1em}} $\sigma = 8$
\tabularnewline
	\includegraphics[width=0.17\linewidth]{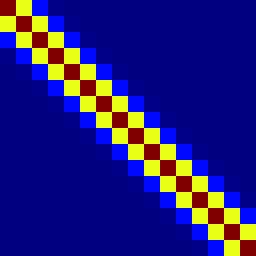} 
 & {\hspace{1em}}
	\includegraphics[width=0.17\linewidth]{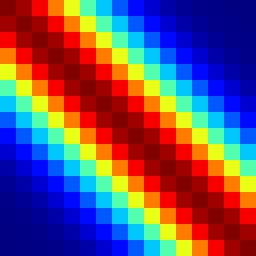} 
 & {\hspace{1em}} 
	\includegraphics[width=0.17\linewidth]{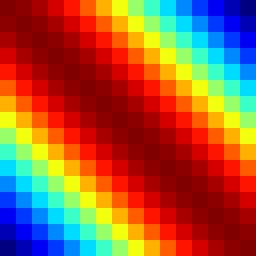} 
	\tabularnewline
\end{tabular}
\hfill{}
\par\end{centering}
\caption{
Visualization of regularization matrix $\Delta A$ with different standard deviation $\sigma$. A Smaller $\sigma$ can enhance the effect of regularization.
}
\label{fig:supp-reg-matrix}
\end{figure*}

The regularization matrix $\Delta A$ is designed as a symmetric Toeplitz matrix with values along the off-diagonal direction following a Gaussian distribution.
In \cref{fig:supp-reg-matrix}, we visualize $\Delta A$ with different standard deviations $\sigma$. 
We can see that as $\sigma$ decreases, the correlation increasingly concentrates on adjacent frames, thereby amplifying the regularization effect.

\subsection{Examples of Real Videos}

\begin{figure}[t]
\begin{centering}
\setlength{\tabcolsep}{0.0em}
\renewcommand{\arraystretch}{0}
\par\end{centering}
\begin{centering}
\hfill{}%
 \begin{tabular}{
    m{0.00\linewidth}<{\centering} @{}
    m{0.16\linewidth}<{\centering} @{}
    m{0.16\linewidth}<{\centering} @{}
    m{0.16\linewidth}<{\centering} @{} %
    m{0.16\linewidth}<{\centering} @{}
    m{0.16\linewidth}<{\centering} @{}
    m{0.16\linewidth}<{\centering} @{}
    }
& GIF &  \multicolumn{5}{c}{
\begin{tabular}{c}
Subsampled Frames
\end{tabular}} 
\tabularnewline
    & \animategraphics[autoplay,loop,width=0.98\linewidth,height=0.65\linewidth]{8}{appendix/figs/real_video/car_turn/frame_}{0}{15}
    & \includegraphics[width=0.98\linewidth,height=0.65\linewidth]{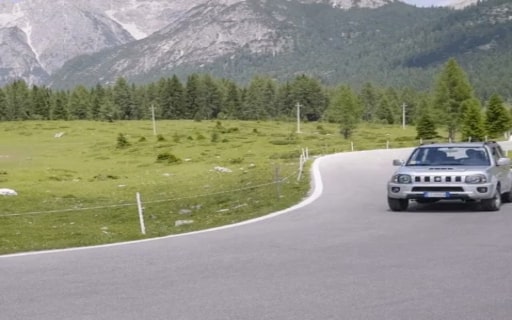}
    & \includegraphics[width=0.98\linewidth,height=0.65\linewidth]{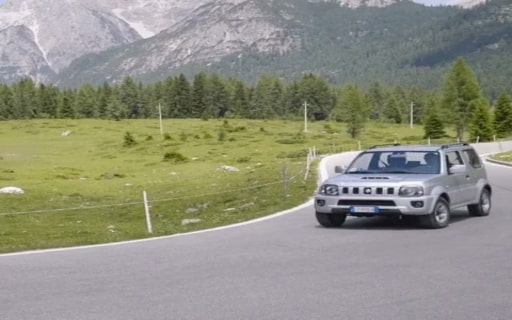} 
    & \includegraphics[width=0.98\linewidth,height=0.65\linewidth]{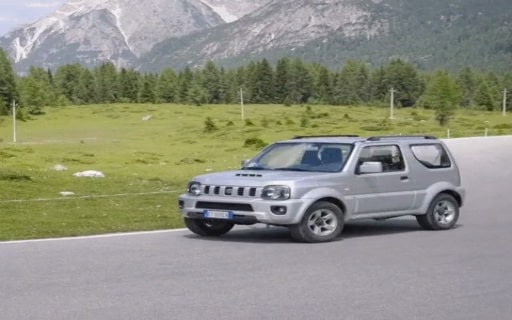} 
    & \includegraphics[width=0.98\linewidth,height=0.65\linewidth]{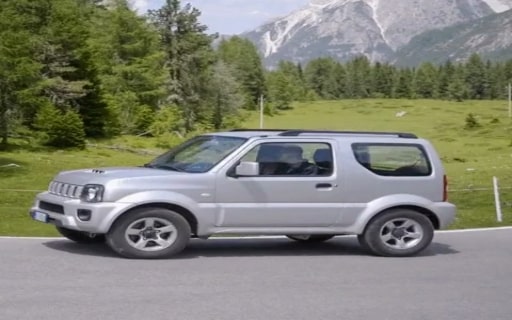} 
    & \includegraphics[width=0.98\linewidth,height=0.65\linewidth]{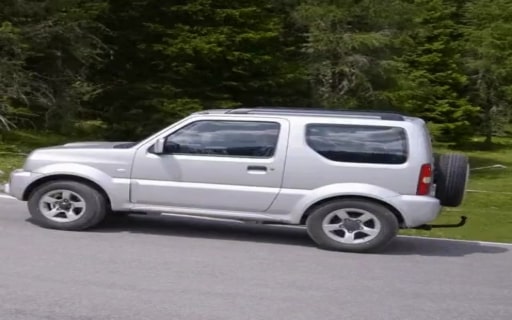}
\tabularnewline
    & \animategraphics[autoplay,loop,width=0.98\linewidth,height=0.65\linewidth]{8}{appendix/figs/real_video/rainbow_disappear/frame_}{0}{15}
    & \includegraphics[width=0.98\linewidth,height=0.65\linewidth]{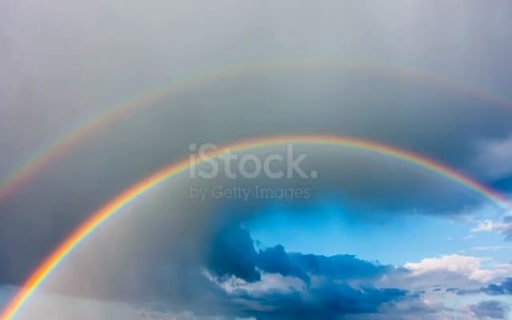}
    & \includegraphics[width=0.98\linewidth,height=0.65\linewidth]{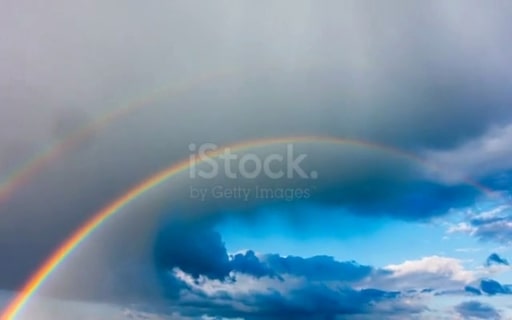} 
    & \includegraphics[width=0.98\linewidth,height=0.65\linewidth]{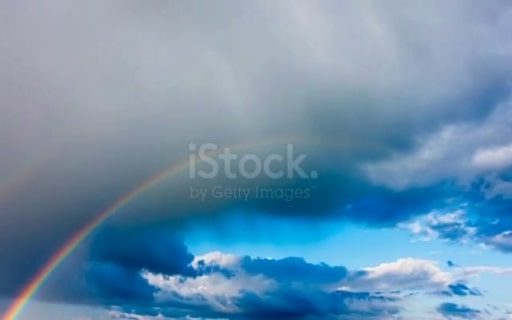} 
    & \includegraphics[width=0.98\linewidth,height=0.65\linewidth]{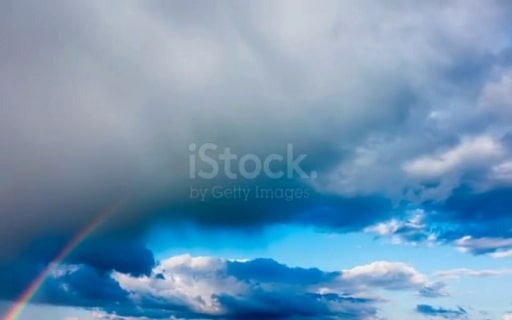} 
    & \includegraphics[width=0.98\linewidth,height=0.65\linewidth]{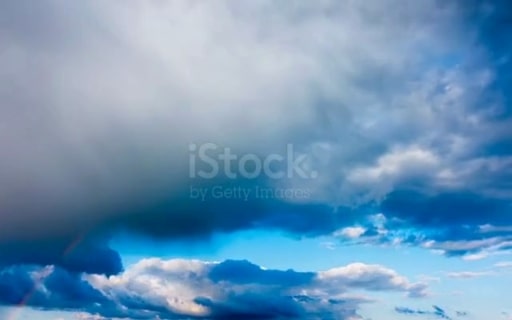}
\tabularnewline
    & \animategraphics[autoplay,loop,width=0.98\linewidth,height=0.65\linewidth]{8}{appendix/figs/real_video/transform_venom/frame_}{0}{15}
    & \includegraphics[width=0.98\linewidth,height=0.65\linewidth]{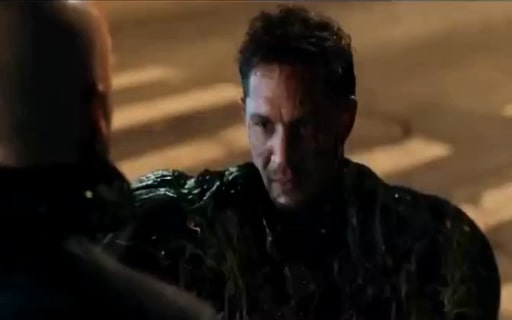}
    & \includegraphics[width=0.98\linewidth,height=0.65\linewidth]{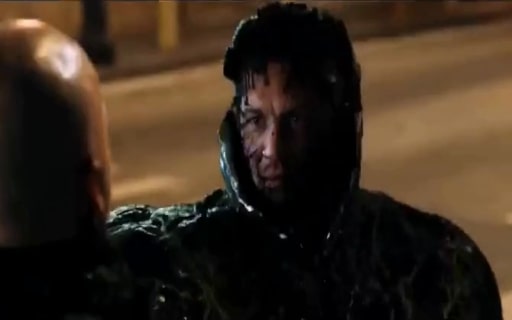} 
    & \includegraphics[width=0.98\linewidth,height=0.65\linewidth]{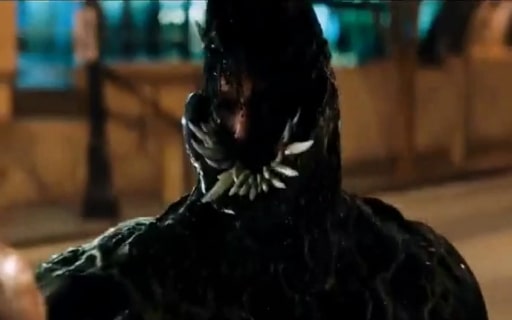} 
    & \includegraphics[width=0.98\linewidth,height=0.65\linewidth]{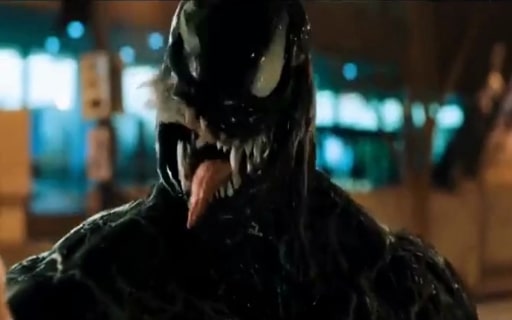} 
    & \includegraphics[width=0.98\linewidth,height=0.65\linewidth]{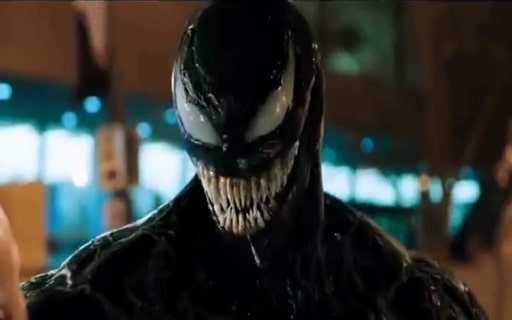}
\tabularnewline
    & \animategraphics[autoplay,loop,width=0.98\linewidth,height=0.65\linewidth]{8}{appendix/figs/real_video/timelapse_sun/frame_}{0}{15}
    & \includegraphics[width=0.98\linewidth,height=0.65\linewidth]{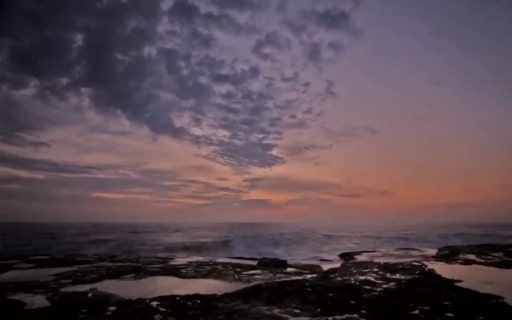}
    & \includegraphics[width=0.98\linewidth,height=0.65\linewidth]{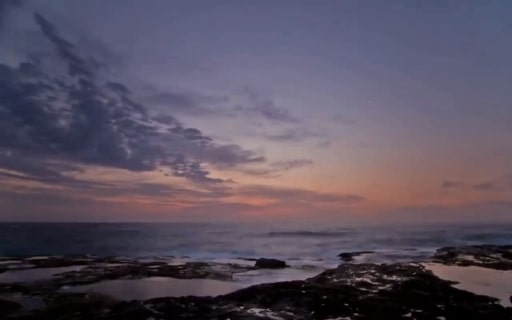} 
    & \includegraphics[width=0.98\linewidth,height=0.65\linewidth]{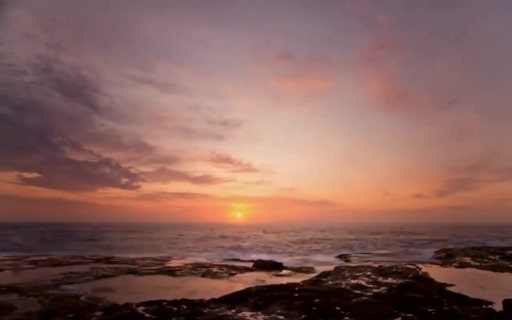} 
    & \includegraphics[width=0.98\linewidth,height=0.65\linewidth]{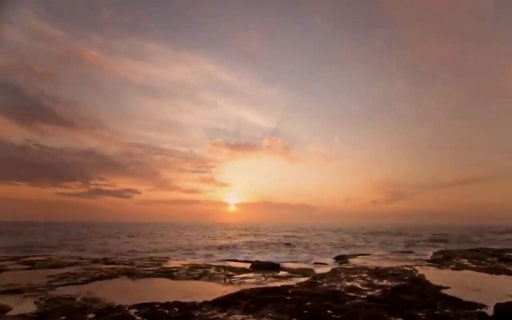} 
    & \includegraphics[width=0.98\linewidth,height=0.65\linewidth]{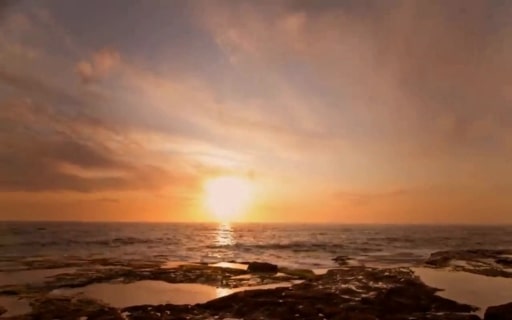}
\tabularnewline
    & \animategraphics[autoplay,loop,width=0.98\linewidth,height=0.65\linewidth]{8}{appendix/figs/real_video/flower/frame_}{0}{15}
    & \includegraphics[width=0.98\linewidth,height=0.65\linewidth]{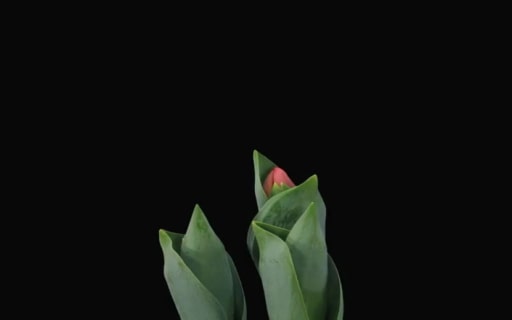}
    & \includegraphics[width=0.98\linewidth,height=0.65\linewidth]{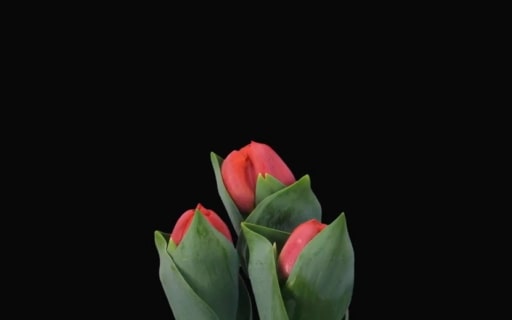} 
    & \includegraphics[width=0.98\linewidth,height=0.65\linewidth]{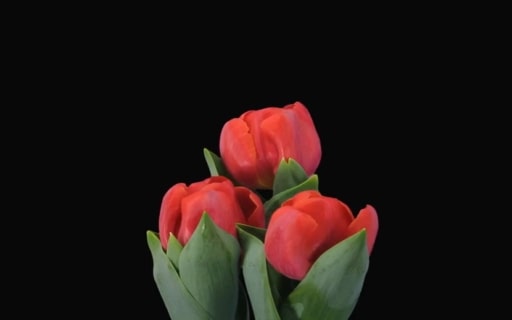} 
    & \includegraphics[width=0.98\linewidth,height=0.65\linewidth]{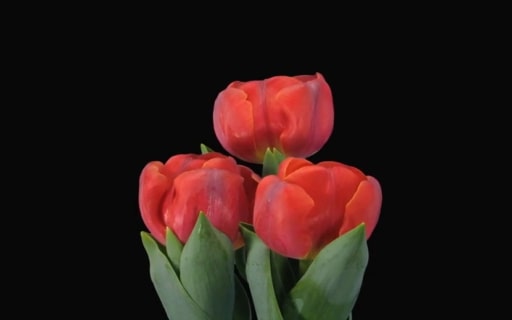} 
    & \includegraphics[width=0.98\linewidth,height=0.65\linewidth]{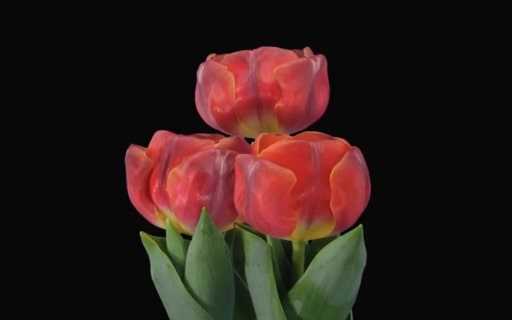}
\tabularnewline
\end{tabular}
\hfill{}
\par\end{centering}
\caption{Examples of diverse real dynamic videos. 
Note that the first column is a GIF, best viewed in \emph{Acrobat Reader}.
}
\label{fig:supp_real_video}
\end{figure}

We provide some examples of real dynamic videos in \cref{fig:supp_real_video}. They are collected from the web, DAVIS dataset\newcite{perazzi2016davis}, etc., showcasing diverse content. The selected videos contain ample visual changes over time, as opposed to static clips, and they are all captured using a single-camera setup.

\end{document}